\documentclass{article}
\usepackage{xcolor}
\usepackage{amssymb}

\usepackage[final]{corl_2025} %
\usepackage{graphicx}
\usepackage{booktabs}
\usepackage{pifont}
\usepackage{wrapfig}
\usepackage{subcaption}
\usepackage{amsmath}
\usepackage{amssymb}
\usepackage{placeins}
\usepackage[T1]{fontenc}

\usepackage{multirow}
\newcommand{\xmark}{\ding{55}}
\newcommand{\figref}[1]{Fig.~\ref{#1}}
\newcommand{\secref}[1]{Section~\ref{#1}}

\newcommand{\tabref}[1]{Table~\ref{#1}}

\makeatletter
\DeclareRobustCommand\onedot{\futurelet\@let@token\@onedot}
\def\@onedot{\ifx\@let@token.\else.\fi}
\def\eg{e.g\onedot} \def\Eg{E.g\onedot}
\def\ie{i.e\onedot}

\def\wrt{wrt\onedot}

\makeatother

\definecolor{darkgreen}{rgb}{0,0.7,0}
\definecolor{darkblue}{RGB}{31,119,180}
\definecolor{darkred}{RGB}{214,39,40}
\definecolor{mediumgray}{rgb}{0.5,0.5,0.5}
\definecolor{mediumteal}{rgb}{0,0.5,0.5}
\definecolor{naviblue}{RGB}{0,0,128}

\definecolor{ellisred}{rgb}{0.87,0.44,0.38} %
\definecolor{ellisgreen}{rgb}{0.69,0.90,0.52} %
\definecolor{elliscyan}{rgb}{0.29,0.77,0.74} %
\definecolor{ellisorange}{rgb}{0.89,0.55,0.28} %
\definecolor{ellisblue}{rgb}{0.41,0.61,0.86} %

\definecolor{customgray}{RGB}{136, 138, 133}

\newcommand{\boldparagraph}[1]{\vspace{0.2cm}\noindent{\bf #1:} }

\definecolor{darkgreen}{rgb}{0,0.7,0}

\newcommand{\pmsd}[1]{{\color{mediumgray}{\scriptsize $\pm$ #1}}}

\let\titleold\title
\renewcommand{\title}[1]{\titleold{#1}\newcommand{\thetitle}{#1}}
\def\maketitlesupplementary{
   \newpage
   \begin{center}
       \Large \textbf{\thetitle} \\  %
       \vspace{0.5em}
       \Large Supplementary Material \\ %
       \vspace{1.0em}
   \end{center}
}

\title{CaRL: Learning Scalable Planning\\ Policies with Simple Rewards}

\author{
  Bernhard Jaeger \quad \quad Daniel Dauner \quad \quad Jens Beißwenger \\
  \textbf{Simon Gerstenecker} \quad \quad \textbf{Kashyap Chitta\thanks{Work done at the University of Tübingen, Kashyap Chitta is currently affiliated with NVIDIA Research.}} \quad \quad \textbf{Andreas Geiger}\\
  University of Tübingen, Tübingen AI Center\\
  \texttt{\{bernhard.jaeger, daniel.dauner, a.geiger\}@uni-tuebingen.de}
}

\begin{document}
\maketitle

\begin{abstract}
We investigate reinforcement learning (RL) for privileged planning in autonomous driving. 
State-of-the-art approaches for this task are rule-based, but these methods do not scale to the long tail.
RL, on the other hand, is scalable and does not suffer from compounding errors like imitation learning.
Contemporary RL approaches for driving use complex shaped rewards that sum multiple individual rewards, \eg~progress, position, or orientation rewards. 
We show that PPO fails to optimize a popular version of these rewards when the mini-batch size is increased, which limits the scalability of these approaches.
Instead, we propose a new reward design based primarily on optimizing a single intuitive reward term: route completion. 
Infractions are penalized by terminating the episode or multiplicatively reducing route completion.
We find that PPO scales well with higher mini-batch sizes when trained with our simple reward, even improving performance.
Training with large mini-batch sizes enables efficient scaling via distributed data parallelism. 
We scale PPO to 300M samples in CARLA and 500M samples in nuPlan with a single 8-GPU node.
The resulting model achieves 64 DS on the CARLA longest6 v2 benchmark, outperforming other RL methods with more complex rewards by a large margin.
Requiring only minimal adaptations from its use in CARLA, the same method is the best learning-based approach on nuPlan.
It scores 91.3 in non-reactive and 90.6 in reactive traffic on the Val14 benchmark while being an order of magnitude faster than prior work.
\end{abstract}  

\section{Introduction}
\label{sec:intro}
We consider the task of privileged planning, in which an autonomous vehicle drives using ground truth perception inputs.
Such planners are traditionally rule-based \citep{Treiber2000PRE, Thrun2006JFR, Jaeger2021, Dauner2023CORL, Sima2024ECCV}. 
While rule-based approaches work well for regular driving \citep{Dauner2023CORL}, they require special scenario-specific rules to solve more complex scenarios \citep{Sima2024ECCV}, which is unlikely to scale to the long tail of driving scenarios.

Training neural planners with imitation learning (IL) is a popular alternative to rule-based approaches \citep{Renz2022CORL, Hallgarten2023ITSC, Cheng2024ICRA, Huang2024NNLS, Sun2024ARXIV, Guo2024ARXIV, Cheng2024ARXIV, Chen2025ARXIV}, because these methods can scale with data. Yet surprisingly, these methods underperform compared to rule-based or hybrid approaches \citep{Dauner2023CORL}.
A common explanation for this behavior is that IL suffers from a distribution shift between the open-loop training objective and the closed-loop inference task.

\begin{figure}[t]
\centering
    \includegraphics[width=\textwidth]{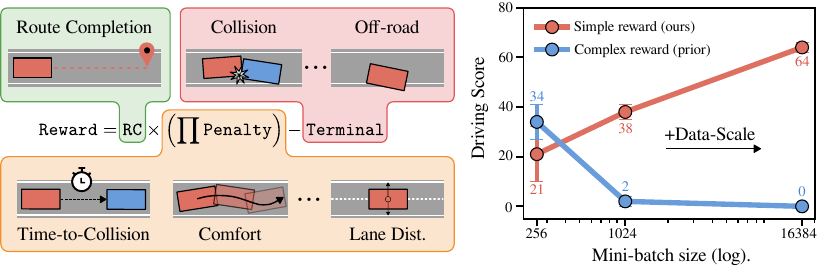}
    \put(-53,107.7){{\fontsize{7}{7}\selectfont \cite{Zhang2021ICCV}}}
    \label{fig:teaser_1}
    \vspace{-0.7cm}
   \caption{\textbf{Simple rewards scale with mini-batch size.} Typical rewards in driving consist of complex rewards that trade off many individual components. This limits scalability as PPO gets stuck in local minima with larger mini-batch sizes. We propose a simple alternative based on maximizing route completion that scales well with mini-batch size.}
\label{fig:teaser}
\vspace{-0.5cm}
\end{figure}

Closed-loop training \citep{Zhang2022ECCV}, in particular Reinforcement Learning (RL) \citep{Sutton2018MIT, Jaeger2024FTO}, is a promising alternative. 
A key problem in RL for driving is designing an appropriate reward function \citep{Knox2023AI}.
Early work in RL for driving used principled, simple rewards \citep{Kendall2019ICRA}, such as maximizing forward speed and terminating upon infractions.
Empirically, however, learning with such simple rewards only succeeded in environments without other actors \citep{Kendall2019ICRA, Wijmans2020ICLR, Fuchs2021RAL, Zeng2024CORL}, where simple behaviors suffice.
In settings where other dynamic actors are present, which require more complex behavior, such as reacting to sudden changes in the environment, simple rewards were empirically found to provide insufficient supervision \citep{Wurman2022Nature}.
As a result, recent designs provide denser feedback to simplify learning, combining many rewards additively, such as one reward for speed, orientation, position, and comfort \citep{Knox2023AI}.
The downsides of this approach are that the tradeoffs between rewards need to be carefully tuned, local minima are introduced, and the agent exploits undesirable shortcuts.
These tradeoffs harm scalability: For example, we observe that increasing the mini-batch size by a factor of 4 with a complex reward reduces the performance of Proximal Policy Optimization (PPO) \citep{Schulman2017ARXIV} drastically, due to a local optimum of the reward.
Larger mini-batch sizes smooth the gradient, which can make optimization more prone to local optima.
Additionally, many popular rewards \citep{Toromanoff2020CVPR, Zhang2021ICCV, Zhang2022ECCV, Chekroun2023ROBOTICS, Li2024ECCV} rely on simplistic rule-based planners to compute their reward terms. This upper bounds the performance, as the decisions from the rules are rewarded as if they were optimal.

We propose an alternative reward design that does not rely on rule-based planners.
The design learns policies with route completion (RC) as the only source of reward. To learn to avoid infractions, we end the episode upon any major infraction, \eg~collision, and reduce the obtained route completion multiplicatively while the agent is violating soft constraints, \eg, exceeding the speed limit.

While conceptually appealing, prior empirical findings suggest that such a simple reward does not provide enough feedback for the policy to learn effectively. 
We reproduce the most popular CARLA \citep{Dosovitskiy2017CORL} RL planner Roach \citep{Zhang2021ICCV} on the CARLA leaderboard 2.0 \citep{leaderboard22022ONLINE} and show that it naively performs worse when trained with the simpler reward.
However, unlike the complex reward, increasing the mini-batch size from 256 to 1024 drastically improves learning performance, with the simple reward outperforming the complex Roach reward.
This is illustrated in \figref{fig:teaser}.

Large mini-batch sizes, made possible by our reward, enable training with much more data, as data collection can be efficiently parallelized.
We scale our training to 300 million samples on CARLA, 30x more than prior work, with one compute server and mini-batch size of 16384. 
This results in a massive performance improvement.

Our final model, named CaRL, outperforms Roach and the recent world model RL-planner Think2Drive \citep{Li2024ECCV} on the longest6 v2 \citep{Chitta2023PAMI} benchmark, by 42 and 57 Driving Score (DS) respectively.
We also implement our method on the nuPlan \citep{Karnchanachari2024ICRA} simulator, which measures performance in realistic everyday scenarios via log replay.
We show that with minimal changes, our method can achieve 91 closed-loop score on the Val14 benchmark in both non-reactive and reactive traffic.
The resulting model is both 1.7 (non-reactive) and 7.9 (reactive) points better than the prior best learning based approach, Diffusion Planner \citep{Zheng2025ICLR}, while being 10x faster at inference time. 

Compared to other approaches, there has been little RL research in planning. 
One reason might be that there is no publicly available RL code base for both the CARLA and nuPlan simulators. 
To foster reproducible research, we published our code at \url{https://github.com/autonomousvision/CaRL}.
\section{Improving Scalability for RL}
\label{sec:scale}

Implementing RL on the CARLA leaderboard 2.0 is challenging because the simulator is slow \cite{Li2024ECCV}. In this section, we hence propose better hyperparameters and an optimized code base for RL in CARLA. These changes make it possible to train on-policy RL methods on the CARLA leaderboard 2.0 efficiently.
Related work is discussed in Appendix \secref{sec:related_work}.

\subsection{Preliminaries}
\label{sec:preliminaries}

We investigate the task of urban navigation from point A to B \citep{Chitta2023PAMI} with ground truth perception.
In sections \hyperref[sec:scale]{2} and \hyperref[sec:single_reward]{3}, we develop our method on CARLA \citep{Dosovitskiy2017CORL, leaderboard22022ONLINE} with a set of ablations where hyperparameters are held constant.
In \secref{sec:experiments}, we perform system-level comparisons with prior work and additionally evaluate our method on the nuPlan \citep{Karnchanachari2024ICRA} simulator.

\boldparagraph{Benchmark} We use the CARLA simulator version 0.9.15 and the longest6 v2 benchmark.
Longest6 v2 consists of 36 routes in towns 1-6 that are between 1 and 2 km long.
Along the route, between 5 and 21 pre-crash safety-critical scenarios \citep{Scenarios2022ONLINE} of 7 different types, as defined by the CARLA leaderboard 2.0, appear. 
Longest6 \citep{Chitta2023PAMI} is a popular CARLA leaderboard 1.0 benchmark. We converted the scenario definitions using the official scenario converter \citep{CARLA2024Online} to the leaderboard 2.0 style and use no modification to the leaderboard code.
We name the resulting benchmark longest6 v2 to avoid confusion, since results on longest6 v2 are not comparable to results on longest6.
The background traffic can drive up to 80 km/h in v2, 2-3 times faster than in leaderboard 1.0, making the benchmark significantly harder.
We focus on this benchmark because it uses the towns 1-6, which run substantially faster than the huge towns 12 and 13, allowing for faster RL and experimentation.
\textbf{Metrics:} We use the standard CARLA leaderboard 2.0 metrics Route Completion (RC), and Driving Score (DS).
DS multiplies RC with a penalty that decreases for every infraction. Details of the metrics can be found in \cite{LB2Metrics2024Online}.
\textbf{Algorithm:} We train our model with Proximal Policy Optimization (PPO) \citep{Schulman2017ARXIV}. PPO is one of the most popular RL algorithms due to its stable convergence and high asymptotic performance.
\textbf{Architecture:} We use bird's eye view semantic segmentation as input and predict the action of the car with a small CNN.
\textbf{Variance:} To combat the variance of CARLA evaluations \citep{Prakash2020CVPR, Behl2020IROS, Chitta2023PAMI} and RL training \citep{Islam2017ARXIV, McIlraith2018AAAI, Lynnerup2019CORL, Agarwal2021NEURIPS} every experiment averages 5 training seeds, each evaluated 3 times, unless noted otherwise. We report standard deviation across training seeds.

\boldparagraph{Simulation speed} Our method builds upon Roach \citep{Zhang2021ICCV}, the best open-source RL method trained with PPO on CARLA.
Roach has previously not been trained on the CARLA leaderboard 2.0.
We first reproduce Roach on the CARLA leaderboard 2.0 and then carefully modify it.
As already documented in \citep{Li2024ECCV}, the CARLA leaderboard 2.0 is slow for RL because of its slow reset step (up to 60 seconds).
To address this problem, we created a separate CARLA leaderboard 2.0, which we optimized for RL training. 
We use the original leaderboard 2.0 code for evaluation to ensure fair comparisons. 
The full list of optimizations can be found in the appendix and code. 
Noteworthy optimizations include avoiding town reloading by specializing each simulator instance to a particular town and pre-processing the A$^\star$ route planning.
Due to these optimizations, we were able to reproduce Roach with 10 million environment samples in 32 hours on a single A100 GPU.

 \subsection{Enabling higher learning rates}
Selecting a robust base set of hyperparameters for PPO is important in planning because the slow simulators and long training times prevent automatic hyperparameter tuning \citep{Fetterman2023ARXIV}.
In this section, we show that the PPO Atari hyperparameter set \citep{Schulman2017ARXIV, Huang2023ICLRBLOG} enables us to train better models faster, compared to the standard Roach hyperparameters used in CARLA \citep{Zhang2021ICCV}, reducing the computational cost of our experiments.
One advantage of the Atari hyperparameters is that they enable training at higher learning rates.
This effect can be intuitively explained by examining the number of off-policy steps performed by PPO.

\boldparagraph{Off-policy steps}
Policy gradient methods like PPO estimate the gradient of the policy's performance and perform gradient ascent.
It is possible to get an unbiased estimate of this gradient for on-policy learning, \ie, if only one optimizer step is used before the data is discarded and new data recollected. 
Strict on-policy methods like REINFORCE \citep{Williams1992ML} or A2C \citep{Mnih2016ICML} do this, but are sample inefficient.
PPO instead increases data efficiency by performing multiple gradient steps per iteration, approximating the off-policy policy gradient \citep{Degris2012ARXIV}.
The resulting approximation error is controlled by limiting policy change via the clipping heuristic.
Another heuristic is that the policy is only updated for a few gradient steps, limiting policy change implicitly.
Policies typically do not change a lot per gradient step, although this is dependent on the learning rate.

\tabref{tab:hyperparameter_values} compares the number of off-policy steps between the standard CARLA parameters from Roach \citep{Zhang2021ICCV} and the Atari \citep{Schulman2017ARXIV, Huang2023ICLRBLOG} parameters.

\begin{table}[h]
\vspace{-0.2cm}
\centering
    \begin{tabular}{c | c c}
        \toprule
        \textbf{Hyperparameter} & \textbf{Atari} \cite{Schulman2017ARXIV, Huang2023ICLRBLOG} & \textbf{CARLA} \cite{Zhang2021ICCV} \\
        \midrule
        Initial Learning rate & 0.00025 & 0.00001 \\
        Steps off-policy (steps$\times$epochs$-1$) &  ($4\times4-1$) = 15 &  ($48\times20-1$) = 959 \\
        \bottomrule
    \end{tabular}
    \vspace{0.1cm}
    \caption{\textbf{Default Hyperparameters of PPO for CARLA and Atari.}}
    \vspace{-0.0cm}
    \label{tab:hyperparameter_values}
\end{table}

Roach performs 959 off-policy steps (the first step is on-policy), 60 times more than the Atari parameters.
One would expect that this introduces large policy changes and hence off-policy error, and it is surprising that Roach converges at all.
The reason for this is the small learning rate, which is 25x lower than the Atari learning rate, leading to smaller policy changes, hence balancing this effect.
Training Roach with the (higher) Atari learning rate and schedule, without changing other hyperparameters, leads to a degenerate policy resulting in a drop of DS from $22\pm14$ to $2\pm1$ DS.

We train Roach with its hyperparameters and the Atari hyperparameters. The results are shown in \tabref{tab:hyperparameter_performance}. Surprisingly, we observe a 10-hour reduction in training time while the policy achieves 11 DS better performance.
The reduction in training time comes from using 5 times fewer epochs, whereas the performance improvement might come from a combination of hyperparameters.
We therefore use the Atari hyperparameters as the basis for our model.

\begin{table}[h]
\centering
    \begin{tabular}{c | c | c | l l }
        \toprule
        \textbf{Hyperparameter set} & \textbf{Train Time h} $\downarrow$ & \textbf{Output} & \textbf{DS} $\uparrow$ & \textbf{RC} $\uparrow$  \\
        \midrule
        CARLA \cite{Zhang2021ICCV} & {32} \pmsd {2}  & mode \cite{Zhang2021ICCV} & {22} \pmsd {14} & {77} \pmsd {19}\\
        \midrule
        \multirow{2}{*}{Atari \cite{Schulman2017ARXIV, Huang2023ICLRBLOG}} & \multirow{2}{*}{\textbf{22} \pmsd {1}} & mode \cite{Zhang2021ICCV} & {33} \pmsd {10} & \textbf{90} \pmsd {6}  \\
        &  & mean & \textbf{34} \pmsd {7} & {86} \pmsd {6} \\
        \bottomrule
    \end{tabular}
    \vspace{0.2cm}
    \caption{\textbf{Atari vs CARLA hyperparameter set on CARLA.} The Atari hyperparameters train faster (-10 h using one A100 (40GB) and 14 EPYC 7742 CPU cores), achieve better performance (+11 DS), and have lower variance (-4 std).}
    \label{tab:hyperparameter_performance}
    \vspace{-0.0cm}
\end{table}

Roach uses a Beta distribution to parametrize its action space. 
During inference, sampling from the distribution is typically replaced with a statistic of the distribution to turn off exploration. 
Roach proposes to use the mode of the distribution when $\alpha, \beta > 1$.
Using the mode biases the policy's behavior, as PPO optimizes the expected performance under the action distribution (the mean).
We observe that using the mean achieves 1 DS better performance with the same models, and reduces the standard deviation by 3 points.

\section{Optimizing a Single Reward}
\label{sec:single_reward}

Many current RL planners \cite{Toromanoff2020CVPR, Zhang2021ICCV, Chekroun2023ROBOTICS, Li2024ECCV} use a reward function that consists of multiple additive terms, including terms to reward the inverse distance to the desired vehicle speed, desired vehicle position, and desired orientation.
This is reminiscent of imitation learning because these ``desired'' states are computed using handcrafted rules akin to a rule-based planning method, or even directly with human labels \cite{Zhang2025ArXiV}. 
As noted in the Appendix of \cite{Zhang2021ICCV}, their rule-based planner is suboptimal and sometimes assigns higher rewards to suboptimal states. 
The advantage of these complex, shaped rewards is that they can simplify learning. 
They provide dense, rich feedback, which simplifies the credit assignment problem of RL.

These rewards have, however, many downsides. The handcrafted rules, being suboptimal, can upper-bound performance, and these rewards introduce loopholes and local minima that the optimization can get stuck in, or the agent can exploit.
One example of a failure case we observe with the Roach reward \cite{Zhang2021ICCV} is that the policy sometimes learns to wait at green traffic lights. This is a simple behavior that obtains a large return with the Roach reward. The model is constantly rewarded by the optimal speed reward if it has speed 0 at a red light. Because traffic lights are often red for a long time and green for a short amount of time in CARLA, the policy sometimes learns to exploit that it can just wait for the next red light for easy rewards, when the light turns green for a short duration. This problem is illustrated in Appendix \figref{fig:wait_green_light_1}. Eventually, many training seeds escape this local minima, but we still observe that the final policy drives slower than usual when approaching green lights, presumably trying to catch the next red light. This is illustrated in Appendix \figref{fig:wait_green_light_2}.

To address these concerns, we are proposing a new type of reward design:

\vspace{-0.25cm}
\begin{equation}
    r_t = RC_t * \left(\prod p_t\right) - T
\label{eqn:infraction_multiplier}
\end{equation}
\vspace{-0.25cm}

The reward at time step $t$ is computed by multiplying the percentage of the route completed during this simulator time step $RC_t$ with soft penalty factors $p_t \in [0,1]$.

The soft penalties $p_t$ are $1$ if their condition is not violated and otherwise have a value $\in [0,1)$ depending on the type of infraction.
Soft penalties are constraints that the agent should typically adhere to, such as staying within the speed limit, but may violate in order to avoid a hard penalty like a collision.
It is important that soft penalty factors that the agent cannot avoid violating early on in training, such as comfort, need to be $>0$.
Otherwise, the agent would receive no reward.
It is possible to apply a soft penalty for multiple frames \eg~to penalize actions while the car is not moving or increase the penalty strength. 

Any major infraction, like a collision, is treated as a hard penalty. 
Hard penalties simply end the episode. 
The principle is that any hard constraints of the optimization problem are formulated as terminal states, ensuring that violating them results in suboptimal performance as no further reward can be collected.
$T$ is a terminal penalty that is applied at the end of the episode, depending on the infraction.
It can be used to induce an ordering between infractions, but it needs to be sufficiently small that the agent is not penalized for starting to drive early on in training. We use $T=1$ for collisions and red light infractions, and $T=0$ for everything else.
The full description of penalties we use can be found in Appendix \secref{sec:reward}.

The reward design follows the following principles: 
(1) \textit{The amount of reward obtainable is finite.} There is only 100 RC to be collected in total, preventing infinite reward loopholes like the car waiting at a green light.
(2) \textit{The reward only specifies what to do, not how to do it.} Our reward does not include any rule-based planners.
(3) \textit{The global optimum of the reward is the same as the global optimum of the metric} \cite{Jaeger2024FTO}. This ensures that solving the learning problem results in an optimal policy, \wrt~the metric. Our reward closely aligns with the DS metric \cite{Koltun2020ICMLWorkshop}, with some additional penalties like speeding. The agent obtains the optimal reward when completing the route without infraction, which also yields the optimal DS.

\begin{table}[t]
\centering
    \begin{tabular}{l | c | c | c }
        \toprule
        \textbf{Reward} & \textbf{Local Hints?} & \textbf{Mini-Batch Size 256} & \textbf{Mini-Batch Size 1024} \\
        \midrule
        Roach \cite{Zhang2021ICCV} & $\checkmark$ & {34} \pmsd {7}  & {2} \pmsd {2} \\
        CaRL (Ours) & \xmark & {21} \pmsd {11} & \textbf{38} \pmsd {3} \\
        \bottomrule
    \end{tabular}
    \vspace{0.2cm}
    \caption{\textbf{Reward and mini-batch size.} The metric is DS. PPO gets stuck in a local minimum when optimizing a complex reward at large mini-batch sizes. The simple reward does not have this problem. PPO even improves with a larger mini-batch size.}
    \label{tab:reward}
    \vspace{-0.0cm}
\end{table}

Our reward is simple and gives much fewer local hints to the policy than other rewards. It avoids the problems of typical reward designs, but might be harder to optimize.
\tabref{tab:reward} shows that PPO struggles to perform well without the rule-based dense supervision of the Roach reward, dropping performance by 13 points at the standard mini-batch size of 256.
Interestingly, this trend reverses when we scale the mini-batch size to 1024. 
When training PPO with the dense Roach reward at 1024 mini-batch size, we observe that the model gets stuck in the local minimum of not driving on many routes, which yields good position and orientation rewards but suboptimal speed rewards. 
At the end of training, the model started to learn to accelerate but hasn't learned to steer yet, leading to a poor driving policy with 2 DS. One hypothesis is that larger mini-batch sizes smooth the optimization, making it more prone to local minima.
In contrast, our reward does not have these tradeoffs between different reward components, and PPO improves drastically when increasing the mini-batch size to 1024, improving DS by 17 points.
Additionally, our reward outperforms the roach reward by 4 points in DS and reduces training variance from $\pm 7$ to $\pm 3$.

\subsection{Scaling Data}

There are two ways to increase mini-batch size in PPO while keeping the number of gradient steps per epoch unchanged. 
In the previous section, we increased the mini-batch size by performing 4x more simulator steps per PPO iteration with 4x fewer PPO iterations, because this method has similar resource requirements.
If more resources are available, the number of parallel simulators can instead be increased, which raises the overall sample throughput.
This also requires additional GPUs for model inference and training, which we utilize via the DD-PPO \cite{Wijmans2020ICLR} scaling approach without preemption for the experiment in this section.
DD-PPO works like distributed data parallelism \cite{Hillis1986ACM} from supervised learning.
\begin{wraptable}{r}{0.5\textwidth}
\vspace{-0.4cm}
\centering
    \begin{tabular}{l | c | c c }
        \toprule
        \textbf{Samples} &  \textbf{Mini-batch} & \textbf{DS} $\uparrow$ & \textbf{RC} $\uparrow$ \\
        \midrule
        10M & 1024 & {31} \pmsd {7} & {63} \pmsd {8} \\
        300M & 16384 & \textbf{64} \pmsd {2} & \textbf{82} \pmsd {1} \\
        \bottomrule
    \end{tabular}
    \caption{\textbf{Effect of scale.}}
    \label{tab:scale}
    \vspace{-0.4cm}
\end{wraptable}
Prior works trained PPO for driving with 1-10 million samples \cite{Zhang2021ICCV, Li2024ECCV}.
This is typical for classic PPO \cite{Schulman2017ARXIV}, but 10-100 times less than what recent breakthroughs in robotics used \cite{Kaufmann2023Nature, Zeng2024CORL}.
Prior work may not have scaled up data because increasing mini-batch size, which is required to scale efficiently, can yield degenerate performance, as \tabref{tab:reward} showed.
Our reward enables us to scale our model from 10 million to 300 million samples using a single node with 8 A100 (40G) GPUs and 108 EPYC Rome CPU cores for 1 week.
\tabref{tab:scale} shows that increasing the samples and mini-batch size leads to a large improvement of 33 DS.
The baseline features some changes compared to the model in the last section, which are discussed in Appendix \secref{sec:method_changes}.
The 300M result is an average of three seeds.
\section{Experiments}
\label{sec:experiments}

This section shows the advantages of our method over several other planning approaches on the CARLA longest6 v2 \cite{Chitta2023PAMI} benchmark and nuPlan Val14 \cite{Dauner2023CORL}.
All experiments were run by us with the same setting to ensure a fair comparison and avoid the common benchmarking errors that permeate the literature \cite{Jaeger2024Online}.

\subsection{CARLA}
\label{sec:experiments_carla}
\vspace{-0.44cm}
\boldparagraph{Benchmark} We use the longest6 v2 benchmark, and metrics as described in \secref{sec:preliminaries}.

\boldparagraph{Baselines} (1) \textbf{Roach} \cite{Zhang2021ICCV} is a popular RL planner trained with the PPO algorithm.
(2) \textbf{Think2Drive} \cite{Li2024ECCV} is a recent RL planner that uses the world model-based approach DreamerV3 \cite{Hafner2025NATURE}. We reproduce both Roach and Think2Drive based on the details provided in the respective  publications.
(3) \textbf{PDM-Lite} \cite{Sima2024ECCV} is a rule-based planning method in CARLA. The planners considered in this work are privileged, meaning they have access to ground truth perception inputs. These inputs could realistically be predicted by a perception stack, albeit at lower accuracy. PDM-Lite is special in that it extracts additional information about the scenarios from the CARLA leaderboard that is not predicted by any existing perception stack. PDM-Lite knows which scenario type will appear and what the scenario parameters are, and uses this information to solve some of the scenarios with specialized rules that are specific to the scenario. Such an approach works for the small variety of scenarios encountered in CARLA, but is unrealistic to scale to the long tail. PDM-Lite is therefore perhaps better viewed as an auto-labeling method for imitation planners. (4) \textbf{PlanT} \cite{Renz2022CORL} is the SotA imitation planner for the CARLA leaderboard 1.0. It uses bounding boxes as input, which are processed with a transformer \cite{Vaswani2017NIPS} and predicts waypoints. We reproduce PlanT for the CARLA leaderboard 2.0 by training it to imitate PDM-Lite.
Implementation details about the baselines can be found in Appendix \secref{sec:baselines}.

\boldparagraph{Results}
Our method, CaRL, achieves 64 DS on longest6 v2 as shown in \tabref{tab:longest6}. 
It significantly advances the SotA in RL planning, outperforming the best prior method, Roach, by 42 DS. 
It particularly reduces pedestrian (Ped) and vehicle (Veh) collisions compared to other RL baselines.
The recent world model-based Think2Drive method only achieves a DS of 7 and is even outperformed by the Roach method when both methods are trained with the hyperparameters proposed by the respective papers.
CaRL is the best learning based method outperforming PlanT by 2 DS while using a smaller model and less inference compute.
The rule-based method PDM-Lite still achieves the best DS with 73 in particular due to its high route completion and low collisions. 
PDM-Lite outperforms CaRL because it is more consistent at solving some of the safety-critical scenarios. 
PDM-Lite drives relatively slow on average, which may help it avoid collisions but as a result incurs many min-speed infractions (MS).
CaRL drives 31\% faster than PDM-Lite on average (16.4 km/h vs 12.5 km/h), which is indicated by its 4.5 times lower MS infraction.

We additionally report the average runtime per time step in milliseconds (ms), including preprocessing observations, on the last route of longest6 v2. Times are measured with an \texttt{RTX 3090} GPU and an \texttt{i9-10850K} CPU. We observe that the RL-based methods are more efficient than the rule-based and imitation-based methods. This is because the model size of RL policies is typically smaller than models used in imitation learning. CaRL runs 2 times faster than the state-of-the-art approach PDM-Lite.
It also uses a similar compute budget at inference as other RL-based approaches while driving substantially better.

\begin{table}[t]
\centering
    \begin{tabular}{l| c | l l | c c c | c }
        \toprule
        \textbf{Method} & \textbf{Type} & \textbf{DS} $\uparrow$ & \textbf{RC} $\uparrow$ & \textbf{Ped} $\downarrow$ & \textbf{Veh} $\downarrow$ & \textbf{MS} $\downarrow$ & \textbf{Time} $\downarrow$ \\
        \midrule
        PDM-Lite \cite{Sima2024ECCV} & Rule & \textbf{73} & \textbf{100} & \textbf{0.00} & \textbf{0.18} & \textbf{7.67} & \textbf{18} \\
        \specialrule{1pt}{0.25em}{0.25em} %
        PlanT \cite{Renz2022CORL} & IL & {62} \pmsd {2} & \textbf{96} \pmsd {2} & {0.07} & {0.43} & {3.18} & {18} \\
        Think2Drive \cite{Li2024ECCV} & RL & {7} \pmsd {1} & {39} \pmsd {19} & {0.97} & {2.55} & {18.05} & \textbf{6} \\
        Roach \cite{Zhang2021ICCV} & RL & {22} \pmsd {14} & {77} \pmsd {19} & {0.45} & {2.42} & {7.12} & {7} \\
        CaRL (Ours) & RL & \textbf{64} \pmsd {2} & {82} \pmsd {1} & \textbf{0.01} & \textbf{0.36} & \textbf{1.71} & 8 \\
        \bottomrule
    \end{tabular}
    \vspace{0.15cm}
    \caption{\textbf{Performance on longest6 v2} (CARLA).}
    \label{tab:longest6}
    \vspace{-0.0cm}
\end{table}

\subsection{nuPlan}
\label{sec:nuPlan}

\boldparagraph{Benchmark} NuPlan is a closed-loop simulator using real-world data.
As metric, we use the closed-loop score (CLS), which is a weighted average of progress, time-to-collision, speed-limit compliance, and comfort, scaled in 0-100.
Additionally, the score is set to zero if any hard penalty is violated \eg, collisions.
We evaluate all methods using non-reactive log replay (NR) and reactive background traffic (R), where other vehicles are controlled with the IDM~\cite{Treiber2000PRE}.
Following~\cite{Dauner2023CORL}, we use the Val14 benchmark that includes 1118 simulations from the validation split.

\boldparagraph{Baselines}(1)~\textbf{PDM-Closed}~\cite{Dauner2023CORL} is an extension of IDM \cite{Treiber2000PRE} with multiple trajectory proposals, internal simulation, and scoring. (2)~\textbf{PLUTO}~\cite{Cheng2024ARXIV} outputs multiple learned trajectories and scores, which can be combined with the scoring mechanism of PDM-Closed. 
(3)~\textbf{PlanTF}~\cite{Cheng2024ICRA} vectorizes agents and map elements and processes them with a transformer to forecast non-ego agents, and regress a trajectory. (4)~\textbf{Diffusion Planner}~\cite{Zheng2025ICLR} applies a diffusion transformer to generate the ego trajectory conditioned on a vectorized scene representation. (5)~\textbf{Log Replay} tracks the human trajectory with an LQR controller~\cite{Liu2021RCAR}.

\boldparagraph{Adaptation} We use different baselines for CARLA and nuPlan because the code of these planners is only compatible with either CARLA or nuPlan. 
Prior work only evaluated on one of the two simulators.
For a more rigorous evaluation, we reproduce CaRL on nuPlan as well.
This requires one special adaptation. 
The simulation duration in nuPlan is a constant 15 seconds, even when the ego agent has already completed the route. 
Since we primarily reward route completion, the agent would get no reward signal after completing the route.
To account for this simulator detail, we additionally reward the agent with a constant survival bonus at every frame during training, which gives the agent an incentive to avoid infractions after completing the route.
We train CaRL with 500M samples on nuPlan since the simulation is faster.
We refer the reader to Appendix \secref{sec:nuPlanChanges} for further implementation details.

\boldparagraph{Results} As shown in \tabref{tab:val14}, rule-based methods are highly effective in nuPlan, with PDM-Closed achieving over 92 reactive and non-reactive CLS. IL planners achieve reasonable performance (85-90) in non-reactive traffic but have difficulties adapting from the non-reactive setting (as seen during training) to the reactive IDM traffic. \Eg, Diffusion Planner (without post-processing) achieves a strong non-reactive CLS of 90 but a weaker reactive score of 83. PLUTO and PlanTF have similar performance drops. CaRL works well in both settings, achieving the highest CLS of all learned planners with 91.3 in the non-reactive and 90.6 CLS in reactive mode, respectively. CaRL runs $7-17\times$ faster than the baselines, due to its small 2M parameter network. We measure runtimes in ms on an \texttt{A5000} with an \texttt{i9-13900K}. The inference time of CaRL is slightly larger on nuPlan than on CARLA because nuPlan has more dynamic actors, which increases rendering times of the observation. We provide further results and baselines in Appendix~\ref{sec:nuPlan_result}.

\begin{table*}[t!]
\centering
    \resizebox{\textwidth}{!}{
    \begin{tabular}{l|c|ccc|ccc|c}
        \toprule
        \multirow{2}{*}{\textbf{Method}} & \multirow{2}{*}{\textbf{Type}} & \multicolumn{3}{c|}{\textbf{Non-Reactive}} & \multicolumn{3}{c|}{\textbf{Reactive}} & \multirow{2}{*}{\textbf{Time} $\downarrow$} \\
        & & \textbf{CLS} $\uparrow$ & \textbf{Col.} $\uparrow$ & \textbf{RC} $\uparrow$ & \textbf{CLS} $\uparrow$ & \textbf{Col.} $\uparrow$ & \textbf{RC} $\uparrow$ \\
        \midrule
        \textit{Log Replay (LQR)}  & \textit{Human} & \textit{93.5} & \textit{98.8} & \textit{99.0} & \textit{80.3} & \textit{85.6} & \textit{99.0} & - \\
        \midrule
        PDM-Closed \cite{Dauner2023CORL} & Rule & \textbf{92.8} & \textbf{98.1} & 92.1 & \textbf{92.1} & \textbf{97.9} & \textbf{90.3} & \textbf{104} \\
        PLUTO \cite{Cheng2024ARXIV} & IL+Rule & 92.6 & 97.9 & \textbf{93.0} & 89.7 & 97.1 & 86.1 & 237 \\ 
        \specialrule{1pt}{0.25em}{0.25em} 
         PlanTF \cite{Cheng2024ICRA} & IL & 84.6 & 94.2 & 90.7 & 76.1 & 95.2 & 77.2 & 107 \\
         Diff. Planner \cite{Zheng2025ICLR} & IL & 89.6 & 95.9 & 94.2 &  82.7 & 93.1 & 85.9 & 138 \\
        CaRL (Ours) & RL & \textbf{91.3} & \textbf{97.4} & \textbf{94.4} & \textbf{90.6} & \textbf{97.1} & \textbf{91.3} & \textbf{14} \\
        \bottomrule
    \end{tabular}}
    \caption{\textbf{Performance on Val14}  (nuPlan)}
    \label{tab:val14}
    \vspace{-0.0cm}
\end{table*}

\section{Conclusion}
\label{sec:conclusion}

Contemporary RL methods for driving often use complex rewards that induce tradeoffs between multiple reward terms. 
We observe that these tradeoffs prevent scalability. 
Training with large mini-batch sizes with PPO leads to degenerate policies, likely since the optimization gets stuck in a local minimum of the reward.
We propose an alternative reward design based on optimizing a single reward: route completion. Infractions either terminate the episode or multiplicatively reduce route completion.
We show that this enables learning with large mini-batch sizes, which in turn enables efficient scaling to more samples by parallelizing data collection.
We scale PPO to 300M samples in CARLA using a mini-batch size of 16384 and show that this leads to a performance improvement of 33 DS on the longest6 v2 benchmark.
CaRL achieves 91 CLS on nuPlan Val14 in both non-reactive and reactive traffic, outperforming all prior learning-based approaches.

\clearpage

\boldparagraph{Limitations}
We investigate urban driving at moderate speeds of up to 80 km/h.
Problems specific to high-speed driving on highways are not considered.

Our work improves the SOTA of RL for driving in simulation. 
For RL to become relevant for real cars, Sim2Real transfer \citep{Miki2022SR, Kaufmann2023Nature, Zeng2024CORL, Lin2025ARXIV} needs to be demonstrated, which we leave for future work. 

We train CaRL with the 7 scenario types of longest6 v2.
The CARLA leaderboard 2.0 offers more scenario types, which we have not investigated in this work.

For the CARLA vehicle physics, there are no human reference values for comfortable driving. 
Comfort is not a physical quantity but a subjective human feeling, so bounds need to be set based on human data. 
We set wide bounds to ensure that the model can comply with the comfort infraction, but the resulting behavior might not be comfortable in a real car. 
Future work may tune these bounds for smoother driving in CARLA.

CaRL has two main failure modes in CARLA: missing exits in highway off-ramps and other cars crashing into its rear in scenarios where another car runs a red light (rear-end collisions are counted as the agent's fault in CARLA). We show examples in the Appendix \secref{sec:failure_cases}.

Our reward does not encode that getting to the goal faster is better.
This is because most RL algorithms naturally encode a notion of urgency via a discount factor.

Both Longest6 v2 and Val14 are level 4 training benchmarks, where training on the evaluation town is allowed. 
We have not investigated level 5 generalization to new towns.

Our reward gives fewer local hints than other rewards. We have deliberately chosen an RL algorithm (PPO) that uses Monte-Carlo returns for optimization, which sum up rewards.
This might alleviate the absence of locality.
RL algorithms based on Q-learning \cite{Watkins1992ML, Mnih2015Nature}, that rely on local TD-prediction, such as Soft-Actor Critic \cite{Haarnoja2018ICML}, may have a harder time optimizing our reward, although we have not investigated other algorithms.

We think these limitations can be addressed and are promising directions for future work.

\acknowledgments{Bernhard Jaeger and Andreas Geiger were supported by the ERC Starting Grant LEGO-3D (850533), the DFG EXC number 2064/1 - project number 390727645, and the Vector Stiftung.
Daniel Dauner was supported by the German Federal Ministry for Economic Affairs and Climate Action within the project NXT GEN AI METHODS (19A23014S).
We thank the International Max Planck Research School for Intelligent Systems (IMPRS-IS) for supporting Bernhard Jaeger, Daniel Dauner, and Kashyap Chitta. We thank Katrin Renz for proofreading.}

{
    \small
    \bibliography{bibliography_long,bibliography,bibliography_custom}

\begin{thebibliography}{109}
\providecommand{\natexlab}[1]{#1}
\providecommand{\url}[1]{\texttt{#1}}
\expandafter\ifx\csname urlstyle\endcsname\relax
  \providecommand{\doi}[1]{doi: #1}\else
  \providecommand{\doi}{doi: \begingroup \urlstyle{rm}\Url}\fi

\bibitem[Treiber et~al.(2000)Treiber, Hennecke, and Helbing]{Treiber2000PRE}
M.~Treiber, A.~Hennecke, and D.~Helbing.
\newblock Congested traffic states in empirical observations and microscopic simulations.
\newblock \emph{Physical review E}, 2000.

\bibitem[Thrun et~al.(2006)Thrun, Montemerlo, Dahlkamp, Stavens, Aron, Diebel, Fong, Gale, Halpenny, Hoffmann, Lau, Oakley, Palatucci, Pratt, Stang, Strohband, Dupont, Jendrossek, Koelen, Markey, Rummel, van Niekerk, Jensen, Alessandrini, Bradski, Davies, Ettinger, Kaehler, Nefian, and Mahoney]{Thrun2006JFR}
S.~Thrun, M.~Montemerlo, H.~Dahlkamp, D.~Stavens, A.~Aron, J.~Diebel, P.~Fong, J.~Gale, M.~Halpenny, G.~Hoffmann, K.~Lau, C.~M. Oakley, M.~Palatucci, V.~R. Pratt, P.~Stang, S.~Strohband, C.~Dupont, L.~Jendrossek, C.~Koelen, C.~Markey, C.~Rummel, J.~van Niekerk, E.~Jensen, P.~Alessandrini, G.~R. Bradski, B.~Davies, S.~Ettinger, A.~Kaehler, A.~V. Nefian, and P.~Mahoney.
\newblock Stanley: The robot that won the {DARPA} grand challenge.
\newblock \emph{Journal of Field Robotics (JFR)}, 23\penalty0 (9):\penalty0 661--692, 2006.

\bibitem[Jaeger(2021)]{Jaeger2021}
B.~Jaeger.
\newblock Expert drivers for autonomous driving.
\newblock Master's thesis, University of Tübingen, 2021.

\bibitem[Dauner et~al.(2023)Dauner, Hallgarten, Geiger, and Chitta]{Dauner2023CORL}
D.~Dauner, M.~Hallgarten, A.~Geiger, and K.~Chitta.
\newblock Parting with misconceptions about learning-based vehicle motion planning.
\newblock In \emph{Proc. Conf. on Robot Learning (CoRL)}, 2023.

\bibitem[Sima et~al.(2024)Sima, Renz, Chitta, Chen, Zhang, Xie, Beißwenger, Luo, Geiger, and Li]{Sima2024ECCV}
C.~Sima, K.~Renz, K.~Chitta, L.~Chen, H.~Zhang, C.~Xie, J.~Beißwenger, P.~Luo, A.~Geiger, and H.~Li.
\newblock Drivelm: Driving with graph visual question answering.
\newblock In \emph{Proc. of the European Conf. on Computer Vision (ECCV)}, 2024.

\bibitem[Renz et~al.(2022)Renz, Chitta, Mercea, Koepke, Akata, and Geiger]{Renz2022CORL}
K.~Renz, K.~Chitta, O.-B. Mercea, A.~S. Koepke, Z.~Akata, and A.~Geiger.
\newblock Plant: Explainable planning transformers via object-level representations.
\newblock In \emph{Proc. Conf. on Robot Learning (CoRL)}, 2022.

\bibitem[Hallgarten et~al.(2023)Hallgarten, Stoll, and Zell]{Hallgarten2023ITSC}
M.~Hallgarten, M.~Stoll, and A.~Zell.
\newblock From prediction to planning with goal conditioned lane graph traversals.
\newblock In \emph{Proc. IEEE Conf. on Intelligent Transportation Systems (ITSC)}, 2023.

\bibitem[Cheng et~al.(2024)Cheng, Chen, Mei, Yang, Li, and Liu]{Cheng2024ICRA}
J.~Cheng, Y.~Chen, X.~Mei, B.~Yang, B.~Li, and M.~Liu.
\newblock Rethinking imitation-based planners for autonomous driving.
\newblock In \emph{Proc. IEEE International Conf. on Robotics and Automation (ICRA)}, 2024.

\bibitem[Huang et~al.(2024)Huang, Liu, Wu, and Lv]{Huang2024NNLS}
Z.~Huang, H.~Liu, J.~Wu, and C.~Lv.
\newblock Differentiable integrated motion prediction and planning with learnable cost function for autonomous driving.
\newblock \emph{{IEEE} Trans. Neural Networks Learn. Syst.}, 2024.

\bibitem[Sun et~al.(2024)Sun, Wang, Zhan, Nie, Wen, Xu, Zhan, Jia, Lang, and Zhao]{Sun2024ARXIV}
Q.~Sun, H.~Wang, J.~Zhan, F.~Nie, X.~Wen, L.~Xu, K.~Zhan, P.~Jia, X.~Lang, and H.~Zhao.
\newblock Generalizing motion planners with mixture of experts for autonomous driving.
\newblock \emph{arXiv.org}, 2410.15774, 2024.

\bibitem[Guo et~al.(2024)Guo, Feng, Zhu, Li, and Pu]{Guo2024ARXIV}
J.~Guo, M.~Feng, P.~Zhu, C.~Li, and J.~Pu.
\newblock Rethinking closed-loop planning framework for imitation-based model integrating prediction and planning.
\newblock \emph{arXiv.org}, 2407.05376, 2024.

\bibitem[Cheng et~al.(2024)Cheng, Chen, and Chen]{Cheng2024ARXIV}
J.~Cheng, Y.~Chen, and Q.~Chen.
\newblock {PLUTO:} pushing the limit of imitation learning-based planning for autonomous driving.
\newblock \emph{arXiv.org}, 2404.14327, 2024.

\bibitem[Chen et~al.(2025)Chen, Yan, Liao, He, and Peng]{Chen2025ARXIV}
X.~Chen, J.~Yan, W.~Liao, T.~He, and P.~Peng.
\newblock Int2planner: An intention-based multi-modal motion planner for integrated prediction and planning.
\newblock \emph{arXiv.org}, 2501.12799, 2025.

\bibitem[Zhang et~al.(2021)Zhang, Liniger, Dai, Yu, and Van~Gool]{Zhang2021ICCV}
Z.~Zhang, A.~Liniger, D.~Dai, F.~Yu, and L.~Van~Gool.
\newblock End-to-end urban driving by imitating a reinforcement learning coach.
\newblock In \emph{Proc. of the IEEE International Conf. on Computer Vision (ICCV)}, 2021.

\bibitem[Zhang et~al.(2022)Zhang, Guo, Zeng, Xiong, Dai, Hu, Ren, and Urtasun]{Zhang2022ECCV}
C.~Zhang, R.~Guo, W.~Zeng, Y.~Xiong, B.~Dai, R.~Hu, M.~Ren, and R.~Urtasun.
\newblock Rethinking closed-loop training for autonomous driving.
\newblock In \emph{Proc. of the European Conf. on Computer Vision (ECCV)}, 2022.

\bibitem[Sutton and Barto(2018)]{Sutton2018MIT}
R.~S. Sutton and A.~G. Barto.
\newblock \emph{Reinforcement learning: An introduction}.
\newblock MIT press, 2018.

\bibitem[Jaeger and Geiger(2024)]{Jaeger2024FTO}
B.~Jaeger and A.~Geiger.
\newblock An invitation to deep reinforcement learning.
\newblock \emph{Foundations and Trends in Optimization}, 2024.

\bibitem[Knox et~al.(2023)Knox, Allievi, Banzhaf, Schmitt, and Stone]{Knox2023AI}
W.~B. Knox, A.~Allievi, H.~Banzhaf, F.~Schmitt, and P.~Stone.
\newblock Reward (mis)design for autonomous driving.
\newblock \emph{Artificial Intelligence (AI)}, 2023.

\bibitem[Kendall et~al.(2019)Kendall, Hawke, Janz, Mazur, Reda, Allen, Lam, Bewley, and Shah]{Kendall2019ICRA}
A.~Kendall, J.~Hawke, D.~Janz, P.~Mazur, D.~Reda, J.~Allen, V.~Lam, A.~Bewley, and A.~Shah.
\newblock Learning to drive in a day.
\newblock In \emph{Proc. IEEE International Conf. on Robotics and Automation (ICRA)}, 2019.

\bibitem[Wijmans et~al.(2020)Wijmans, Kadian, Morcos, Lee, Essa, Parikh, Savva, and Batra]{Wijmans2020ICLR}
E.~Wijmans, A.~Kadian, A.~Morcos, S.~Lee, I.~Essa, D.~Parikh, M.~Savva, and D.~Batra.
\newblock {DD-PPO:} learning near-perfect pointgoal navigators from 2.5 billion frames.
\newblock In \emph{Proc. of the International Conf. on Learning Representations (ICLR)}, 2020.

\bibitem[Fuchs et~al.(2021)Fuchs, Song, Kaufmann, Scaramuzza, and D{\"{u}}rr]{Fuchs2021RAL}
F.~Fuchs, Y.~Song, E.~Kaufmann, D.~Scaramuzza, and P.~D{\"{u}}rr.
\newblock Super-human performance in gran turismo sport using deep reinforcement learning.
\newblock \emph{IEEE Robotics and Automation Letters (RA-L)}, 2021.

\bibitem[Zeng et~al.(2024)Zeng, Zhang, Ehsani, Hendrix, Salvador, Herrasti, Girshick, Kembhavi, and Weihs]{Zeng2024CORL}
K.~Zeng, Z.~Zhang, K.~Ehsani, R.~Hendrix, J.~Salvador, A.~Herrasti, R.~B. Girshick, A.~Kembhavi, and L.~Weihs.
\newblock Poliformer: Scaling on-policy {RL} with transformers results in masterful navigators.
\newblock In \emph{Proc. Conf. on Robot Learning (CoRL)}, volume 2406.20083, 2024.

\bibitem[Wurman et~al.(2022)Wurman, Barrett, Kawamoto, MacGlashan, Subramanian, Walsh, Capobianco, Devlic, Eckert, Fuchs, Gilpin, Khandelwal, Kompella, Lin, MacAlpine, Oller, Seno, Sherstan, Thomure, Aghabozorgi, Barrett, Douglas, Whitehead, D{\"{u}}rr, Stone, Spranger, and Kitano]{Wurman2022Nature}
P.~R. Wurman, S.~Barrett, K.~Kawamoto, J.~MacGlashan, K.~Subramanian, T.~J. Walsh, R.~Capobianco, A.~Devlic, F.~Eckert, F.~Fuchs, L.~Gilpin, P.~Khandelwal, V.~Kompella, H.~Lin, P.~MacAlpine, D.~Oller, T.~Seno, C.~Sherstan, M.~D. Thomure, H.~Aghabozorgi, L.~Barrett, R.~Douglas, D.~Whitehead, P.~D{\"{u}}rr, P.~Stone, M.~Spranger, and H.~Kitano.
\newblock Outracing champion gran turismo drivers with deep reinforcement learning.
\newblock \emph{Nature}, 2022.

\bibitem[Schulman et~al.(2017)Schulman, Wolski, Dhariwal, Radford, and Klimov]{Schulman2017ARXIV}
J.~Schulman, F.~Wolski, P.~Dhariwal, A.~Radford, and O.~Klimov.
\newblock Proximal policy optimization algorithms.
\newblock \emph{arXiv.org}, 1707.06347, 2017.

\bibitem[Toromanoff et~al.(2020)Toromanoff, Wirbel, and Moutarde]{Toromanoff2020CVPR}
M.~Toromanoff, E.~Wirbel, and F.~Moutarde.
\newblock End-to-end model-free reinforcement learning for urban driving using implicit affordances.
\newblock In \emph{Proc. IEEE Conf. on Computer Vision and Pattern Recognition (CVPR)}, 2020.

\bibitem[Chekroun et~al.(2023)Chekroun, Toromanoff, Hornauer, and Moutarde]{Chekroun2023ROBOTICS}
R.~Chekroun, M.~Toromanoff, S.~Hornauer, and F.~Moutarde.
\newblock {GRI:} general reinforced imitation and its application to vision-based autonomous driving.
\newblock \emph{Robotics}, 2023.

\bibitem[Li et~al.(2024)Li, Jia, Wang, and Yan]{Li2024ECCV}
Q.~Li, X.~Jia, S.~Wang, and J.~Yan.
\newblock Think2drive: Efficient reinforcement learning by thinking with latent world model for autonomous driving (in {CARLA-V2)}.
\newblock In \emph{Proc. of the European Conf. on Computer Vision (ECCV)}, 2024.

\bibitem[Dosovitskiy et~al.(2017)Dosovitskiy, Ros, Codevilla, Lopez, and Koltun]{Dosovitskiy2017CORL}
A.~Dosovitskiy, G.~Ros, F.~Codevilla, A.~Lopez, and V.~Koltun.
\newblock {CARLA}: {An} open urban driving simulator.
\newblock In \emph{Proc. Conf. on Robot Learning (CoRL)}, 2017.

\bibitem[team(2022)]{leaderboard22022ONLINE}
C.~team.
\newblock Carla autonomous driving leaderboard 2.0.
\newblock \url{https://leaderboard.carla.org/}, 2022.

\bibitem[Chitta et~al.(2023)Chitta, Prakash, Jaeger, Yu, Renz, and Geiger]{Chitta2023PAMI}
K.~Chitta, A.~Prakash, B.~Jaeger, Z.~Yu, K.~Renz, and A.~Geiger.
\newblock Transfuser: Imitation with transformer-based sensor fusion for autonomous driving.
\newblock \emph{Transactions on Pattern Analysis and Machine Intelligence (T-PAMI)}, 2023.

\bibitem[Karnchanachari et~al.(2024)Karnchanachari, Geromichalos, Tan, Li, Eriksen, Yaghoubi, Mehdipour, Bernasconi, Fong, Guo, and Caesar]{Karnchanachari2024ICRA}
N.~Karnchanachari, D.~Geromichalos, K.~S. Tan, N.~Li, C.~Eriksen, S.~Yaghoubi, N.~Mehdipour, G.~Bernasconi, W.~K. Fong, Y.~Guo, and H.~Caesar.
\newblock Towards learning-based planning: The nuplan benchmark for real-world autonomous driving.
\newblock In \emph{Proc. IEEE International Conf. on Robotics and Automation (ICRA)}, 2024.

\bibitem[Zheng et~al.(2025)Zheng, Liang, Zheng, Zheng, Mao, Li, Gu, Ai, Li, Zhan, et~al.]{Zheng2025ICLR}
Y.~Zheng, R.~Liang, K.~Zheng, J.~Zheng, L.~Mao, J.~Li, W.~Gu, R.~Ai, S.~E. Li, X.~Zhan, et~al.
\newblock Diffusion-based planning for autonomous driving with flexible guidance.
\newblock \emph{Proc. of the International Conf. on Learning Representations (ICLR)}, 2025.

\bibitem[CARLA(2022)]{Scenarios2022ONLINE}
CARLA.
\newblock Carla autonomous driving leaderboard 2.0 scenarios.
\newblock \url{https://leaderboard.carla.org/scenarios}, 2022.

\bibitem[CARLA(2024)]{CARLA2024Online}
CARLA.
\newblock Leaderboard 1.0 to 2.0 scenario converter.
\newblock \url{https://github.com/carla-simulator/leaderboard/blob/40d507ce67b189e66458a12492acaab706b28e92/scripts/route_bridge.py}, 2024.

\bibitem[team(2024)]{LB2Metrics2024Online}
C.~team.
\newblock Carla leaderboard 2.0 metrics.
\newblock \textsc{url:}~\url{https://leaderboard.carla.org/evaluation_v2_0}, 2024.

\bibitem[Prakash et~al.(2020)Prakash, Behl, Ohn-Bar, Chitta, and Geiger]{Prakash2020CVPR}
A.~Prakash, A.~Behl, E.~Ohn-Bar, K.~Chitta, and A.~Geiger.
\newblock Exploring data aggregation in policy learning for vision-based urban autonomous driving.
\newblock In \emph{Proc. IEEE Conf. on Computer Vision and Pattern Recognition (CVPR)}, 2020.

\bibitem[Behl et~al.(2020)Behl, Chitta, Prakash, Ohn-Bar, and Geiger]{Behl2020IROS}
A.~Behl, K.~Chitta, A.~Prakash, E.~Ohn-Bar, and A.~Geiger.
\newblock Label efficient visual abstractions for autonomous driving.
\newblock In \emph{Proc. IEEE International Conf. on Intelligent Robots and Systems (IROS)}, 2020.

\bibitem[Islam et~al.(2017)Islam, Henderson, Gomrokchi, and Precup]{Islam2017ARXIV}
R.~Islam, P.~Henderson, M.~Gomrokchi, and D.~Precup.
\newblock Reproducibility of benchmarked deep reinforcement learning tasks for continuous control.
\newblock \emph{arXiv.org}, 1708.04133, 2017.

\bibitem[Henderson et~al.(2018)Henderson, Islam, Bachman, Pineau, Precup, and Meger]{McIlraith2018AAAI}
P.~Henderson, R.~Islam, P.~Bachman, J.~Pineau, D.~Precup, and D.~Meger.
\newblock Deep reinforcement learning that matters.
\newblock In \emph{Proc. of the Conf. on Artificial Intelligence (AAAI)}, 2018.

\bibitem[Lynnerup et~al.(2019)Lynnerup, Nolling, Hasle, and Hallam]{Lynnerup2019CORL}
N.~A. Lynnerup, L.~Nolling, R.~Hasle, and J.~Hallam.
\newblock A survey on reproducibility by evaluating deep reinforcement learning algorithms on real-world robots.
\newblock In \emph{Proc. Conf. on Robot Learning (CoRL)}, 2019.

\bibitem[Agarwal et~al.(2021)Agarwal, Schwarzer, Castro, Courville, and Bellemare]{Agarwal2021NEURIPS}
R.~Agarwal, M.~Schwarzer, P.~S. Castro, A.~C. Courville, and M.~G. Bellemare.
\newblock Deep reinforcement learning at the edge of the statistical precipice.
\newblock In \emph{Advances in Neural Information Processing Systems (NeurIPS)}, 2021.

\bibitem[Fetterman et~al.(2023)Fetterman, Kitanidis, Albrecht, Polizzi, Fogelman, Knutins, Wr{\'{o}}blewski, Simon, and Qiu]{Fetterman2023ARXIV}
A.~J. Fetterman, E.~Kitanidis, J.~Albrecht, Z.~Polizzi, B.~Fogelman, M.~Knutins, B.~Wr{\'{o}}blewski, J.~B. Simon, and K.~Qiu.
\newblock Tune as you scale: Hyperparameter optimization for compute efficient training.
\newblock \emph{arXiv.org}, 2306.08055, 2023.

\bibitem[Huang et~al.(2023)Huang, Dossa, Raffin, Kanervisto, and Wang]{Huang2023ICLRBLOG}
S.~Huang, R.~F.~J. Dossa, A.~Raffin, A.~Kanervisto, and W.~Wang.
\newblock The 37 implementation details of proximal policy optimization.
\newblock \emph{The ICLR Blog Track 2023}, 2023.

\bibitem[Williams(1992)]{Williams1992ML}
R.~J. Williams.
\newblock Simple statistical gradient-following algorithms for connectionist reinforcement learning.
\newblock \emph{Machine Learning}, 8:\penalty0 229--256, 1992.

\bibitem[Mnih et~al.(2016)Mnih, Badia, Mirza, Graves, Lillicrap, Harley, Silver, and Kavukcuoglu]{Mnih2016ICML}
V.~Mnih, A.~P. Badia, M.~Mirza, A.~Graves, T.~P. Lillicrap, T.~Harley, D.~Silver, and K.~Kavukcuoglu.
\newblock Asynchronous methods for deep reinforcement learning.
\newblock In \emph{Proc. of the International Conf. on Machine learning (ICML)}, 2016.

\bibitem[Degris et~al.(2012)Degris, White, and Sutton]{Degris2012ARXIV}
T.~Degris, M.~White, and R.~S. Sutton.
\newblock Off-policy actor-critic.
\newblock \emph{arXiv.org}, 1205.4839, 2012.

\bibitem[Zhang et~al.(2025)Zhang, Liang, Guo, Lu, Wang, Xiong, Miao, and Wang]{Zhang2025ArXiV}
D.~Zhang, J.~Liang, K.~Guo, S.~Lu, Q.~Wang, R.~Xiong, Z.~Miao, and Y.~Wang.
\newblock Carplanner: Consistent auto-regressive trajectory planning for large-scale reinforcement learning in autonomous driving.
\newblock \emph{arXiv.org}, 2502.19908, 2025.

\bibitem[Koltun(2020)]{Koltun2020ICMLWorkshop}
V.~Koltun.
\newblock Autonomous driving: The way forward.
\newblock \textsc{url:}~\url{https://www.youtube.com/watch?v=XmtTjqimW3g}, 2020.

\bibitem[Hillis and Jr.(1986)]{Hillis1986ACM}
W.~D. Hillis and G.~L.~S. Jr.
\newblock Data parallel algorithms.
\newblock \emph{Communications of the ACM}, 1986.

\bibitem[Kaufmann et~al.(2023)Kaufmann, Bauersfeld, Loquercio, Müller, Koltun, and Scaramuzza]{Kaufmann2023Nature}
E.~Kaufmann, L.~Bauersfeld, A.~Loquercio, M.~Müller, V.~Koltun, and D.~Scaramuzza.
\newblock Champion-level drone racing using deep reinforcement learning.
\newblock \emph{Nature}, 2023.

\bibitem[Jaeger et~al.(2024)Jaeger, Chitta, Dauner, Renz, and Geiger]{Jaeger2024Online}
B.~Jaeger, K.~Chitta, D.~Dauner, K.~Renz, and A.~Geiger.
\newblock Common mistakes in benchmarking autonomous driving.
\newblock \textsc{url:}~\url{https://github.com/autonomousvision/carla_garage/blob/leaderboard_2/docs/common_mistakes_in_benchmarking_ad.md}, 2024.

\bibitem[Hafner et~al.(2025)Hafner, Pasukonis, Ba, and Lillicrap]{Hafner2025NATURE}
D.~Hafner, J.~Pasukonis, J.~Ba, and T.~Lillicrap.
\newblock Mastering diverse control tasks through world models.
\newblock \emph{Nature}, 2025.

\bibitem[Vaswani et~al.(2017)Vaswani, Shazeer, Parmar, Uszkoreit, Jones, Gomez, Kaiser, and Polosukhin]{Vaswani2017NIPS}
A.~Vaswani, N.~Shazeer, N.~Parmar, J.~Uszkoreit, L.~Jones, A.~N. Gomez, L.~Kaiser, and I.~Polosukhin.
\newblock Attention is all you need.
\newblock In \emph{Advances in Neural Information Processing Systems (NIPS)}, pages 5998--6008, 2017.

\bibitem[Liu et~al.(2021)Liu, Yang, Huang, Li, Dang, and Li]{Liu2021RCAR}
J.~Liu, Z.~Yang, Z.~Huang, W.~Li, S.~Dang, and H.~Li.
\newblock Simulation performance evaluation of pure pursuit, stanley, lqr, mpc controller for autonomous vehicles.
\newblock In \emph{IEEE international conference on real-time computing and robotics (RCAR)}. IEEE, 2021.

\bibitem[Miki et~al.(2022)Miki, Lee, Hwangbo, Wellhausen, Koltun, and Hutter]{Miki2022SR}
T.~Miki, J.~Lee, J.~Hwangbo, L.~Wellhausen, V.~Koltun, and M.~Hutter.
\newblock Learning robust perceptive locomotion for quadrupedal robots in the wild.
\newblock \emph{Science Robotics}, 2022.

\bibitem[Lin et~al.(2025)Lin, Sachdev, Fan, Malik, and Zhu]{Lin2025ARXIV}
T.~Lin, K.~Sachdev, L.~Fan, J.~Malik, and Y.~Zhu.
\newblock Sim-to-real reinforcement learning for vision-based dexterous manipulation on humanoids.
\newblock \emph{arXiv.org}, 2502.20396, 2025.

\bibitem[Watkins and Dayan(1992)]{Watkins1992ML}
C.~J. Watkins and P.~Dayan.
\newblock Q-learning.
\newblock \emph{Machine Learning}, 1992.

\bibitem[Mnih et~al.(2015)Mnih, Kavukcuoglu, Silver, Rusu, Veness, Bellemare, Graves, Riedmiller, Fidjeland, Ostrovski, Petersen, Beattie, Sadik, Antonoglou, King, Kumaran, Wierstra, Legg, and Hassabis]{Mnih2015Nature}
V.~Mnih, K.~Kavukcuoglu, D.~Silver, A.~A. Rusu, J.~Veness, M.~G. Bellemare, A.~Graves, M.~A. Riedmiller, A.~Fidjeland, G.~Ostrovski, S.~Petersen, C.~Beattie, A.~Sadik, I.~Antonoglou, H.~King, D.~Kumaran, D.~Wierstra, S.~Legg, and D.~Hassabis.
\newblock Human-level control through deep reinforcement learning.
\newblock \emph{Nature}, 2015.

\bibitem[Haarnoja et~al.(2018)Haarnoja, Zhou, Abbeel, and Levine]{Haarnoja2018ICML}
T.~Haarnoja, A.~Zhou, P.~Abbeel, and S.~Levine.
\newblock Soft actor-critic: Off-policy maximum entropy deep reinforcement learning with a stochastic actor.
\newblock In \emph{Proc. of the International Conf. on Machine learning (ICML)}, 2018.

\bibitem[Rajamani(2011)]{Rajamani2011BOOK}
R.~Rajamani.
\newblock \emph{Vehicle dynamics and control}.
\newblock Springer Science \& Business Media, 2011.

\bibitem[Polack et~al.(2017)Polack, Altché, d'Andréa Novel, and de~La~Fortelle]{Polack2017IV}
P.~Polack, F.~Altché, B.~d'Andréa Novel, and A.~de~La~Fortelle.
\newblock The kinematic bicycle model: A consistent model for planning feasible trajectories for autonomous vehicles?
\newblock In \emph{Proc. IEEE Intelligent Vehicles Symposium (IV)}, 2017.

\bibitem[Chekroun et~al.(2024)Chekroun, Gilles, Toromanoff, Hornauer, and Moutarde]{Chekroun2024IV}
R.~Chekroun, T.~Gilles, M.~Toromanoff, S.~Hornauer, and F.~Moutarde.
\newblock {MBAPPE:} mcts-built-around prediction for planning explicitly.
\newblock In \emph{Proc. IEEE Intelligent Vehicles Symposium (IV)}, 2024.

\bibitem[Yang et~al.(2024)Yang, Su, Gkanatsios, Ke, Jain, Schneider, and Fragkiadaki]{Yang2024CVPR}
B.~Yang, H.~Su, N.~Gkanatsios, T.~Ke, A.~Jain, J.~G. Schneider, and K.~Fragkiadaki.
\newblock Diffusion-es: Gradient-free planning with diffusion for autonomous and instruction-guided driving.
\newblock In \emph{Proc. IEEE Conf. on Computer Vision and Pattern Recognition (CVPR)}, 2024.

\bibitem[Gulino et~al.(2023)Gulino, Fu, Luo, Tucker, Bronstein, Lu, Harb, Pan, Wang, Chen, Co{-}Reyes, Agarwal, Roelofs, Lu, Montali, Mougin, Yang, White, Faust, McAllister, Anguelov, and Sapp]{Gulino2023NEURIPS}
C.~Gulino, J.~Fu, W.~Luo, G.~Tucker, E.~Bronstein, Y.~Lu, J.~Harb, X.~Pan, Y.~Wang, X.~Chen, J.~D. Co{-}Reyes, R.~Agarwal, R.~Roelofs, Y.~Lu, N.~Montali, P.~Mougin, Z.~Yang, B.~White, A.~Faust, R.~McAllister, D.~Anguelov, and B.~Sapp.
\newblock Waymax: An accelerated, data-driven simulator for large-scale autonomous driving research.
\newblock In \emph{Advances in Neural Information Processing Systems (NeurIPS)}, 2023.

\bibitem[Sun et~al.(2020)Sun, Kretzschmar, Dotiwalla, Chouard, Patnaik, Tsui, Guo, Zhou, Chai, Caine, Vasudevan, Han, Ngiam, Zhao, Timofeev, Ettinger, Krivokon, Gao, Joshi, Zhang, Shlens, Chen, and Anguelov]{Sun2020CVPR}
P.~Sun, H.~Kretzschmar, X.~Dotiwalla, A.~Chouard, V.~Patnaik, P.~Tsui, J.~Guo, Y.~Zhou, Y.~Chai, B.~Caine, V.~Vasudevan, W.~Han, J.~Ngiam, H.~Zhao, A.~Timofeev, S.~Ettinger, M.~Krivokon, A.~Gao, A.~Joshi, Y.~Zhang, J.~Shlens, Z.~Chen, and D.~Anguelov.
\newblock Scalability in perception for autonomous driving: Waymo open dataset.
\newblock In \emph{Proc. IEEE Conf. on Computer Vision and Pattern Recognition (CVPR)}, 2020.

\bibitem[Xiao et~al.(2024)Xiao, Liu, Ye, Yang, and Wang]{Xiao2024ArXiV}
L.~Xiao, J.~Liu, X.~Ye, W.~Yang, and J.~Wang.
\newblock Easychauffeur: {A} baseline advancing simplicity and efficiency on waymax.
\newblock \emph{arXiv.org}, 2408.16375, 2024.

\bibitem[Charraut et~al.(2025)Charraut, Tournaire, Doulazmi, and Buhet]{Charraut2025ArXiV}
V.~Charraut, T.~Tournaire, W.~Doulazmi, and T.~Buhet.
\newblock V-max: Making rl practical for autonomous driving.
\newblock \emph{arXiv.org}, 2503.08388, 2025.

\bibitem[Cusumano-Towner et~al.(2025)Cusumano-Towner, Hafner, Hertzberg, Huval, Petrenko, Vinitsky, Wijmans, Killian, Bowers, Sener, Krähenbühl, and Koltun]{Cusumano-Towner2025ARXIV}
M.~Cusumano-Towner, D.~Hafner, A.~Hertzberg, B.~Huval, A.~Petrenko, E.~Vinitsky, E.~Wijmans, T.~Killian, S.~Bowers, O.~Sener, P.~Krähenbühl, and V.~Koltun.
\newblock Robust autonomy emerges from self-play.
\newblock \emph{arXiv.org}, 2502.03349, 2025.

\bibitem[Ross et~al.(2011)Ross, Gordon, and Bagnell]{Ross2011AISTATS}
S.~Ross, G.~J. Gordon, and D.~Bagnell.
\newblock A reduction of imitation learning and structured prediction to no-regret online learning.
\newblock In \emph{Conference on Artificial Intelligence and Statistics (AISTATS)}, 2011.

\bibitem[Chen et~al.(2024)Chen, Wu, Chitta, Jaeger, Geiger, and Li]{Chen2024PAMI}
L.~Chen, P.~Wu, K.~Chitta, B.~Jaeger, A.~Geiger, and H.~Li.
\newblock End-to-end autonomous driving: Challenges and frontiers.
\newblock \emph{Transactions on Pattern Analysis and Machine Intelligence (T-PAMI)}, 2024.

\bibitem[Lu et~al.(2023)Lu, Fu, Tucker, Pan, Bronstein, Roelofs, Sapp, White, Faust, Whiteson, Anguelov, and Levine]{Lu2023IROS}
Y.~Lu, J.~Fu, G.~Tucker, X.~Pan, E.~Bronstein, R.~Roelofs, B.~Sapp, B.~White, A.~Faust, S.~Whiteson, D.~Anguelov, and S.~Levine.
\newblock Imitation is not enough: Robustifying imitation with reinforcement learning for challenging driving scenarios.
\newblock In \emph{Proc. IEEE International Conf. on Intelligent Robots and Systems (IROS)}, 2023.

\bibitem[Liang et~al.(2018)Liang, Wang, Yang, and Xing]{Liang2018ECCVb}
X.~Liang, T.~Wang, L.~Yang, and E.~P. Xing.
\newblock {CIRL:} controllable imitative reinforcement learning for vision-based self-driving.
\newblock In \emph{Proc. of the European Conf. on Computer Vision (ECCV)}, 2018.

\bibitem[Song et~al.(2021)Song, Lin, Kaufmann, D{\"{u}}rr, and Scaramuzza]{Song2021ICRA}
Y.~Song, H.~Lin, E.~Kaufmann, P.~D{\"{u}}rr, and D.~Scaramuzza.
\newblock Autonomous overtaking in gran turismo sport using curriculum reinforcement learning.
\newblock In \emph{Proc. IEEE International Conf. on Robotics and Automation (ICRA)}, 2021.

\bibitem[Zhang et~al.(2021)Zhang, Liniger, Dai, Yu, and Gool]{Zhang2021ARXIVb}
Z.~Zhang, A.~Liniger, D.~Dai, F.~Yu, and L.~V. Gool.
\newblock End-to-end urban driving by imitating a reinforcement learning coach.
\newblock \emph{arXiv.org}, 2108.08265v3, 2021.

\bibitem[Wu et~al.(2022)Wu, Jia, Chen, Yan, Li, and Qiao]{Wu2022NEURIPS}
P.~Wu, X.~Jia, L.~Chen, J.~Yan, H.~Li, and Y.~Qiao.
\newblock Trajectory-guided control prediction for end-to-end autonomous driving: {A} simple yet strong baseline.
\newblock In \emph{Advances in Neural Information Processing Systems (NeurIPS)}, 2022.

\bibitem[Jia et~al.(2023{\natexlab{a}})Jia, Wu, Chen, Xie, He, Yan, and Li]{Jia2023CVPR}
X.~Jia, P.~Wu, L.~Chen, J.~Xie, C.~He, J.~Yan, and H.~Li.
\newblock Think twice before driving: Towards scalable decoders for end-to-end autonomous driving.
\newblock In \emph{Proc. IEEE Conf. on Computer Vision and Pattern Recognition (CVPR)}, 2023{\natexlab{a}}.

\bibitem[Jia et~al.(2023{\natexlab{b}})Jia, Gao, Chen, Yan, Liu, and Li]{Jia2023ICCV}
X.~Jia, Y.~Gao, L.~Chen, J.~Yan, P.~L. Liu, and H.~Li.
\newblock Driveadapter: Breaking the coupling barrier of perception and planning in end-to-end autonomous driving.
\newblock In \emph{Proc. of the IEEE International Conf. on Computer Vision (ICCV)}, 2023{\natexlab{b}}.

\bibitem[Ishida et~al.(2024)Ishida, Corrado, Fedoseev, Yeo, Russell, Shotton, Henriques, and Hu]{Ishida2024ICLRW}
S.~Ishida, G.~Corrado, G.~Fedoseev, H.~Yeo, L.~Russell, J.~Shotton, J.~F. Henriques, and A.~Hu.
\newblock Langprop: A code optimization framework using large language models applied to driving.
\newblock In \emph{ICLR 2024 Workshop on Large Language Model (LLM) Agents}, 2024.

\bibitem[Huang et~al.(2022)Huang, Dossa, Ye, Braga, Chakraborty, Mehta, and Ara{\'{u}}jo]{Huang2022JMLR}
S.~Huang, R.~F.~J. Dossa, C.~Ye, J.~Braga, D.~Chakraborty, K.~Mehta, and J.~G.~M. Ara{\'{u}}jo.
\newblock Cleanrl: High-quality single-file implementations of deep reinforcement learning algorithms.
\newblock \emph{Journal of Machine Learning Research (JMLR)}, 2022.

\bibitem[Ha and Schmidhuber(2018)]{Ha2018NEURIPS}
D.~Ha and J.~Schmidhuber.
\newblock Recurrent world models facilitate policy evolution.
\newblock In \emph{Advances in Neural Information Processing Systems (NeurIPS)}, 2018.

\bibitem[Hafner et~al.(2020)Hafner, Lillicrap, Ba, and Norouzi]{Hafner2020ICLR}
D.~Hafner, T.~P. Lillicrap, J.~Ba, and M.~Norouzi.
\newblock Dream to control: Learning behaviors by latent imagination.
\newblock In \emph{Proc. of the International Conf. on Learning Representations (ICLR)}, 2020.

\bibitem[Hafner et~al.(2021)Hafner, Lillicrap, Norouzi, and Ba]{Hafner2021ICLR}
D.~Hafner, T.~P. Lillicrap, M.~Norouzi, and J.~Ba.
\newblock Mastering atari with discrete world models.
\newblock In \emph{Proc. of the International Conf. on Learning Representations (ICLR)}, 2021.

\bibitem[Kazemkhani et~al.(2025)Kazemkhani, Pandya, Cornelisse, Shacklett, and Vinitsky]{Kazemkhani2025ICLR}
S.~Kazemkhani, A.~Pandya, D.~Cornelisse, B.~Shacklett, and E.~Vinitsky.
\newblock Gpudrive: Data-driven, multi-agent driving simulation at 1 million {FPS}.
\newblock In \emph{Proc. of the International Conf. on Learning Representations (ICLR)}, 2025.

\bibitem[Chen and Kr{\"a}henb{\"u}hl(2022)]{Chen2022CVPRa}
D.~Chen and P.~Kr{\"a}henb{\"u}hl.
\newblock Learning from all vehicles.
\newblock In \emph{Proc. IEEE Conf. on Computer Vision and Pattern Recognition (CVPR)}, 2022.

\bibitem[Weng et~al.(2022)Weng, Lin, Huang, Liu, Makoviichuk, Makoviychuk, Liu, Song, Luo, Jiang, Xu, and Yan]{Weng2022NEURIPS}
J.~Weng, M.~Lin, S.~Huang, B.~Liu, D.~Makoviichuk, V.~Makoviychuk, Z.~Liu, Y.~Song, T.~Luo, Y.~Jiang, Z.~Xu, and S.~Yan.
\newblock Envpool: {A} highly parallel reinforcement learning environment execution engine.
\newblock In \emph{Advances in Neural Information Processing Systems (NeurIPS)}, 2022.

\bibitem[Vinyals et~al.(2019)Vinyals, Babuschkin, Czarnecki, Mathieu, Dudzik, Chung, Choi, Powell, Ewalds, Georgiev, Oh, Horgan, Kroiss, Danihelka, Huang, Sifre, Cai, Agapiou, Jaderberg, Vezhnevets, Leblond, Pohlen, Dalibard, Budden, Sulsky, Molloy, Paine, G{\"{u}}l{\c{c}}ehre, Wang, Pfaff, Wu, Ring, Yogatama, W{\"{u}}nsch, McKinney, Smith, Schaul, Lillicrap, Kavukcuoglu, Hassabis, Apps, and Silver]{Vinyals2019Nature}
O.~Vinyals, I.~Babuschkin, W.~M. Czarnecki, M.~Mathieu, A.~Dudzik, J.~Chung, D.~H. Choi, R.~Powell, T.~Ewalds, P.~Georgiev, J.~Oh, D.~Horgan, M.~Kroiss, I.~Danihelka, A.~Huang, L.~Sifre, T.~Cai, J.~P. Agapiou, M.~Jaderberg, A.~S. Vezhnevets, R.~Leblond, T.~Pohlen, V.~Dalibard, D.~Budden, Y.~Sulsky, J.~Molloy, T.~L. Paine, {\c{C}}.~G{\"{u}}l{\c{c}}ehre, Z.~Wang, T.~Pfaff, Y.~Wu, R.~Ring, D.~Yogatama, D.~W{\"{u}}nsch, K.~McKinney, O.~Smith, T.~Schaul, T.~P. Lillicrap, K.~Kavukcuoglu, D.~Hassabis, C.~Apps, and D.~Silver.
\newblock Grandmaster level in starcraft {II} using multi-agent reinforcement learning.
\newblock \emph{Nature}, 2019.

\bibitem[Berner et~al.(2019)Berner, Brockman, Chan, Cheung, Debiak, Dennison, Farhi, Fischer, Hashme, Hesse, J{\'{o}}zefowicz, Gray, Olsson, Pachocki, Petrov, de~Oliveira~Pinto, Raiman, Salimans, Schlatter, Schneider, Sidor, Sutskever, Tang, Wolski, and Zhang]{Berner2019ARXIV}
C.~Berner, G.~Brockman, B.~Chan, V.~Cheung, P.~Debiak, C.~Dennison, D.~Farhi, Q.~Fischer, S.~Hashme, C.~Hesse, R.~J{\'{o}}zefowicz, S.~Gray, C.~Olsson, J.~Pachocki, M.~Petrov, H.~P. de~Oliveira~Pinto, J.~Raiman, T.~Salimans, J.~Schlatter, J.~Schneider, S.~Sidor, I.~Sutskever, J.~Tang, F.~Wolski, and S.~Zhang.
\newblock Dota 2 with large scale deep reinforcement learning.
\newblock \emph{arXiv.org}, 1912.06680, 2019.

\bibitem[Petrenko et~al.(2020)Petrenko, Huang, Kumar, Sukhatme, and Koltun]{Petrenko2020ICML}
A.~Petrenko, Z.~Huang, T.~Kumar, G.~S. Sukhatme, and V.~Koltun.
\newblock Sample factory: Egocentric 3d control from pixels at 100000 {FPS} with asynchronous reinforcement learning.
\newblock In \emph{Proc. of the International Conf. on Machine learning (ICML)}, 2020.

\bibitem[Python(2025)]{Python2025Online}
Python.
\newblock Pep 703 – making the global interpreter lock optional in cpython.
\newblock \textsc{url:}~\url{https://peps.python.org/pep-0703/}, 2025.

\bibitem[Paszke et~al.(2019)Paszke, Gross, Massa, Lerer, Bradbury, Chanan, Killeen, Lin, Gimelshein, Antiga, Desmaison, K{\"{o}}pf, Yang, DeVito, Raison, Tejani, Chilamkurthy, Steiner, Fang, Bai, and Chintala]{Paszke2019NeurIPS}
A.~Paszke, S.~Gross, F.~Massa, A.~Lerer, J.~Bradbury, G.~Chanan, T.~Killeen, Z.~Lin, N.~Gimelshein, L.~Antiga, A.~Desmaison, A.~K{\"{o}}pf, E.~Z. Yang, Z.~DeVito, M.~Raison, A.~Tejani, S.~Chilamkurthy, B.~Steiner, L.~Fang, J.~Bai, and S.~Chintala.
\newblock Pytorch: An imperative style, high-performance deep learning library.
\newblock In \emph{Advances in Neural Information Processing Systems (NeurIPS)}, 2019.

\bibitem[Lyle et~al.(2023)Lyle, Zheng, Nikishin, Pires, Pascanu, and Dabney]{Lyle2023ICML}
C.~Lyle, Z.~Zheng, E.~Nikishin, B.~{\'{A}}. Pires, R.~Pascanu, and W.~Dabney.
\newblock Understanding plasticity in neural networks.
\newblock In \emph{Proc. of the International Conf. on Machine learning (ICML)}, Proceedings of Machine Learning Research, 2023.

\bibitem[Dohare et~al.(2024)Dohare, Hernandez{-}Garcia, Lan, Rahman, Mahmood, and Sutton]{Dohare2024NATURE}
S.~Dohare, J.~F. Hernandez{-}Garcia, Q.~Lan, P.~Rahman, A.~R. Mahmood, and R.~S. Sutton.
\newblock Loss of plasticity in deep continual learning.
\newblock \emph{Nature}, 2024.

\bibitem[Juliani and Ash(2024)]{Juliani2024NeurIPS}
A.~Juliani and J.~Ash.
\newblock A study of plasticity loss in on-policy deep reinforcement learning.
\newblock In \emph{Advances in Neural Information Processing Systems (NeurIPS)}, 2024.

\bibitem[Moalla et~al.(2024)Moalla, Miele, Pyatko, Pascanu, and Gulcehre]{Moalla2024NEURIPS}
S.~Moalla, A.~Miele, D.~Pyatko, R.~Pascanu, and C.~Gulcehre.
\newblock No representation, no trust: Connecting representation, collapse, and trust issues in {PPO}.
\newblock In \emph{Advances in Neural Information Processing Systems (NeurIPS)}, 2024.

\bibitem[Chou et~al.(2017)Chou, Maturana, and Scherer]{Chou2017ICML}
P.~Chou, D.~Maturana, and S.~A. Scherer.
\newblock Improving stochastic policy gradients in continuous control with deep reinforcement learning using the beta distribution.
\newblock In \emph{Proc. of the International Conf. on Machine learning (ICML)}, 2017.

\bibitem[Petrazzini and Antonelo(2021)]{Petrazzini2021SSCI}
I.~G.~B. Petrazzini and E.~A. Antonelo.
\newblock Proximal policy optimization with continuous bounded action space via the beta distribution.
\newblock In \emph{IEEE Symposium Series on Computational Intelligence, (SSCI)}, 2021.

\bibitem[Jaeger et~al.(2023)Jaeger, Chitta, and Geiger]{Jaeger2023ICCV}
B.~Jaeger, K.~Chitta, and A.~Geiger.
\newblock Hidden biases of end-to-end driving models.
\newblock In \emph{Proc. of the IEEE International Conf. on Computer Vision (ICCV)}, 2023.

\bibitem[Brockman et~al.(2016)Brockman, Cheung, Pettersson, Schneider, Schulman, Tang, and Zaremba]{Brockman2016ARXIV}
G.~Brockman, V.~Cheung, L.~Pettersson, J.~Schneider, J.~Schulman, J.~Tang, and W.~Zaremba.
\newblock Openai gym.
\newblock \emph{arXiv.org}, 1606.01540, 2016.

\bibitem[Towers et~al.(2024)Towers, Kwiatkowski, Terry, Balis, Cola, Deleu, Goul{\~{a}}o, Kallinteris, Krimmel, KG, Perez{-}Vicente, Pierr{\'{e}}, Schulhoff, Tai, Tan, and Younis]{Towers2024ArXiv}
M.~Towers, A.~Kwiatkowski, J.~K. Terry, J.~U. Balis, G.~D. Cola, T.~Deleu, M.~Goul{\~{a}}o, A.~Kallinteris, M.~Krimmel, A.~KG, R.~Perez{-}Vicente, A.~Pierr{\'{e}}, S.~Schulhoff, J.~J. Tai, H.~Tan, and O.~G. Younis.
\newblock Gymnasium: {A} standard interface for reinforcement learning environments.
\newblock \emph{arXiv.org}, 2407.17032, 2024.

\bibitem[Hintjens(2013)]{Hintjens2013BOOK}
P.~Hintjens.
\newblock \emph{ZeroMQ: Messaging for Many Applications}.
\newblock O'Reilly Media, 2013.

\bibitem[Huang et~al.(2022)Huang, Dossa, Raffin, Kanervisto, and Wang]{Shengyi2022Online}
S.~Huang, R.~F.~J. Dossa, A.~Raffin, A.~Kanervisto, and W.~Wang.
\newblock The 37 implementation details of proximal policy optimization.
\newblock In \emph{ICLR Blog Track}, 2022.
\newblock URL \url{https://iclr-blog-track.github.io/2022/03/25/ppo-implementation-details/}.

\bibitem[Hanselmann et~al.(2022)Hanselmann, Renz, Chitta, Bhattacharyya, and Geiger]{Hanselmann2022ECCV}
N.~Hanselmann, K.~Renz, K.~Chitta, A.~Bhattacharyya, and A.~Geiger.
\newblock King: Generating safety-critical driving scenarios for robust imitation via kinematics gradients.
\newblock In \emph{Proc. of the European Conf. on Computer Vision (ECCV)}, 2022.

\bibitem[Hao et~al.(2023)Hao, Cui, Luo, Xie, Bai, Yang, Yan, Pan, and Yang]{Hao2023TITS}
K.~Hao, W.~Cui, Y.~Luo, L.~Xie, Y.~Bai, J.~Yang, S.~Yan, Y.~Pan, and Z.~Yang.
\newblock Adversarial safety-critical scenario generation using naturalistic human driving priors.
\newblock \emph{IEEE Trans. on Intelligent Transportation Systems (TITS)}, 2023.

\bibitem[Yin et~al.(2024)Yin, Khayatan, Éloi Zablocki, Boulch, , and Cord]{Yin2024ECCVW}
Y.~Yin, P.~Khayatan, Éloi Zablocki, A.~Boulch, , and M.~Cord.
\newblock Regents: Real-world safety-critical driving scenario generation made stable.
\newblock In \emph{ECCV 2024 W-CODA Workshop}, 2024.

\bibitem[Hipp(2000)]{Hipp2000Online}
D.~R. Hipp.
\newblock Sqlite.
\newblock \textsc{url:}~\url{http://sqlite. org/}, 2000.

\bibitem[den Bossche et~al.(2024)den Bossche, Jordahl, Fleischmann, Richards, McBride, Wasserman, Badaracco, Snow, Ward, Tratner, Gerard, Perry, cjqf, Hjelle, Taves, ter Hoeven, Cochran, Bell, rraymondgh, Bartos, Roggemans, Culbertson, Caria, Tan, Eubank, sangarshanan, Flavin, Rey, and Gardiner]{Bossche2024Online}
J.~V. den Bossche, K.~Jordahl, M.~Fleischmann, M.~Richards, J.~McBride, J.~Wasserman, A.~G. Badaracco, A.~D. Snow, B.~Ward, J.~Tratner, J.~Gerard, M.~Perry, cjqf, G.~A. Hjelle, M.~Taves, E.~ter Hoeven, M.~Cochran, R.~Bell, rraymondgh, M.~Bartos, P.~Roggemans, L.~Culbertson, G.~Caria, N.~Y. Tan, N.~Eubank, sangarshanan, J.~Flavin, S.~Rey, and J.~Gardiner.
\newblock geopandas/geopandas: v1.0.1, 2024.
\newblock URL \url{https://doi.org/10.5281/zenodo.12625316}.

\bibitem[Zimmerlin et~al.(2024)Zimmerlin, Beißwenger, Jaeger, Geiger, and Chitta]{Zimmerlin2024ARXIV}
J.~Zimmerlin, J.~Beißwenger, B.~Jaeger, A.~Geiger, and K.~Chitta.
\newblock Hidden biases of end-to-end driving datasets.
\newblock \emph{arXiv.org}, 2412.09602, 2024.

\bibitem[Hafner et~al.(2023)Hafner, Pasukonis, Ba, and Lillicrap]{Hafner2023ARXIV}
D.~Hafner, J.~Pasukonis, J.~Ba, and T.~Lillicrap.
\newblock Mastering diverse domains through world models.
\newblock \emph{arXiv.org}, 2023.

\bibitem[Scheel et~al.(2021)Scheel, Bergamini, Wolczyk, Osi{\'n}ski, and Ondruska]{Scheel2021CORL}
O.~Scheel, L.~Bergamini, M.~Wolczyk, B.~Osi{\'n}ski, and P.~Ondruska.
\newblock Urban driver: Learning to drive from real-world demonstrations using policy gradients.
\newblock In \emph{Proc. Conf. on Robot Learning (CoRL)}, 2021.

\end{thebibliography}
}

\clearpage
\appendix
\maketitlesupplementary
\FloatBarrier
\section{Related work}
\label{sec:related_work}

\boldparagraph{Driving simulations} We use simulations to train and evaluate our methods in this work.
\textbf{CARLA} \cite{Dosovitskiy2017CORL} is a 3D open source autonomous driving simulator, based on the Unreal Engine, enabling real-time graphical rendering and simulation. CARLA enables research on the full driving stack from perception to planning and control, and has an extensive set of functionalities extended and developed over the last 8 years. The environments, assets, and scenarios are designed to offer a high-definition simulation of safety-critical events with arbitrary duration. In return, CARLA is considered computationally expensive, due to the 3D simulation and flexibility of the simulator. In this work, we use the simulation code of the CARLA leaderboard 2.0 together with the longest6 v2 benchmark \cite{Chitta2023PAMI}.

\textbf{nuPlan} \cite{Karnchanachari2024ICRA} is a data-driven simulator for vehicle motion planning that uses the real-world nuPlan dataset for simulation initialization or replay-based background traffic. As such, nuPlan simulates short scenes (\ie 15s) in a lightweight 2D setting on real-world maps with a kinematic bicycle model~\citep{Rajamani2011BOOK, Polack2017IV}. Compared to CARLA, nuPlan uses fewer resources at the cost of complexity. We use the CARLA leaderboard 2.0 to benchmark long-form driving with safety-critical scenarios and complement the results with nuPlan to test realistic everyday driving situations. A particularly interesting result of the nuPlan simulator and its benchmark was that it showed that rule-based \citep{Dauner2023CORL} and search-based \citep{Chekroun2024IV, Yang2024CVPR} methods still outperform pure learning-based approaches \citep{Hallgarten2023ITSC, Cheng2024ICRA, Cheng2024ARXIV}.

\textbf{Waymax} \cite{Gulino2023NEURIPS} is another recent bird's-eye-view (BEV) data-driven simulator, like nuPlan. It is different in that it uses a different underlying dataset \cite{Sun2020CVPR} and is implemented in Jax, which allows for GPU acceleration. It otherwise has the same strengths and weaknesses as nuPlan.
Several concurrent works \cite{Xiao2024ArXiV, Charraut2025ArXiV, Cusumano-Towner2025ARXIV} have reported that route information needed for proper simulation is currently proprietary in the WaymoOpen dataset.
As a workaround, these works have used the human expert trajectory as conditioning input to the model.
This can leak label information about an optimal trajectory to the policy, potentially leading to misleading results. 
Examples include leaking the lane the human expert drove on \cite{Cusumano-Towner2025ARXIV}, or even a fine-grained human path \cite{Charraut2025ArXiV}, trivializing lateral planning.
Due to this problem and because Waymax doesn't seem to offer fundamental advantages compared to nuPlan for benchmarking purposes, we are not using this benchmark in this work.
In nuPlan, conditioning is based on road blocks, which masks information about precise locations or which lane the human expert drove on.

\boldparagraph{Reinforcement Learning} Deep Reinforcement Learning \cite{Sutton2018MIT, Jaeger2024FTO} (RL) methods train neural networks, referred to as policy, model, or agent, by learning from trial and error. The goal and hence optimization objective of the trial and error learning is to maximize the (discounted) sum of rewards.
The RL framework offers two unique advantages compared to popular behavior cloning for the problem of driving. 
First, RL methods learn online, which avoids the compounding error problem \cite{Ross2011AISTATS} typical for behavior cloning methods.
Second, RL methods can optimize non-differentiable rewards \cite{Jaeger2024FTO}, which makes it possible to align the optimization objective closely with the desired target metric.

\boldparagraph{Rewards for Driving} Due to the flexibility in optimization, there are many ways to define the reward for driving, making reward design an important open problem \cite{Knox2023AI, Chen2024PAMI}.

 Early work on RL methods in autonomous driving \cite{Kendall2019ICRA} focused on simplified settings such as lane keeping on an empty street. This simplified setting allowed the methods to use simple principled reward functions such as maximizing forward speed and ending the episode upon an infraction.
Similarly, such simple reward functions are also popular in other robotics fields, such as autonomous racing \cite{Fuchs2021RAL} or indoor navigation \cite{Wijmans2020ICLR, Zeng2024CORL}, but only in settings where there are no other dynamic actors in the environment.

Lu et al. \cite{Lu2023IROS} optimize a simple reward, with only collision and an off-road penalty, in dynamic traffic. This reward does not include any term for progress, so a policy that always stays stationary would be optimal in reactive traffic.
The model only learns to drive because it is additionally trained with behavior cloning and because other cars will crash into a static ego car in log replay traffic.
Therefore, such a reward is not reasonable for pure RL in reactive traffic, which is the problem we investigate in this work.

When training with RL, in settings with dynamic actors, such as other cars, it is reported that simple rewards do not suffice anymore \cite{Liang2018ECCVb, Toromanoff2020CVPR, Song2021ICRA, Wurman2022Nature}.
To address this, complex reward terms are engineered that give additional local signals to the learning algorithm. 
This makes optimization easier, because it simplifies the credit assignment problem, allowing weaker learning recipes to succeed but potentially upper bounds performance as the reward may not be well aligned anymore with the original objective.

Toromanoff et al. 2020 \cite{Toromanoff2020CVPR} was the first RL method to train a planning policy head from scratch with RL in the dynamic and reactive traffic of the CARLA leaderboard 1.0.
They proposed to reward the difference between a desired vehicle state and the current vehicle state (desired position, rotation, and speed). 
This desired vehicle state is computed with privileged rules and provides very dense information for the learning agent.
This provides dense feedback, simplifying the learning problem, but limits the agent's performance to the performance of the rule-based planner used to compute the reward.
Due to its success, this general reward design became the dominant paradigm for state-of-the-art RL methods in CARLA \cite{Zhang2021ICCV, Chekroun2023ROBOTICS, Li2024ECCV}.
Our work shows that this trade-off is not necessary. We propose a more principled reward, free of rule-based experts, and show that it enabled data scaling via large mini-batch sizes.

\boldparagraph{RL Planners} \textbf{Roach} \cite{Zhang2021ICCV} is the first method to train a planner in CARLA with PPO \cite{Schulman2017ARXIV} uses a similar reward to \cite{Toromanoff2020CVPR}, which computes its reward based on the difference to a desired vehicle state. 
While the paper worked with the CARLA simulator, it is important to note that the conference paper erroneously claims SOTA performance on the CARLA leaderboard. The method was instead trained and evaluated on a newly created CARLA benchmark, which reused the route definitions of the CARLA leaderboard but did not include any safety-critical scenarios. This was later corrected in version 3 of the ArXiv \cite{Zhang2021ARXIVb} report.
A major problem with Roach was that its open-source training code is incompatible with the CARLA scenario runner that is used to evaluate safety-critical scenarios.
Some works made the inference code compatible \cite{Wu2022NEURIPS, Jia2023CVPR, Jia2023ICCV}, but the model achieved low scores when evaluated with safety-critical scenarios \cite{Ishida2024ICLRW} because it has never encountered these scenarios during training. Methods that used Roach for data collection combined the model with rule-based planners to post-process its actions.
Instead of this workaround, we rewrote the training code from scratch for the CARLA leaderboard 2.0.
We build our code upon CleanRL \cite{Huang2022JMLR} and use the open-source Roach codebase to match implementation details precisely.
We reproduce the Roach method on the CARLA leaderboard 2.0 longest6 v2 benchmark. 
We observe that Roach achieves much lower scores than rule-based \cite{Sima2024ECCV} or imitation learning \cite{Renz2022CORL} based baselines.
We build upon Roach and show that PPO can optimize our more principled reward when the mini-batch size is increased.

\textbf{Think2Drive} \cite{Li2024ECCV} is a recent RL planner trained with the CARLA leaderboard 2.0 that advocates for the use of world models \cite{Ha2018NEURIPS, Hafner2020ICLR, Hafner2021ICLR, Hafner2025NATURE} instead of using PPO.
World models simulate the original simulator with a faster neural network-based simulator. 
This is necessary because the default CARLA leaderboard 2.0 is too slow to converge PPO. 
The resulting world model simulator is imperfect but able to produce more samples in the same wall clock time. 
The policy is then trained exclusively with the approximated data from the world model. 
Using World models in a simulator is a compromise that trades off larger sample efficiency against lower asymptotic performance of the policy, because flaws in the world model translate to flaws in the policy \cite{Jaeger2024FTO}.
We show that this compromise is not necessary on CARLA.
We demonstrate that a two orders of magnitude increase in training throughput can be achieved through a combination of software engineering, better hyperparameters, scale, and a more efficient scaling technique in the CARLA leaderboard 2.0.

Think2Drive claims to achieve 56.8 Driving Score on the official test routes of the CARLA leaderboard 2.0 test routes.
It is not described what these "official" test routes refer to. The CARLA leaderboard 2.0 has an official test server, however, it can only be used for sensorimotor agents. 
Privileged agents like Think2Drive cannot be evaluated on it, because the terms of service explicitly prohibit it. 
The CARLA leaderboard 2.0 has a set of 2 routes called DevTest, which are in the training town 12 and are likely meant for development testing (debugging). 
Evaluating on these or the more principled validation routes (which contain 20 routes and are on the validation town 13) requires solving roughly 90 safety-critical scenarios consecutively along roughly 10 km-long routes. 
Think2Drive reports an average success rate per scenario of 84\% (average of Table 2 in the paper). One can estimate the Driving Score when solving 90 scenarios consecutively at an 84\% individual success rate. We ran a simple Monte Carlo simulation, averaging 10000 trials assuming perfect route completion and no other infractions, and using the median penalty factor of 0.65. This resulted in an average of 14 infractions per route and an average of 0.6 DS.
This is much lower than the reported 56.8 DS, which makes it unlikely that the unmodified devtest or validation routes are meant.
The authors uploaded a video of the model driving on the test routes. The video does not showcase that the model can solve multiple scenarios consecutively, but instead cuts after every individual scenario.

The code and model of Think2Drive are not public, so it is not directly possible to verify the claims made in the paper.
To alleviate this issue, we reproduce Think2Drive on the CARLA leaderboard 2.0  longest 6 v2 benchmark and will publish the code and models.

\textbf{GPUDrive} \cite{Kazemkhani2025ICLR} and \textbf{GIGAFLOW} \cite{Cusumano-Towner2025ARXIV} are concurrent works that speed up RL for planning by coding GPU-accelerated driving simulators. 
These driving simulators approximate other simulators by removing functionality and approximating expensive computations. 
GPUDrive approximates Waymax \cite{Gulino2023NEURIPS} using the assets of the Waymo Open dataset \cite{Sun2020CVPR}, whereas GIGAFLOW uses the map assets from CARLA \cite{Dosovitskiy2017CORL}.
In some sense, writing approximate faster simulators is similar to learning world models, with the advantage that with manual coding, the approximation errors can be controlled precisely.
Since the approximate simulator might be easier to solve than the original problem, it is important to evaluate on the original benchmark after training with the approximate simulator.
In this work, we similarly train with a faster version of the CARLA leaderboard 2.0, but evaluate with the original code.
GPUDrive does not evaluate its policy on Waymax and has no non-RL baselines, so it is hard to evaluate the performance of the policy.
GIGAFLOW, on the other hand, demonstrates Sim2Sim transfer evaluating their policy on nuPlan, Waymax, and the CARLA leaderboard 1.0.
While the GIGAFLOW training speed is remarkably efficient, achieving over 1 million FPS training throughput, their learning recipe is remarkably inefficient, requiring 1 trillion samples to converge.
Our best model (see \secref{sec:nuPlan_result}) achieves similar scores (93.1 vs 93.8) on nuPlan val14 reactive with 1 billion samples, 1000 times less data, while using the same learning algorithm \cite{Wijmans2020ICLR}.
Since GIGAFLOW simulates faster, this leads to similar overall compute requirements.
Our method is perhaps a better candidate for future extensions to end-to-end RL with vision because our 1B samples requirement can be achieved with a roughly 1000 FPS training throughput.
Some works have achieved such throughputs with vision in indoor navigation simulators \citep{Wijmans2020ICLR, Zeng2024CORL}.
Training the deep networks required for vision at the 1 Million FPS requirement of GIGAFLOW might need LLM pre-training level compute.

GIGAFLOW tries to address the problem of trading off different reward components by aggressively randomizing the tradeoff parameters.
They learn a variety of policies by conditioning the model on the reward parameters. 
The parameters then need to be tuned at inference to recover a good planner.
This leads to a strong driving policy, but might increase sample requirements as many policies need to be learned.
Instead, we propose to eliminate the reward tradeoffs by optimizing a single reward component.

Both GIGAFLOW and GPUDrive learn from all vehicles \cite{Chen2022CVPRa} because the policy is trained via self-play to control all actors in the scene. 
The number of samples used in both works is multiplied by the number of actors in each simulator because of that. 
Whether multi-agent samples have the same value as time step samples is unclear, since they could also be viewed as augmented versions of the same sample.
Self-play is a complementary technique to our work that can make a policy robust to varying traffic behaviors and therefore a promising approach for future attempts at Sim2Real transfer.

\boldparagraph{Asynchronous collection} PPO synchronizes many parallel environments by running batch model forward passes during data collection, which can lead to idle resources if the time of environment steps is variable. 
Envpool \citep{Weng2022NEURIPS} proposed to alleviate the PPO collection synchronization problem by introducing a collection mini-batchsize where only a minibatch of ready environments is forward passed at a time.
This alleviates the synchronization problem by reducing the number of synchronized environments, but has the downside that the number of sequential forward passes needed to collect the data is increased, which can slow down data collection in cases where the environment step times are similar. 
Additionally, it oversamples data from fast environments, which in CARLA are situations with fewer vehicles, hence easier samples.
Our asynchronous collection solution, proposed in \secref{sec:acppo}, does not bias the data collection towards faster environments and keeps the number of sequential forward passes to the network the same.

Various works have proposed fully asynchronous policy gradient methods \cite{Mnih2016ICML, Vinyals2019Nature, Berner2019ARXIV, Petrenko2020ICML}, which are methods that do not synchronize the backward pass.
These methods train faster but often do not study the effect of policy lag. 
By policy lag, we mean the \textit{variable and uncontrollable} amount of off-policy policy gradient error that gets introduced if training and data collection are decoupled.
This makes it unclear whether fully asynchronous methods outperform approaches that do synchronize the backward pass, given a fixed compute budget.
Anecdotal evidence \cite{Berner2019ARXIV} (figure 5b, policy lag is called data staleness) suggests that the negative effect of policy lag can be quite severe.
A more rigorous comparison between backward synchronized and fully asynchronous policy gradient methods is an interesting direction for future work.
Our proposed method, AC-PPO, synchronizes the backward pass during training which avoids the problem of policy lag.

\section{Method changes}
\label{sec:method_changes}

For our final model, trained with 300M samples, we have made several changes to Roach. In particular, increasing the field of view (FOV) of the input is important to achieve Driving Scores > 50 on longest6 v2, although it decreases performance in low data regimes, since the task becomes harder with the resulting smaller objects and larger amount of pixels.
The following tables are structured such that each row is the baseline for the next.
We found these changes helpful in preliminary experiments, although some of them have little impact on the model trained with a 1024 mini-batch size since it seems to already be close to the best performance that it can get given its limited FOV.

\subsection{Asynchronous data collection with PPO}
\label{sec:acppo}

PPO parallelizes data collection across multiple environments.
In environments whose time steps take a variable amount of time, also called nonhomogeneous environments \citep{Wijmans2020ICLR}, PPO is suboptimal in terms of efficiency \citep{Weng2022NEURIPS}.
PPO performs batched forward passes with the model during data collection, which synchronizes each environment after each environment step. 
CARLA is a nonhomogeneous environment, where the speed of a step varies between towns and traffic densities.
Additionally, environment resets are significantly slower. 
This means all parallel environments wait for the slowest, leading to unnecessary idle times.

\begin{table}[h]
\centering
    \begin{tabular}{l | c c }
        \toprule
        \textbf{Method} & \textbf{FPS} $\uparrow$ & \textbf{Train time h} $\downarrow$  \\
        \midrule
        PPO \cite{Schulman2017ARXIV} & {130} & {22} \\
        AC-PPO & \textbf{144} & \textbf{19} \\
        \bottomrule
    \end{tabular}
    \vspace{0.2cm}
    \caption{\textbf{PPO vs AC-PPO}}
    \label{tab:ac_ppo}
\end{table}

We propose to remove the collection synchronization, running each forward pass concurrently with batch size 1 in parallel. 
We name this approach Asynchronous Collection Proximal Policy Optimization (AC-PPO).
Running multiple model forward passes concurrently on a GPU efficiently requires using multi-threading and CUDA streams to share the model weights and CUDA contexts.
Unfortunately, Python does not support concurrent multi-threading because the global interpreter lock prevents concurrent execution of threads.
This is a known limitation of the Python programming language, and the developers are actively working on improving it \citep{Python2025Online}.
For now, AC-PPO can be implemented in other languages. We re-implemented our training code in C++, using the LibTorch interface of PyTorch \citep{Paszke2019NeurIPS}.
\tabref{tab:ac_ppo} shows that AC-PPO trains 11 percent faster than PPO using the same hardware and performing the same computations (up to random seeds). This reduces training time by 3 hours. 

\subsection{Hyperparameter}
We further reduced the number of training epochs from 4 (Atari) to 3 in CARLA.
As \tabref{tab:epochs} shows, this increased DS by 2 points and slightly reduced training time.
\begin{table}[h]
\centering
    \begin{tabular}{c | c c }
        \toprule
        \textbf{Epochs} & \textbf{DS} $\uparrow$ & \textbf{RC} $\uparrow$ \\
        \midrule
        4 & {38} \pmsd {3} & \textbf{85} \pmsd {5} \\
        3 & \textbf{40} \pmsd {2} & {84} \pmsd {1} \\
        \bottomrule
    \end{tabular}
    \vspace{0.2cm}
    \caption{\textbf{Effect of reducing training epochs.}}
    \label{tab:epochs}
\end{table}

The full list of hyperparameters used in Atari, Roach, and our method is displayed in \tabref{tab:hyperparameter_all}.

\begin{table}[th]
\small
\centering
    \begin{tabular}{c | c c c c}
        \toprule
        \textbf{Hyperparameter} & \textbf{Roach} \cite{Zhang2021ICCV} & \textbf{Atari} \cite{Schulman2017ARXIV, Huang2023ICLRBLOG} & \textbf{CaRL}(CARLA)  & \textbf{CaRL}(nuPlan) \\
        \midrule
        Samples & 10M & 10M & 300M & 500M/\textbf{1B} \\
        Initial Learning rate & 0.00001 & \textbf{0.00025} & \textbf{0.00025} & \textbf{0.00025} \\
        Learning rate schedule & KL patience & linear & linear & linear \\
        num envs & 6 &  8  & 128 & 512/\textbf{1024} \\
        env steps per iteration & \textbf{2048} &  128  & 512 & 32/32\\
        batch size & 12288 & 1024 & \textbf{65536} &  16384/32768 \\
        mini-batch size & 256 & 256 & \textbf{16384} & 4096/8192 \\
        Steps per epoch & \textbf{48} & 4  & 4 & 4  \\
        Epochs & \textbf{20} &  4  & 3 & 4 \\
        Steps off-policy &  \textbf{959} & 15 & 11 & 15 \\
        Exploration Hints & $\checkmark$ & \xmark & \xmark & \xmark \\
        norm advantage & \xmark & $\checkmark$ & $\checkmark$ & $\checkmark$ \\
        clip value loss & \xmark & $\checkmark$ & $\checkmark$ & $\checkmark$ \\
        entropy coefficient  & \textbf{0.01} & \textbf{0.01} & \textbf{0.01} & \textbf{0.01} \\
        value function coefficient & \textbf{0.5} & \textbf{0.5} & \textbf{0.5} & \textbf{0.5} \\
        discount factor & \textbf{0.99} & \textbf{0.99} & \textbf{0.99} & \textbf{0.99} \\
        GAE $\lambda$ & 0.9 & \textbf{0.95} & \textbf{0.95} & \textbf{0.95} \\
        clipping coefficient & \textbf{0.2} & 0.1 & 0.1 & 0.1 \\
        Max gradient norm & \textbf{0.5} & \textbf{0.5} & \textbf{0.5} & \textbf{0.5} \\
        \bottomrule
    \end{tabular}
    \vspace{0.1cm}
    \caption{\textbf{PPO hyperparameters used in Atari, Roach and CaRL}}
    \label{tab:hyperparameter_all}
\end{table}

\subsection{Architecture}
\label{sec:changes_architecture}

We follow the architecture of \cite{Zhang2021ICCV}, consisting of 6 convolutional layers for the bird's-eye-view (BEV) semantic image, encode the measurements with 2 layer MLPs, merge the features with a 2 layer MLP and decode the features with a 2 layer MLP into $\alpha$ and $\beta$ values for the action probablity distribution and a seperate MLP for the value head.
All layers are followed by a ReLU activation function.

In addition to this basic architecture, we found it helpful to introduce a layernorm before every ReLU activation function.
\begin{table}[h]
\centering
    \begin{tabular}{l | c c }
        \toprule
        \textbf{Architecture} & \textbf{DS} $\uparrow$ & \textbf{RC} $\uparrow$ \\
        \midrule
        Cnn \cite{Zhang2021ICCV} & {40} \pmsd {2} & {84} \pmsd {1} \\
        + Layernorm & \textbf{42} \pmsd {4} & {93} \pmsd {2} \\
        + Asymmetric Critic & {41} \pmsd {3} & \textbf{94} \pmsd {3} \\
        \bottomrule
    \end{tabular}
    \vspace{0.2cm}
    \caption{\textbf{Architecture.}}
    \label{tab:architecture}
\end{table}

\tabref{tab:architecture} shows that layernorm improved the driving score by 2 points and the route completion by 9 points.
One hypothesis for why the layernorm is helpful might be that it increases the plasticity of the network \cite{Lyle2023ICML}. 
Maintaining plasticity is particularly important in continual learning tasks \cite{Dohare2024NATURE, Juliani2024NeurIPS}. 
On-policy reinforcement learning is a non-stationary optimization problem, because if the policy changes, so will the states it observes. 
To fit the new states, the network needs to maintain plasticity \cite{Moalla2024NEURIPS}.
By increasing plasticity layernorm might be a generally helpful tool for on-policy RL.

Additionally, \tabref{tab:architecture} shows the impact of using an asymmetric critic. 
The result is similar, with the driving score reduced by 1 point and the route completion increased by 1 point.
Asymmetric critic means that the value function gets additional information that the policy doesn't know.
These are inputs that are unnecessary for driving but help predict the cumulative rewards.
In particular, we use the time until a timeout happens, a blocked infraction happens, the remaining length of the route, how long any time to collision penalty is still applied, and for how many frames a comfort penalty is still applied, all appropriately normalized.

\subsection{Action distribution}
\label{sec:changes_action}
The action space of the car is steering $\in [-1,1]$, throttle $\in [0,1]$ and brake $\in [0,1]$. Throttle and brake are merged into one action dimension $\in [-1,1]$, making them mutually exclusive.
These continuous values require discretization or a continuous probability distribution.
Zhang et al. \cite{Zhang2021ICCV} report that the Normal distribution performs much worse than the Beta distribution, which our preliminary experiments confirmed.

The Beta distribution introduces an additional choice:
The Beta distribution is parametrized by two parameters $\alpha, \beta \in (0, \infty)$. 
For $\alpha, \beta \in (0,1)$, the distribution assigns a lot of probability to the extreme ends of the bounded action space and can become multi-modal.
This overweights extreme actions such as full throttle or brake, which should be used very rarely as they lead to uncomfortable driving. Additionally, the distribution degenerates as $\alpha, \beta \rightarrow 0$.
Instead of using the full range of $\alpha, \beta$ like \cite{Zhang2021ICCV}, we restrict the network to predict values $\in [1, \infty)$ by adding 1 to the Softplus activation function, following \cite{Chou2017ICML, Petrazzini2021SSCI}.
This ensures that the distribution is unimodal and does not degenerate as $\alpha, \beta \rightarrow \infty$ is needed for the distribution to collapse.

\begin{table}[h]
\centering
    \begin{tabular}{l | c c }
        \toprule
        \textbf{Action distribution} & \textbf{DS} $\uparrow$ & \textbf{RC} $\uparrow$ \\
        \midrule
        Beta $\alpha,\beta \in (0, \infty)$ & \textbf{41} \pmsd {3} & \textbf{94} \pmsd {3} \\
        Beta $\alpha,\beta \in (1, \infty)$ & \textbf{41} \pmsd {2} & \textbf{94} \pmsd {2} \\
        \bottomrule
    \end{tabular}
    \vspace{0.2cm}
    \caption{\textbf{Action distribution parameterization}}
    \label{tab:action_distribution}
\end{table}

\tabref{tab:action_distribution} shows that both variants achieve the same driving score. The driving score does not consider comfort, so improvements in comfort are not measured.

\begin{figure}[htp]
    \centering
    \includegraphics[width=\textwidth]{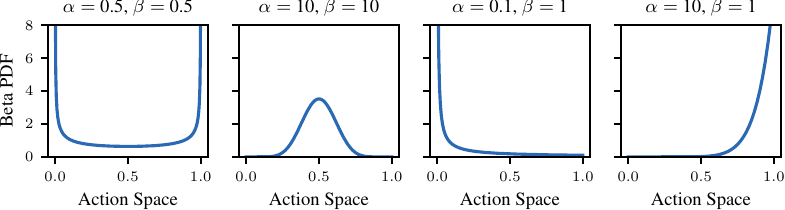}
    \caption{\textbf{Beta distribution examples.}}
    \label{fig:beta}
\end{figure}

\figref{fig:beta} shows 4 different shapes of the Beta distribution PDF to give the reader some intuition. 
The $\alpha, \beta$ values are displayed in the title of the plot. The first and third plots show examples for $\alpha, \beta <= 1$. 
The distribution can become multimodal (first plot) or degenerate towards either extreme (third plot). 
When $\alpha, \beta >= 1$, then the distribution behaves more like a normal distribution, if $\alpha \approx \beta$ (second plot), and does not degenerate quickly if $\alpha < \beta$ or vice versa (fourth plot).

\subsection{Input}
\label{sec:changes_input}
Similar to prior work \cite{Zhang2021ICCV, Li2024ECCV}, we use a BEV semantic segmentation image as input. The channels consist of a channel for the road, an A$^\star$ route mask, lane markings, vehicles, pedestrians, traffic lights, stop signs, speed signs, static objects, and shoulder lanes.
Unlike prior work \cite{Zhang2021ICCV, Li2024ECCV}, we do not use additional channels for the states of pedestrians and vehicles from past time steps. Instead of using a binary mask for the vehicles and pedestrians, we encode their speed value in the bounding box.
This reduces GPU memory requirements.
Similar to concurrent work \cite{Cusumano-Towner2025ARXIV}, we find historical information, aside from the speed of vehicles, to not be necessary for the CARLA and nuPlan benchmarks.
Additionally, we render open doors, a constant velocity forecast, the blinker, stop, and emergency lights of the car into the car channel via special bounding boxes.

The route mask channel renders the lane-level route from the CARLA A$^\star$ route planner that determines where the car should go. 
This is meant to condition the model such that it knows where it should drive at intersections, similar to the target points used in sensor-based agents \cite{Jaeger2023ICCV}, but much denser.
We observed that even if the reward function allows the agent to deviate from the lanes that the A$^\star$ planner picked, it would still follow the A$^\star$ lane most of the time.
This is likely because the model learns to ignore the lane markings and road channel and mostly abuses the A$^\star$ route mask channel for steering. 
This leads to the same pathologies that the other CARLA baselines have (see \secref{sec:failure_cases}), where the agent lane changes at pre-defined locations, leading to crashes on highways, even if the reward does not directly enforce the behavior as done in prior work.
These models have the characteristic 0.0 route deviation infractions, which are also typical for imitation learning models that abuse a similar bias \cite{Jaeger2023ICCV}.
We were able to break this behavior by only rendering the A$^\star$ route inside intersections, which enabled the policy to learn how to choose its driving lane by itself even inside intersections where there is conditioning.
This reduced the mentioned lane change collision, however, it also had the downside that the agent now sometimes takes the wrong turn at intersections.

The current policy incurs many infractions in situations where the FOV of the input is too small to see fast approaching cars in time.
Roach used 30m front, 8 m to the back, and 19m to each side.
This might have been fine for low-speed driving, but when driving at higher speeds, the car needs to be able to see other vehicles earlier since the time to stop the vehicle increases roughly quadratically with driving speed.
We increased the visible range to 78 m in front, 50 m in back, and 64 m to either side. 
This would naively make the input image much larger. To stay in a similar range, we decreased the pixels per meter resolution from 5 to 2 and increased image resolution from 192x192 to 256x256. 
To account for the increase in resolution, we added an additional Conv2D layer with stride 2 to the network to keep the number of output features from the CNN constant.
Our final model has 2 million parameters.

\begin{figure}[th]
\centering
\includegraphics[width=0.95\textwidth]{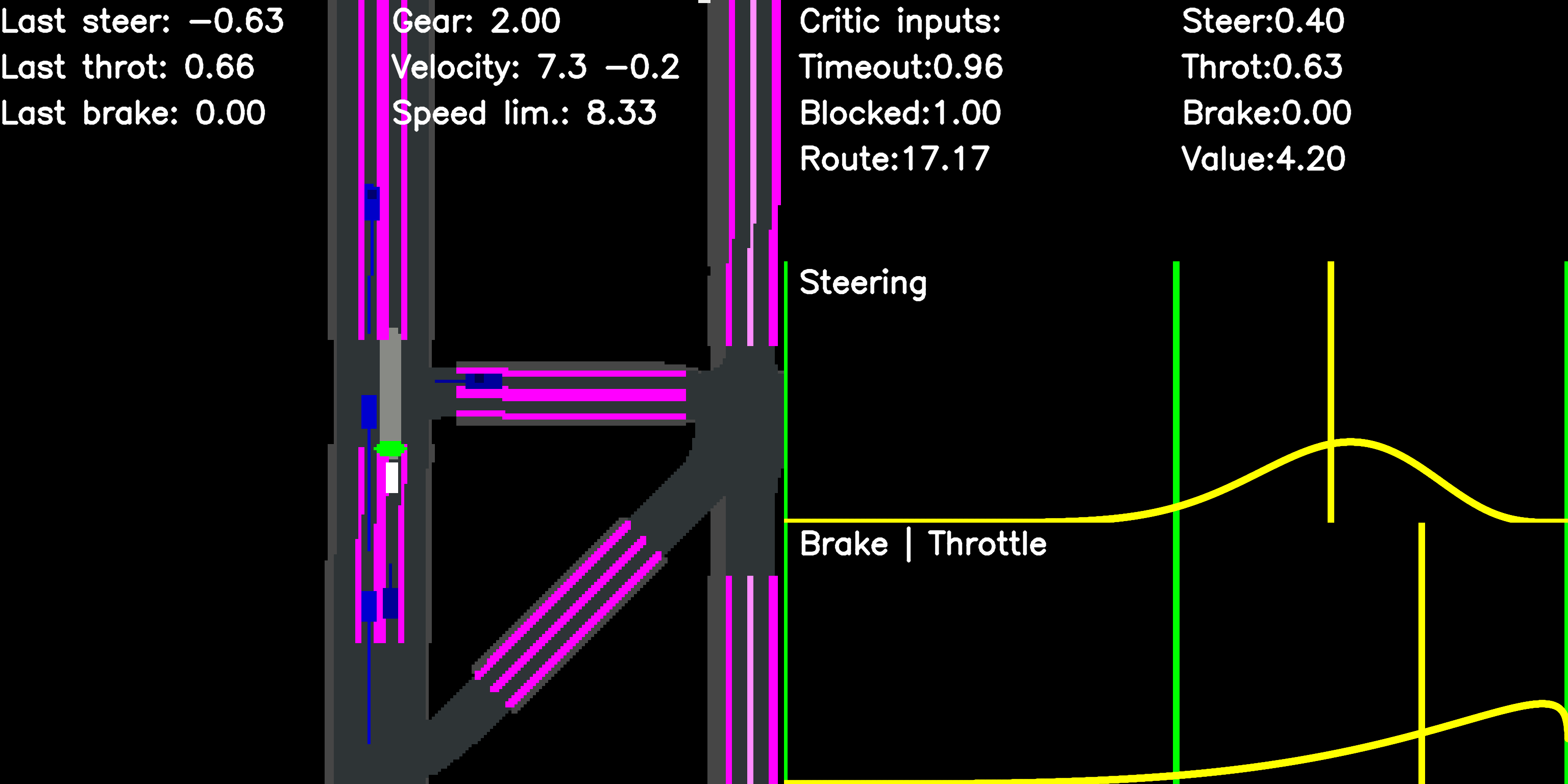}
\caption{\textbf{A rendering of our input}. The distribution shows the model action distribution predictions. The yellow vertical line denotes the mean of the distribution. Other cars are rendered in blue. The brightness of blue encodes their speed. A constant velocity forecast is rendered as a line in front of other vehicles. The ego car is depicted in white. Conditioning is in light grey and only rendered inside intersections. Dark grey depicts the road. The lane markings are pink.}
\label{fig:observation}
\end{figure}

Similar to prior work we also use scalar measurements as input to the policy. 
We use the last steering, throttle, and brake of the car, the gear state, velocity, speed as well as the current speed limit.
As mentioned before, we additionally use extra scalar measurements that are input only to the value function, which may be helpful for value estimation but are unnecessary for driving.
\figref{fig:observation} shows an example of a rendering of the  BEV observation.

\tabref{tab:input} shows that using the observation with the larger field of view and no conditioning outside junctions decreases the driving score by 10 points and increases variance.
A closer look at the auxiliary metrics reveals that our input reduces vehicle collisions (Veh) by 0.5, because the policy can choose where to lane change and sees other vehicles earlier.
The driving score is worse because of a large drop of 31 points in RC which the driving score puts a strong emphasis on. 
Without the conditioning bias, the model is not consistent at taking intersections, leading to an increase of 0.3 in route deviations (Dev).
Our observation is preferable, since, as shown in the main paper, a lot of the lost route completion will be recovered by scaling up samples and mini-batch size.

\begin{table}[h]
\centering
    \begin{tabular}{l | l l l  l | c c | c c }
        \toprule
        \textbf{BEV Style} & \textbf{Size HWC} & \textbf{Front} & \textbf{Back} & \textbf{L/R} & \textbf{DS} $\uparrow$ & \textbf{RC} $\uparrow$ & \textbf{Veh} $\downarrow$ & \textbf{Dev} $\downarrow$ \\
        \midrule
        Roach \cite{Zhang2021ICCV} & 192x192x15 & 30m & 8m & 19m & \textbf{41} \pmsd {2} & \textbf{94} \pmsd {2} & {1.41} & \textbf{0.00} \\
        Ours & 256x256x10 & \textbf{78m} &\textbf{50m} & \textbf{64m} & {31} \pmsd {7} & {63} \pmsd {8} & \textbf{0.91}  & {0.30} \\
        \bottomrule
    \end{tabular}
    \vspace{0.2cm}
    \caption{\textbf{Input.} Our input has a larger Field of view but at lower resolution (2 pixels per meter instead of 5). This makes driving harder but enables the model to avoid collisions that are otherwise impossible to solve.}
    \label{tab:input}
\end{table}

\subsection{nuPlan}
\label{sec:nuPlanChanges}

This section provides the implementation details for training CaRL on nuPlan.

\boldparagraph{Hyperparamters} Compared to CARLA, a single nuPlan environment has a low memory and compute footprint, with fewer crashes during training. This enabled us to scale the number of parallel training environments for faster data collection and train with 500M/1B samples. We use the regular number of 4 training epochs. All parameters can be found in \tabref{tab:hyperparameter_all}.

\boldparagraph{Architecture} We use the same architectecture as described in \secref{sec:changes_architecture} and merely adapt the input layer dimensions. 

\boldparagraph{Action Distribution} The kinematic bicycle model of nuPlan takes the longitudinal acceleration (in $\text{m}/\text{s}^2$) and steering rate (in $\text{rad}/\text{s}$) of the planner output to propagate the ego agent. 
Most methods using the nuPlan simulator predict waypoints as an output representation because this was a requirement of the 2023 nuPlan challenge~\cite{Karnchanachari2024ICRA}.
Limiting the design space for output representations of planners to waypoint trajectories may limit their performance because the representation, while recently successful, is both arbitrary and ambiguous \cite{Jaeger2023ICCV}.
Since the input to the vehicle model is not waypoints, these predictions are converted by processing the trajectory with an LQR controller.
We demonstrate that this choice of trajectory representation, combined with LQR controller, can reduce the performance of the planner by comparing the performance of the human ground truth trajectory with three different controllers in \secref{sec:nuPlan_result}.

Instead of using a modular trajectory + controller solution, we integrate the control into the planner in this work, such that the neural network also has to learn control. 
We map the action distribution from \secref{sec:changes_action} to values for the bicycle model. We scale action output from $[-1, 1]$ to $[-3.2, 2.4]\text{ } \text{m}/\text{s}^2$ and $[-0.84, 0.84] \text{ rad}$, for the target acceleration and steering respectively. 
These limits were derived by applying the default controller on a large set of human trajectories. 
Next, we use a discrete-time derivative to convert the target steering angle to a steering rate and clip the longitudinal acceleration to restrict reverse driving. 

\boldparagraph{Input} We conceptually use the same input and extent as in CARLA from \secref{sec:changes_input}, with minor adjustments for nuPlan. For the route, we render the polygons of the route roadblocks that are provided as route information to a planner in nuPlan. We render speed limits and traffic-light states as polylines of the lane center. Since nuPlan does not consider stop-sign infractions, we dropped the corresponding channel to reduce rendering time. We adapt the scalar measurements and input the last steering and acceleration action, the velocity and acceleration along the longitudinal and lateral axis, the current steering angle and rate, as well as the angular velocity and acceleration.

\section{Failure cases}
\label{sec:failure_cases}

\subsection{Roach}
In this section, we show some examples of characteristic failure cases of the Roach reward function.
The Roach model only achieves 22 DS on longest6 v2, so there are more failure cases than displayed here.

\begin{figure}[th]
\centering
    \includegraphics[width=0.95\textwidth]{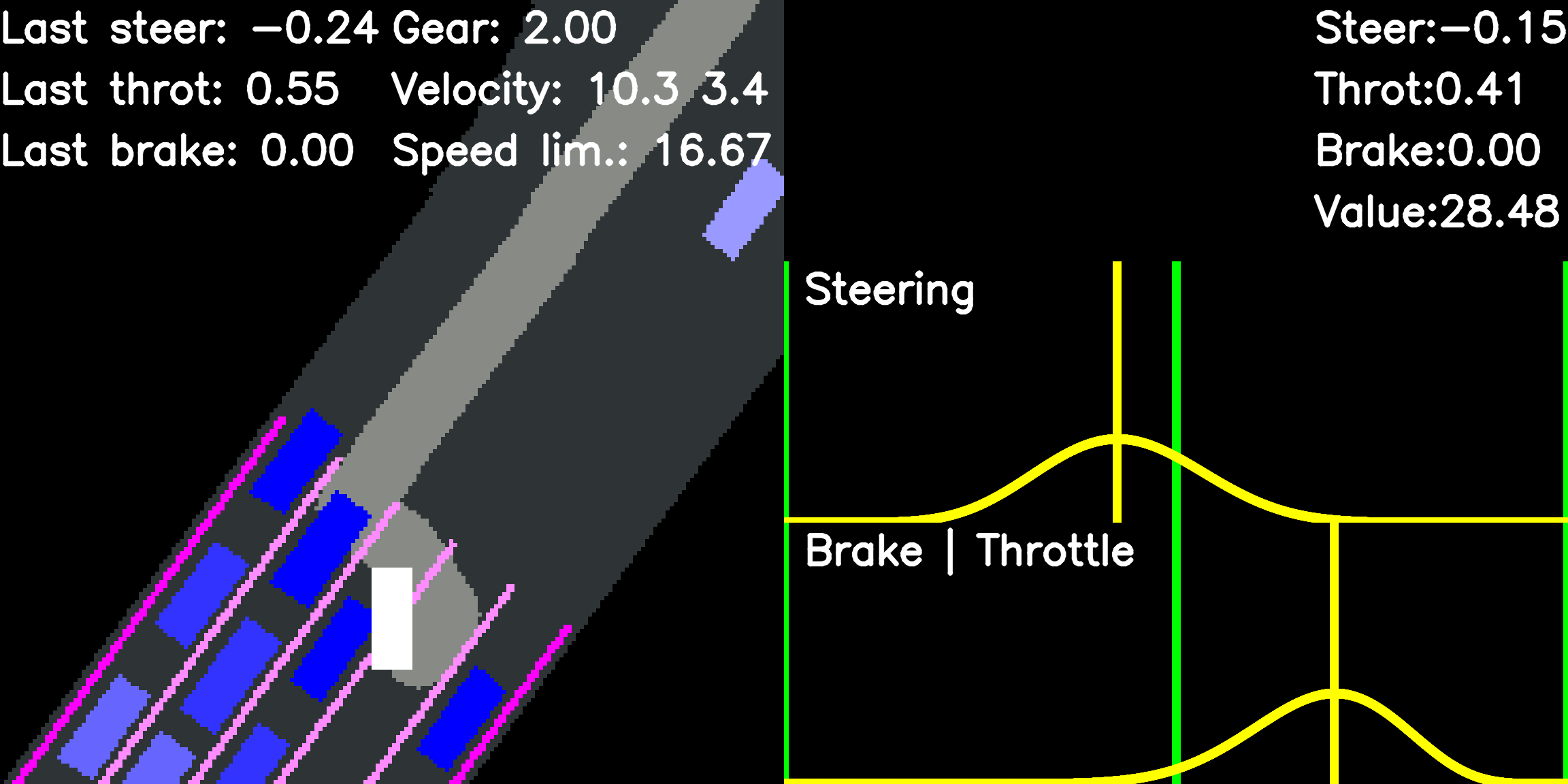}
   \caption{\textbf{Roach lane changes at predefined locations leading to rear collisions.} White is the ego car. \textcolor{blue}{Blue} other cars. Lighter shades of blue represent past time steps. \textcolor{customgray}{Gray} is the precomputed route from the A$^\star$ planner, and decides which lane to drive on.}
\label{fig:lane_change_collision}
\end{figure}

Another problem of the roach reward is the dependence on an A$^\star$ algorithm for local path planning. The route that the agent should follow is computed by an A$^\star$ algorithm. While this is, in principle, not a problem, the episode is also terminated if the policy drives 3.5 meters away from the lane chosen by the A$^\star$ algorithm. This is a problem on multi-lane roads because it means the agent cannot decide which lane to drive on based on the traffic situation. Additionally, the A$^\star$ algorithm lane changes at specific locations, which the agent has to adhere to. In case there is an oncoming vehicle in the target lane, the agent has to stop and wait for it to pass. Stopping at random locations on a highway is a dangerous policy that disrupts traffic flow and may lead to rear-end collisions.
Additionally, we observe that a frequent failure case of Roach is that it does not stop, leading to a collision with an oncoming vehicle, when changing lanes at these predefined locations. This is illustrated in \figref{fig:lane_change_collision}.
Aside from RL agents, who inherit it, this is also a major limitation of SotA rule-based planners \cite{Jaeger2023ICCV, Dauner2023CORL, Sima2024ECCV}.

\begin{figure}[th]
    \centering
    \includegraphics[width=0.95\textwidth]{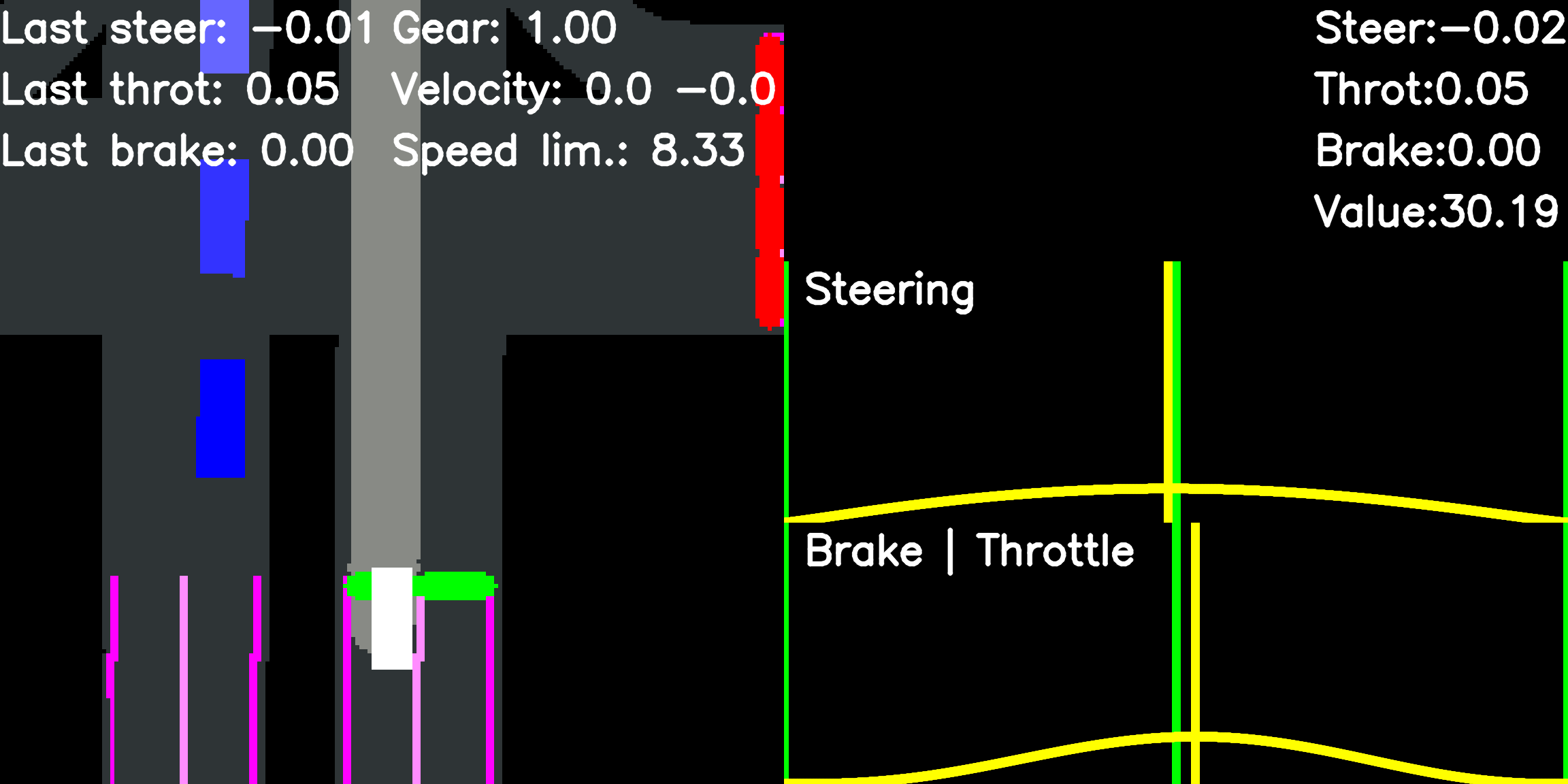}
    \caption{\textbf{Roach learns to wait at a green light early during training.} Note the 0 m/s velocity at the top of the image.}
    \label{fig:wait_green_light_1}
\end{figure}

\figref{fig:wait_green_light_1} shows the behavior of an early model after roughly 2 million samples. Roach waits at a green light, rendered as a \textcolor{green}{green} stop line. This is a very simple behavior that leads to high rewards since the agent will be given large speed rewards in future frames after the light turns red again. Unlike driving, waiting at a traffic light incurs no risk of collision.

\begin{figure}[th]
    \centering
    \includegraphics[width=0.95\textwidth]{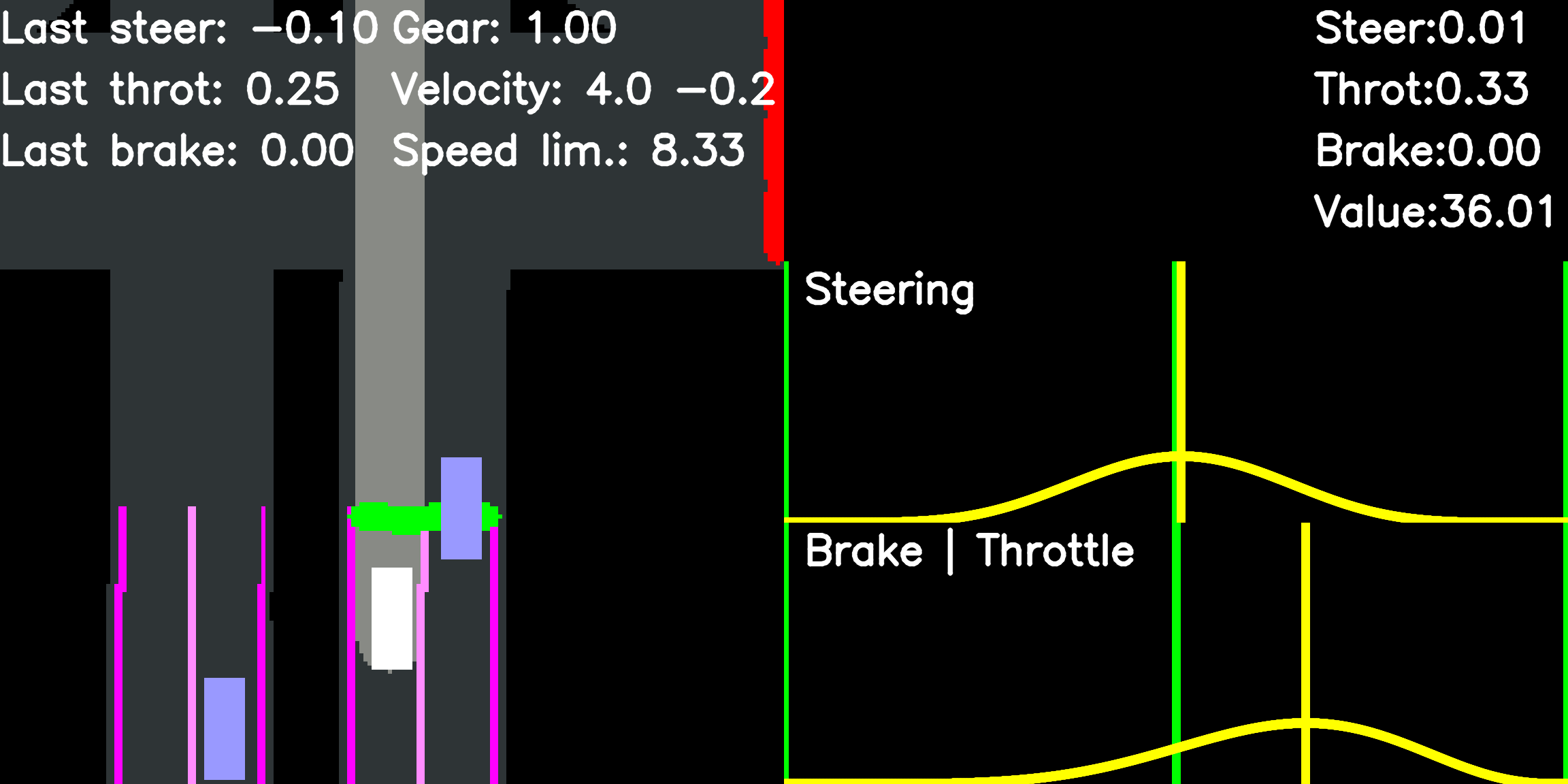}
    \includegraphics[width=0.95\textwidth]{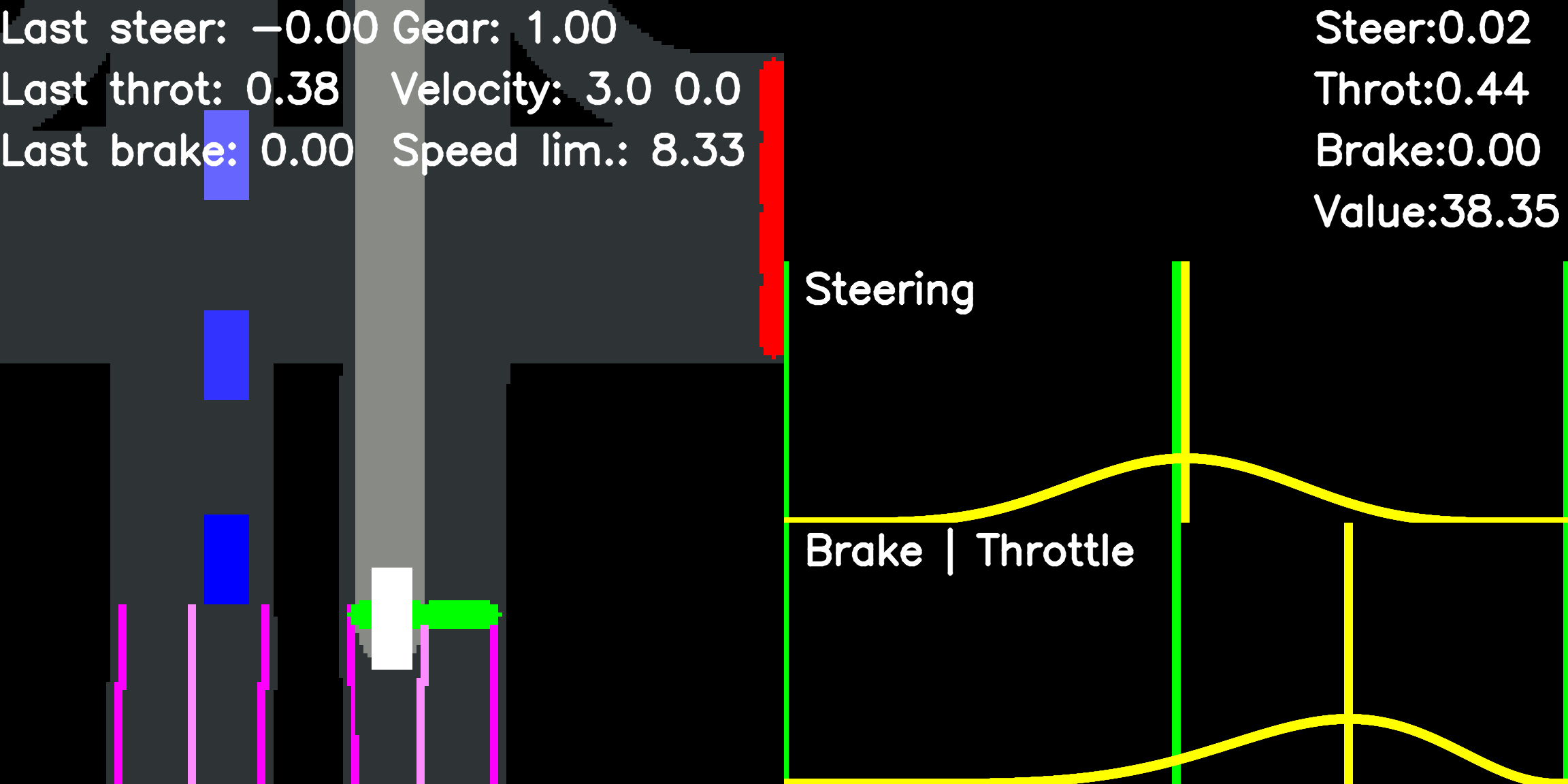}
    \includegraphics[width=0.95\textwidth]{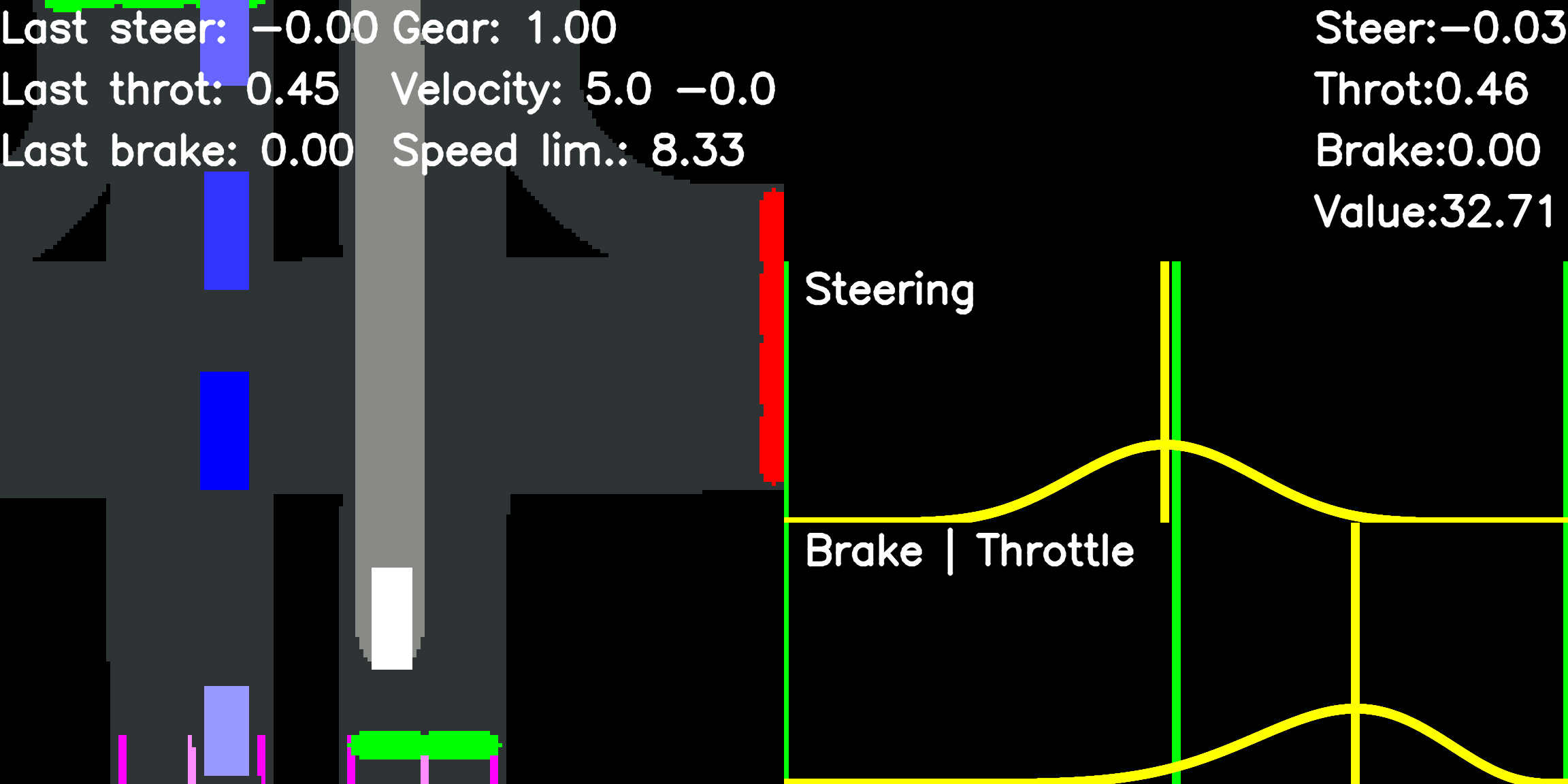}
    \caption{\textbf{Roach slows down at green lights.} Top to bottom are 3 different time steps. Note how the velocity initially slows down from 4 m/s to 3 m/s. Once the agent passed the green light, the model started accelerating again to 5 m/s.}
    \label{fig:wait_green_light_2}
\end{figure}

\figref{fig:wait_green_light_2} shows the behavior of a converged seed. Many training seeds eventually escape the behavior of waiting at green lights. They do, however, still slow down when approaching a green light, presumably trying to increase the chance of catching the next red light. Here, the model initially drives at 4 m/s, slows down to 3 m/s when it gets close to the traffic light. Once it passes the traffic light, the model starts accelerating again here to 5 m/s.
Roach is trained via its reward function to drive at most 6 m/s.

\subsection{CaRL}

\subsubsection{CARLA}
\begin{figure}[th]
    \centering
    \includegraphics[width=0.95\textwidth]{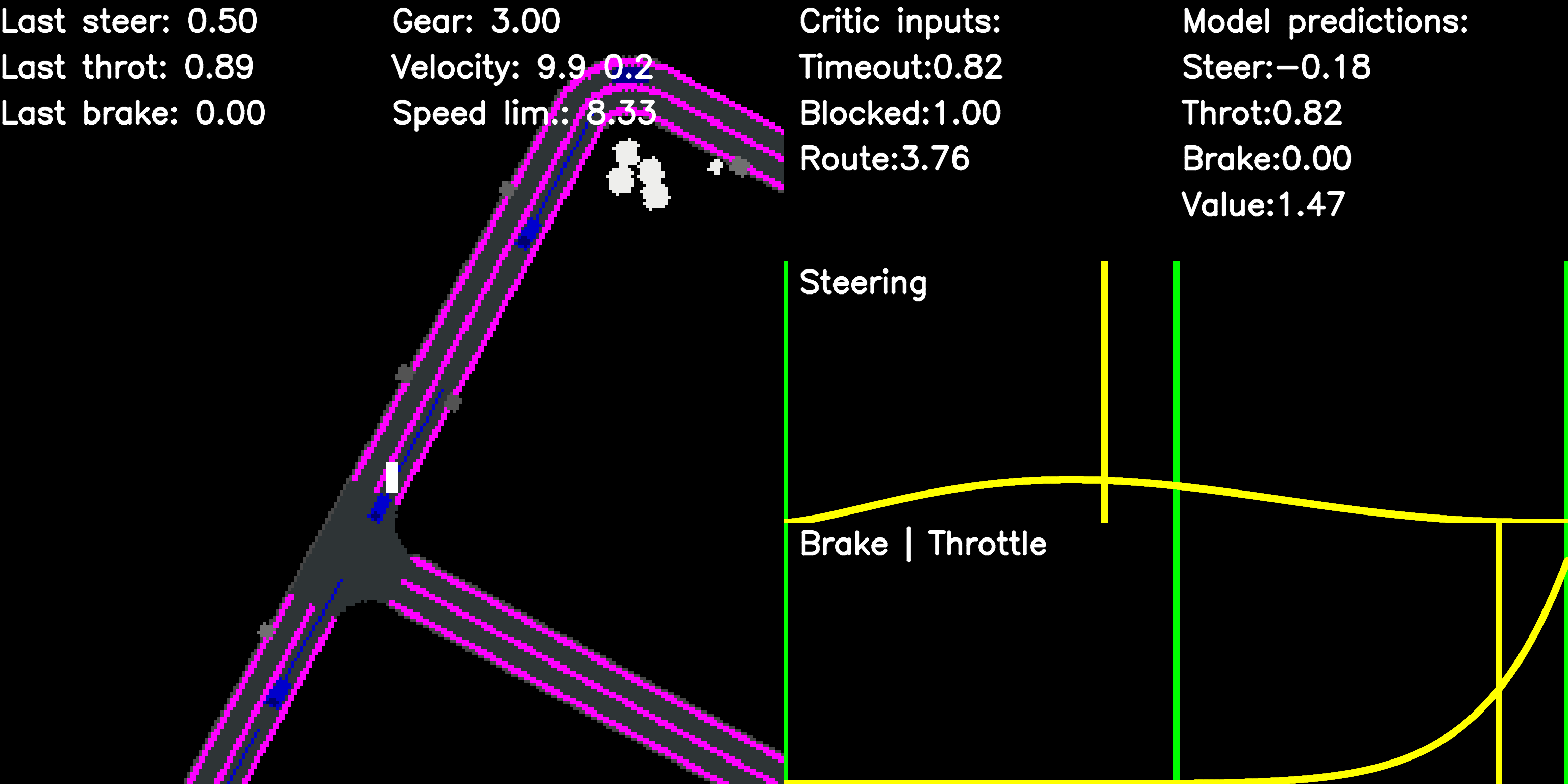}
    \caption{\textbf{Another car runs a red light and rams the CaRL from behind.}}
    \label{fig:run_red_light}
\end{figure}

Our final model, CaRL, has two common failure modes in CARLA, accounting for the majority of its infractions.
First, CaRL often fails in the safety-critical scenarios that involve other cars running red lights.
\figref{fig:run_red_light} shows an example where a non-reactive car is running a red light at high speed and crashes into CaRL from behind. CaRL tries to evade the opposing lane but accelerates too slowly to avoid the collision.

\begin{figure}[th]
    \centering
    \includegraphics[width=0.95\textwidth]{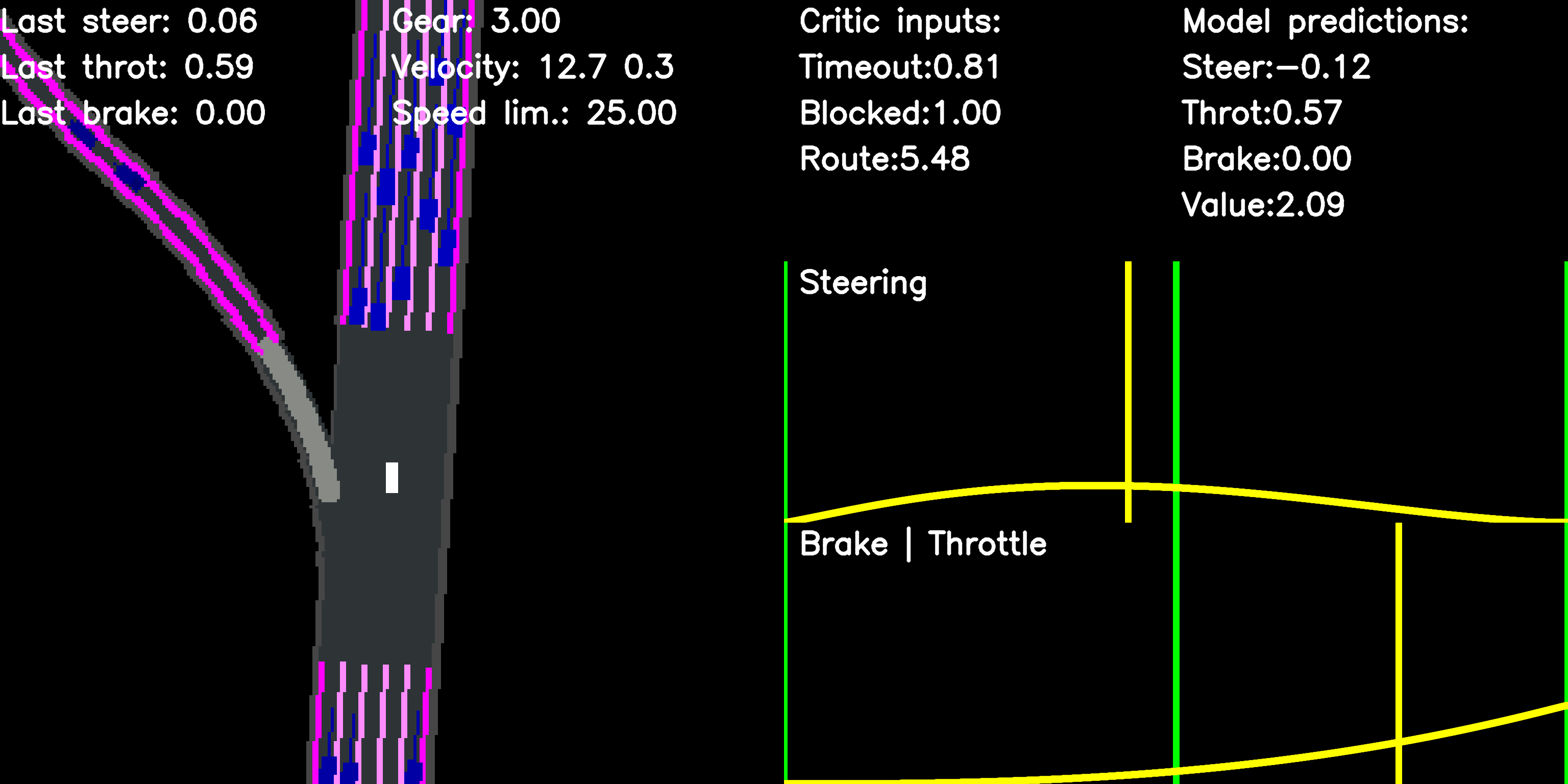}
    \caption{\textbf{CaRL misses a highway exit.}}
    \label{fig:highway_offramp}
\end{figure}

The second failure case is taking the wrong turn, most frequently missing highway off-ramps.
An example is illustrated in \figref{fig:highway_offramp} where CaRL continues driving on the highway instead of exiting on the left.

\subsubsection{nuPlan}

\begin{figure}[t!]
    \centering
    \subfloat[Pedestrian collision.]{%
       \includegraphics[width=0.3\textwidth]{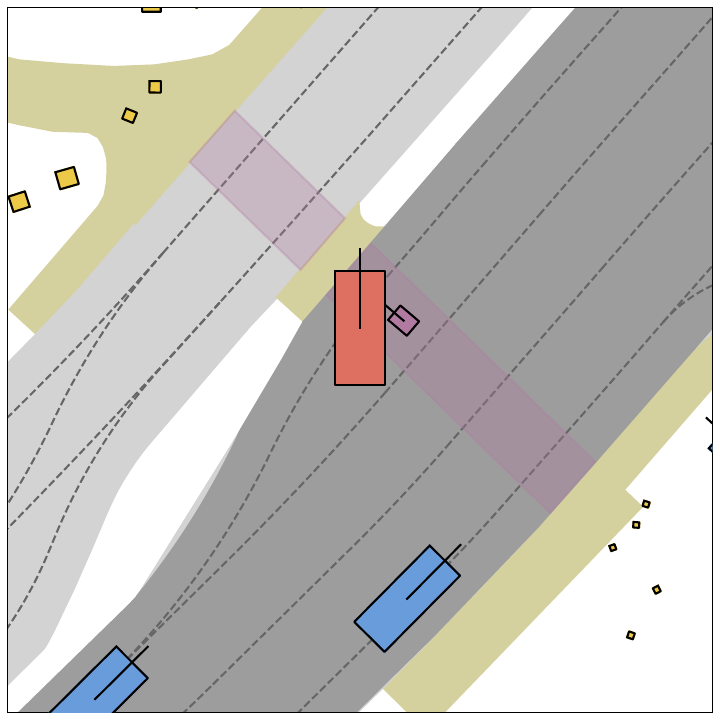}
     }
     \hfill
     \subfloat[Vehicle collision.]{%
       \includegraphics[width=0.3\textwidth]{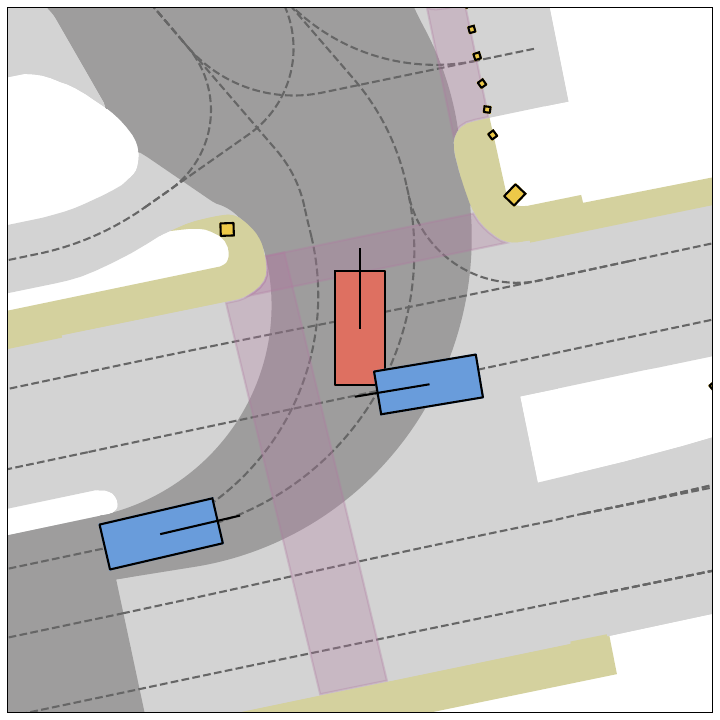}
     }
     \hfill
     \subfloat[Ego stuck in traffic.]{%
       \includegraphics[width=0.3\textwidth]{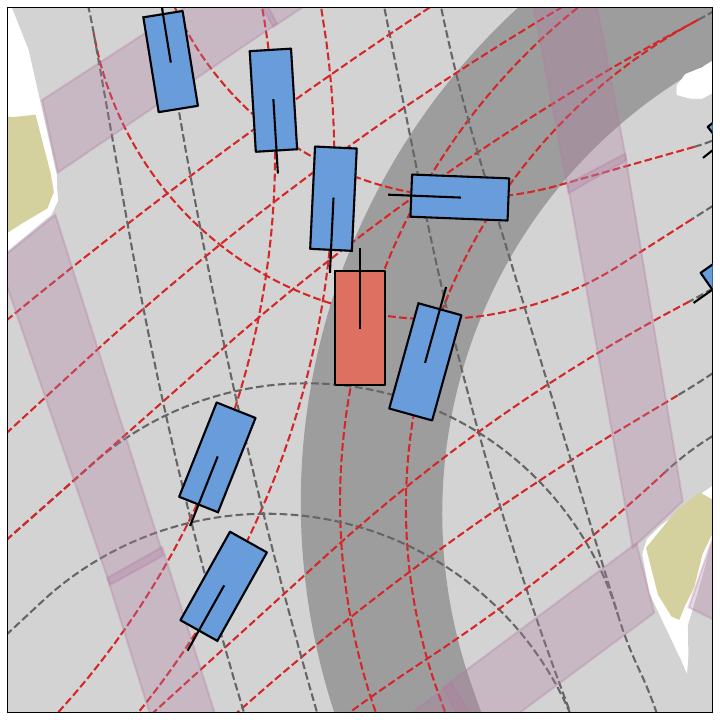}
     }
    \caption{\textbf{Failures of CaRL in nuPlan.} (a) The ego vehicle collides with non-reactive pedestrians and veers off-road in an attempt at collision avoidance. (b) During an unprotected turn, the ego vehicle has a rear-side collision with a non-reactive vehicle. (c) Non-ego vehicles regularly block the road in the reactive simulation, resulting in low progress scores.}
    \label{fig:nuplan_fails}
\end{figure}

We show different failure modes of CaRL in \figref{fig:nuplan_fails}. Avoiding collisions remains a challenge in nuPlan, partially due to the non-reactive agents or the abrupt behavior of reactive vehicles in the simulation modes. Specifically, pedestrians pose challenges to the planner. We estimate the small size and more flexible movement of pedestrians to be difficult for CaRL. Moreover, we point out that route completion is generally more difficult in the reactive nuPlan simulation, due to IDM vehicles blocking the road.

\section{Engineering}

In supervised learning models are much larger than in reinforcement learning which has the consequence that most of the computation happens in highly optimized library functions.
In RL in contrast models are comparatively tiny (we use a 2 million parameter model in this work) and most of the computation happens in user code.
Since data is simulated and PPO scales well with more data (as shown in this work and others), this has the consequence that \textit{the efficiency of the engineering has a direct impact on the performance of the model}, given a fixed time and compute budget. 
Doing (gradient) synchronized on-policy learning at scale poses special requirements to the simulation software, such as low loading times, fast time steps, and low variance between time steps.
Standard evaluation simulation software is often suboptimal on one or more of these axes.
In the following, we describe several changes we made to the CARLA leaderboard 2.0 and nuPlan simulation code (for training only) that improved our training throughput and enabled us to scale to 300 million and 1 billion samples.
We are focusing on the most important changes since it is not possible to describe every little engineering detail in writing.
The full details can be found in the code at \url{https://github.com/autonomousvision/CaRL}.

\subsection{CARLA}

Since we are training privileged models, we run CARLA in CPU mode using the "-nullrhi" option, which saves a lot of GPU memory.
This feature was silently introduced in CARLA 0.9.14 alongside the multi-GPU rendering feature.
The multi-GPU rendering feature decomposes the simulation into a CPU primary server and multiple GPU rendering servers.
It is possible, as long as no sensors are used, to use zero GPU rendering servers, which effectively introduced CPU-only simulation.

In the CARLA leaderboard 2.0, the environment controls the agent, whereas in almost all RL codebases, the agent controls the environment. This makes applying the standard gym(nasium) interface \citep{Brockman2016ARXIV, Towers2024ArXiv} hard. To resolve this, we split the environment and training code into two separate processes that communicate with each other via inter-process communication \citep{Hintjens2013BOOK}. This incurs an overhead for sending the observations to the training process, but it is not significant compared to CARLA's expensive runtime.
This enables us to use the gymnasium interface and build our code upon the CleanRL \cite{Shengyi2022Online, Huang2022JMLR} PPO implementation.

As documented in \cite{Li2024ECCV}, the CARLA leaderboard 2.0 has very slow reset times that can be up to 1 minute.
The reset consists of reloading the town, computing a new route with an A$^\star$ algorithm, instantiating all cars and scenarios, and setting up the ego agent.

We speed up this process by specializing each simulator to a particular town. This avoids the need to reload the entire town assets and only requires destroying past cars and re-instantiating new once.
PPO (Atari hyperparameters) uses 8 different simulators, which align nicely with the 8 small towns of CARLA (towns 1-7 and 10).
Roach (CARLA hyperparameters) uses 6 simulators, we use town 1-6 which aligns with the evaluation towns of longest6 v2.

We remove the computation bottleneck of the A$^\star$ algorithm by pre-computing it once for all of our experiments for a fixed set of randomly generated routes and storing the results on disk (see \secref{sec:scenario_generation}).
This increases RAM requirements, but modern servers have sufficient RAM capacity (our servers have > 100 GB per GPU).

When instantiating the agent, we do not reinitialize everything from scratch and only recompute values that are route-specific.

We train with the CARLA towns 1-7 and 10 and do not use the CARLA towns 11 - 13 or 15. This is because these towns are massive in size which significantly increases the runtime and reset speed of CARLA (and the reset can not be entirely avoided because only parts of the towns are loaded in memory at a time).

In this work, we have only modified the Python code of the CARLA leaderboard 2.0 client and used the original C++ simulation server.
A particular problem with the CARLA server is that it has various bugs and crashes occasionally, particularly with the autogenerated scenarios we use.
When scaling to over 100 concurrent CARLA servers, we observe that the occasional crashes become frequent crashes, as the probability that one of the 100 servers crashes increases.
We monitor our training code for such crashes and restart everything if one occurs. 
We have tuned tight timeout values and optimized startup time to minimize the impact of these CARLA crashes.
Nevertheless, the crashes are still a major bottleneck in our scaled-up runs, roughly doubling training time.
A promising direction for future work to speed up RL training with CARLA further is to fork the simulator, fix the bugs, and optimize bottleneck functions like the slow CARLA HD-map retrieval.

We run the simulator at 10 Hz during training, which is the lowest frequency at which the CARLA physics are still accurate according to the documentation.
During model evaluation, we run the simulator at the standard 20 Hz of the CARLA leaderboard 2.0 and use an action repeat of 2 for the policy.
In preliminary experiments, we found that decreasing the simulator frequency was more efficient than using action repeat during training.

\subsection{Scenario Generation}
\label{sec:scenario_generation}
An important problem when training with the CARLA leaderboard 2.0 is that there are only a very small number of training scenario definitions available.
These scenario definitions specify where a particular scenario type takes place and set the scenario hyperparameters.
These scenario definitions were manually labeled, which is why they are limited in quantity.
To generate the required number of samples for PPO with only the few available scenario definitions would require repeating the same routes many times and would likely lead to overfitting.
Instead, we wrote an automatic route and scenario generation script based on rejection sampling and various heuristics.
More principled solutions like self-play \cite{Cusumano-Towner2025ARXIV} or scenario generation methods \cite{Hanselmann2022ECCV, Hao2023TITS, Yin2024ECCVW} are not able to generate scenario types like construction sites, vehicle opening doors, or emergency vehicles required for the CARLA leaderboard 2.0. Extending these works to generate a wider variety of scenarios is a promising direction for future work.

The script works roughly as follows: First, we select a random point from the HD-map topology. 
We then pick the next point iteratively at 1-meter distance, choosing randomly at intersections. 
Outside of intersections, we change the lane with 10\% probability.
We generate more route points until the route is 1 km long.
We run various checks to check if the generated route is valid, such as can the ego vehicle be spawned at the initial point. 
If not, we discard the route.
Along the route, we then try to place scenarios of random types roughly every 100 meters. 
The scenario parameters are drawn from a random distribution, which is distributed according to how frequently the parameter occurs in the released handcrafted scenarios of CARLA.
We then try to instantiate the scenario with the CARLA leaderboard 2.0 code and reject scenarios that fail to set up.
If a generated scenario passes all validation checks, it gets added to the list of scenarios for that route. 
If not, we try another scenario type until all types have been tried.
Various scenario types have special rules to account for the scenario's specific needs.

The resulting generated routes have up to 10 scenarios in them.
The generated route and scenarios are noisy because our rejection tests are imperfect.
Additionally, the number of scenarios generated may not be balanced because scenarios that can be placed in many locations are oversampled.
Nevertheless, this script enables us to generate the large volume of unique training routes and scenarios needed to train PPO at scale.

On nuPlan, a large number of training scenarios are available, so generating scenarios was not necessary.

\subsection{nuPlan}

\boldparagraph{Dataset} The nuPlan recordings are stored in an SQLite database~\cite{Hipp2000Online}, whereas the maps are saved in a GeoPandas dataframe~\cite{Bossche2024Online}. To reduce database interactions, we first collect the scenarios in the selected training set and store them in smaller pickle files with gzip compression. We additionally limit the number of entities at each frame by selecting the closest objects and agents to the position of the human operator. This is beneficial in nuPlan due to frequent scenes with 50+ pedestrians and cars, which slow down dataloading and rendering, effectively idling other parallel environments. To this end, we set a maximum of 50 vehicles, 50 pedestrians, 10 bicycles, and 10 static objects. Our final dataset uses all temporally non-overlapping scenarios (of 15s duration) with valid route information from nuPlan's training split, resulting in 269,075 samples and a storage size of 84 GB. Notably, we do not pre-process and cache any maps and provide the full planner input interface of nuPlan in the environment. We expect further performance could be gained from pre-processing the map and recordings, which would come at the cost of flexibility when computing the reward and observation of an RL agent. 

\boldparagraph{Background Traffic} We directly integrate the reactive and non-reactive background traffic of the nuPlan simulator. In the non-reactive mode, the vehicles, pedestrians, bicycles, and objects are replayed as observed in logs. The reactive traffic simulates the vehicles along the centerline of the lane graph with an IDM lane following model~\cite{Treiber2000PRE}, while the remaining agent categories are replayed from the logs. For training CaRL on nuPlan, we randomly select between reactive and non-reactive traffic during initialization with equal probability. We note that the reactive traffic is considerably slower, making it less adequate for short-cycled testing. We leave further optimization to future work.

\boldparagraph{Rendering} The observation rendering remains a considerable performance bottleneck (in addition to the reactive background traffic). In each iteration, we need to query nearby map elements from the map interface, retrieve relevant agents, and render several polygons and polylines in the respective channels. We use OpenCV for CPU-based rendering, which we found to be the fastest CPU-based library during development. 

\subsection{Hardware}
Our code is CPU bottlenecked and highly parallelized, and benefits strongly from modern CPUs with many cores. We use servers with 12-24 physical cores per GPU for our experiments.
Contemporary machine learning servers are primarily built around powerful high-memory GPUs for supervised training, often neglecting the CPU  (with as low as 4 cores per GPU). 
We do not recommend such machines for our compute task since the low number of CPU cores means the compute job runs several times slower.
Our jobs were run on expensive A100 and H100 GPUs simply because servers with high-end CPUs typically also have high-end GPUs. We do not fully utilize these expensive GPUs since our model is small. More economical GPUs, like the A6000, might suffice for our compute tasks.
\section{Baselines}
\label{sec:baselines}

\subsection{CARLA}

In this section we describe our reproduction of the CARLA baselines PlanT and Think2Drive.

\subsubsection{PlanT}
We implement PlanT \citep{Renz2022CORL}, which is the state-of-the-art imitation-learning based baseline on CARLA. PlanT is a simple transformer-based planner that uses a sparse object representation for its inputs.
All objects are input as bounding boxes with six attributes: x, y, yaw, speed, x-extent, and y-extent.
Route boxes input the box ID instead of speed. The model outputs 4 waypoints for control using a GRU, which is additionally supplied with the target point and a traffic light flag.

Since the original PlanT model was trained using the leaderboard 1.0 version of longest6, we need to make some adjustments for a fair comparison. Most of these changes result from the increased ego speed, as described in \secref{sec:preliminaries}.

\begin{figure}[th]
\centering
    \includegraphics[height=0.37\textwidth]{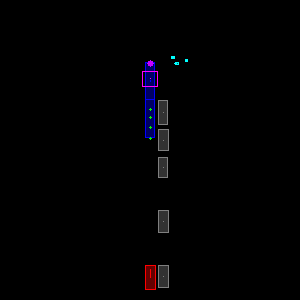}
    \includegraphics[height=0.37\textwidth]{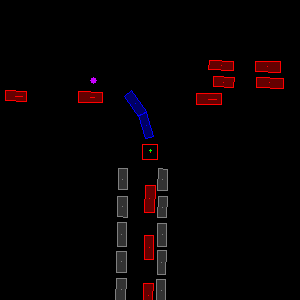}
    \includegraphics[height=0.37\textwidth]{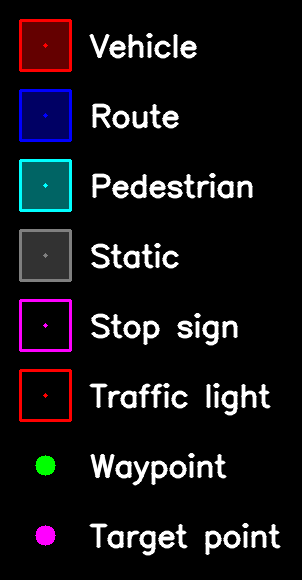}
   \caption{Training examples with the updated PlanT inputs. The image on the left shows the ego vehicle approaching a stop sign with pedestrians crossing the road. The second example shows the ego vehicle waiting to turn left at a red light. The ego vehicle is always at the center of the image.}
\label{fig:PlanT_input}
\end{figure}

\boldparagraph{Input} We add pedestrians and static obstacles as new object classes, following the representation of cars, since the original model did not consider them. Additionally, we add trigger boxes for ego-relevant stop signs and traffic lights to the object representation since the original flag-based approach was found to be ineffective at higher speeds and with the complexities of approaching and passing a stop sign. We only input red and yellow traffic lights and remove stop signs when the ego vehicle has slowed down enough to consider the stop sign as passed. \figref{fig:PlanT_input} shows two training examples with the updated inputs.

\boldparagraph{Model} We keep the model architecture unchanged except for removing the traffic light flag in the GRU and adding the 4 new class embeddings. Since the maximum driving speed increases from 60 to 120 km/h, we double the number of bins for forecasting the other vehicles' speed from 16 to 32 to keep the bin size the same.

\boldparagraph{Control} To accommodate faster driving speeds, we replace the original control logic. We use the lateral controller implemented by PDM-Lite \cite{Sima2024ECCV}, using the interpolated waypoints as the target route and calculate the desired speed based on the distance between the second and fourth waypoints. Even though the sampling rate of our dataset is 4 Hz, twice that of the original paper, we keep the number of waypoints at 4. During testing, both increasing the number of waypoints to 8 and sampling the waypoints at 2 Hz showed poor performance, which may be caused by the limited amount of route information given by the route boxes, combined with the long distance forecasts resulting from higher driving speeds.

\boldparagraph{Training} We use a publicly available dataset collected using PDM-Lite \cite{Zimmerlin2024ARXIV},  containing all 36 scenarios relevant to the leaderboard 2.0. At around 470k samples, it is similar in size to the 2x dataset used in the original paper.
We follow the original training regime, training for 47 epochs with a batch size of 128 and a learning rate of 1e-4.

\subsubsection{Think2Drive}
In this section, we describe our reimplementation of Think2Drive \cite{Li2024ECCV}, the code of which has not been officially released. Since several key hyperparameters are missing in the paper, we inferred them ourselves or used Roach \cite{Zhang2021ICCV} as a reference. Think2Drive is based on the model-based reinforcement learning algorithm DreamerV3 \cite{Hafner2025NATURE}, combining a world model with an actor-critic approach.

\textbf{Model:} The Think2Drive paper does not specify which code version they have used. We tested both versions corresponding to the first \citep{Hafner2023ARXIV} and second \citep{Hafner2025NATURE} DreamerV3 paper versions, finding that only the latter functioned properly. Consequently, we used the GitHub commit 251910d associated with the second paper version, which often outperformed the first version on most benchmarks, reported in the DreamerV3 paper. We selected the 100 million parameter model to match the model size used in Think2Drive and adjusted key hyperparameters accordingly. \tabref{tab:hyperparams_think2drive} shows the key hyperparameters used in our implementation.
\begin{table}[]
    \centering
    \begin{tabular}{l| c }
        \toprule
        \textbf{Hyperparameter} & \textbf{Value} \\
        \midrule
        Hidden size & 768 \\
        Recurrent units & 6,144 \\
        Base CNN channels & 96 \\
        Codes per latent & 48 \\
        Reward loss scale & 10.0 \\
        Replay buffer size & 300,000 \\
        Parameters & 125 Million \\
        \bottomrule
    \end{tabular}
    \vspace{0.2cm}
    \caption{Hyperparameters of our Think2Drive implementation. We used the second code version of DreamerV3 \cite{Hafner2025NATURE} and selected the 100M model to match Think2Drive's model capacity, while increasing encoder/decoder channels and reward loss scale similar to Think2Drive.}
    \label{tab:hyperparams_think2drive}
\end{table}

\textbf{Input:} As input, we use the same BEV and scalar values as Think2Drive. Each image channel represents a binary mask for specific static elements and dynamic objects. For dynamic objects, we render previous locations (transformed to the current ego vehicle's coordinate frame) from four timesteps: current (t) and historical frames (t-5, t-10, t-15), each timestep being 0.05 seconds. Scalar inputs are sampled at these same timesteps.

\textbf{Actions:} We adopt the 30 discrete actions defined in the Think2Drive paper. Our simulator runs at 20 Hz while action selection occurs at 10 Hz, resulting in each action being repeated once (action repeat = 2).

\textbf{Reward:} Similar to Think2Drive, we use a weighted sum as our reward function:
\begin{equation}
    r=1 \cdot r_{speed} + 1 \cdot r_{travel}+2 \cdot p_{deviation}+0.5 \cdot c_{steer}
\end{equation}
The speed reward $r_{speed}$ encourages the agent to match a target speed, calculated as in \eqref{equ:speed_reward}.
\begin{equation}
    r_{speed}=1-\frac{|v-v_{target}|}{7.5}
    \label{equ:speed_reward}
\end{equation}
This target speed is determined by taking the minimum of: (1) 80\% of the speed limit, (2) target speed considering the distance to the next red traffic light, (3) target speed considering the distance to the next uncleared stop sign, and (4) the target speed considering the distance to the next vehicle on the ego vehicles route. We linearly decrease the target speed given the distance to an obstructing actor, using \eqref{equ:target_speed}.
\begin{equation}
v_{target}=0.8 \cdot v_{limit} \cdot \frac{\text{clip}\{d - \text{safe margin}, 0, 12.5\}}{12.5}
\label{equ:target_speed}
\end{equation}
The safe margin is 8.0, 4.0, and 2.5 meters for vehicles, traffic lights and stop signs respectively.
For the travel reward $r_{travel}$, we use the distance traveled in meters. 
The deviation penalty $p_{deviation}$ penalizes deviations from the planned route as in \eqref{equ:route_deviation}.

\begin{equation}
p_{deviation}=-\frac{||p_{ego} - p_{route}||_2}{8.0}
\label{equ:route_deviation}
\end{equation}
where $p_{ego}$ is the ego vehicle position and $p_{route}$ is the nearest point on the route.
We terminate the episode if the ego vehicle runs a red light, runs a stop sign, collides with any type of object, or deviates more than 15 m from the route. In case the ego has reached the end of the route up to 10 m or did not drive faster than 0.1 m/s for at least once during the last 100 s, we truncate the episode. For terminal rewards, we set the reward to $-1$ if the episode terminates due to exceeding the route deviation. In case of a collision, running a red light, or running a stop sign, we set the reward to $-1-speed$ (where speed is the current velocity of the vehicle in m/s). If the vehicle reaches the end of the route, we assign a reward of $+1$.

\textbf{Data Collection:} We run 8 parallel CARLA instances (one for Towns 1-7 and Town 10), each generating 6 FPS, for a total throughput of 48 FPS.

\textbf{Training:} Following Think2Drive's approach, we implement curriculum learning over two training phases. First, we train 400,000 CARLA steps (800,000 in total, due to action repeat = 2) on basic routes with traffic lights, stop signs, and other vehicles. We then switch to the second training phase with 1.5 million CARLA steps on 1 km routes with scenarios along the route. We reinitialize both actor and critic parameters at 400,000 CARLA steps and subsequently every 800,000 steps throughout the second training phase.
Following the first parameter reinitialization, we linearly increase the actor-critic train ratio from 16 to 64, which means each sample is ultimately used an average of 64 times for training. The world model maintains a constant train ratio of 16 throughout the training. This creates training iterations where only the actor-critic parameters are updated while the world model remains frozen. During these actor-critic-only updates, we prevent gradient propagation through the world model to avoid the critic's loss from dominating the world model's training dynamics. Our sampling strategy mirrors Think2Drive's approach: 50\% samples are drawn uniformly from the replay buffer, while the remaining 50\% come from the $K = 64$ frames immediately preceding episode termination.

\subsection{nuPlan}
\label{sec:nuPlan_result}

This section provides additional baselines for nuPlan.
All baselines were evaluated using the officially published code and model.

\boldparagraph{Metrics} The nuPlan closed-loop score (CLS) is assigned to each simulation clip and combines sets of multiplier penalties $M$ and weighted metrics $W$. 
\begin{equation}
    \text{CLS} = \left( \prod_{m\in M} \text{score}_m \right) \times \left(\frac{\sum_{w \in W} \text{weight}_w \times \text{score}_w }{\sum_{w \in W} \text{weight}_w}\right) \in [0,100]
\end{equation}
The multiplier penalties include scores for no at-fault collisions (Coll.), drivable area compliance (Driv.), driving direction compliance (Dir.), and making progress (MP) to evaluate if the agent is stuck in a scenario. The weighted metrics contain scores for time-to-collision (TTC), route completion (RC), speed-limit compliance (Speed.), and comfort (Comf.). All scores are between 0-100, where higher values indicate better performance. The final CLS is an average over all scenario scores. We refer to~\cite{Karnchanachari2024ICRA} for further information. 

\boldparagraph{Baselines} (1) The IDM planner is a rule-based baseline provided in the official \texttt{nuplan-devkit} repository\footnote{\href{https://github.com/motional/nuplan-devkit}{https://github.com/motional/nuplan-devkit}}. The planner first retrieves a centerline path with a breadth-first-search on the lane graph and applies the IDM car-following model to control the ego-agent along the center path. (2) PDM-Closed is an extension of the IDM planner, that creates several trajectory proposals by combining different target speeds and centerline offsets. These proposals are simulated and scored with a similar metric to the nuPlan closed-loop score. (3) PLUTO uses a transformer-based architecture to forecast background agents and to regress multiple ego trajectories with corresponding scores. These outputs are then post-processed with the simulation and scoring modules of PDM-Closed to select a suitable ego trajectory, resulting in a hybrid planner. We use the code and checkpoint of the PLUTO repository\footnote{\href{https://github.com/jchengai/pluto}{https://github.com/jchengai/pluto}} for reproduction. (4) Urban Driver applies PointNet layers on a vector representation of the scene with multi-head attention for aggregation and outputs an ego trajectory. (5) GC-PGP utilizes a graph neural network on the lane graph and agents to output multiple route-conditioned trajectories. (6) PlanCNN uses a CNN encoder on a semantic BEV rendering of the scene and outputs an ego trajectory with an MLP head. (7) PlanTF uses a Transformer architecture to encode agents and map elements in vector representation, forecast non-ego agents, and regress a single ego trajectory. The code and checkpoint are publicly provided in the official repository\footnote{\href{https://github.com/jchengai/plantf}{https://github.com/jchengai/plantf}} (8) Diffusion Planner uses a DiT architecture to generate trajectories for ego and non-ego agents, while conditioning on a vector representation of the scene. Currently, the repository of Diffusion Planner~\footnote{\href{https://github.com/ZhengYinan-AIR/Diffusion-Planner}{https://github.com/ZhengYinan-AIR/Diffusion-Planner}} does not provide the guidance terms used for post-processing, but only code and checkpoints for purely learned planning. For PDM-Closed, Urban Driver, GC-PGP, and PlanCNN, we use the scripts and checkpoints from \texttt{tuplan\_garage}\footnote{\href{https://github.com/autonomousvision/tuplan\_garage}{https://github.com/autonomousvision/tuplan\_garage}}.

\boldparagraph{Results}
\tabref{tab:val14_2} shows the results on Val14 non-reactive traffic, while \tabref{tab:val14_3} displays the results on Val14 reactive traffic.
As shown in prior work \citep{Dauner2023CORL}, the scoring mechanism of PDM-Closed improves IDM significantly. 
Rule-based and hybrid planners outperform all imitation learning based baselines.
There is a particularly large gap in reactive traffic where the best open-source method, Diffusion Planner, still achieves 10 points CLS less than PDM-Closed.
CaRL closes this gap to 1.5 points.

To achieve the best results in a metric, the global optimum of the optimization objective should be aligned with the global optimum of the metric \cite{Jaeger2024FTO}.
The closed-loop score of nuPlan uses a weighted sum of 4 terms: route completion (RC), time to collision (TTC), speed limit compliance (Speed), and comfort (Comf.), which are traded off in a 5:5:4:2 ratio.
While our reward works on both CARLA and nuPlan, it might not be optimal for this particular metric and its tradeoffs.
We train another CaRL model, noted as CaRL (nuPlan tuned), where we modify the reward to reflect the tradeoffs of the nuPlan CLS and double the samples to 1 billion to maximize the CLS score.
Instead of using multiplicative penalties for TTC, comfort, and speed, we add them to RC with the same ratios as used in the metric.
This improves the score of CaRL by 2.6 points in non-reactive traffic and 2.5 points CLS in reactive traffic.
This results in CLS of 93.87 in non-reactive and 93.12 in reactive traffic, marking the first time that a purely learning-based model outperforms the rule-based PDM-closed method in both traffic modes (the concurrent work GIGAFLOW \citep{Cusumano-Towner2025ARXIV} also outperforms PDM-Closed but only reports numbers for reactive traffic).

This reward introduces tradeoffs that introduce local minima that the optimization gets stuck in.
When training with nuPlans (log-replay) traffic, this is not as much of a problem because the car is initialized at the (high) human speed, which means the model cannot avoid driving.
Additionally, staying stationary can result in collisions with non-reactive traffic.
This enables PPO to escape the local minimum of not driving (which gives optimal speed, comfort, and TTC rewards) early during training.
We observed in preliminary experiments that training with this reward results in degraded performance in CARLA because of this problem.
We therefore do not recommend this reward in general, but it is useful when the goal is to achieve maximum performance in nuPlan.

The trajectory of the human data annotator can be seen as an optimal for nuPlan, and it is often used as an upper-bound for the simulator.
Most simulators, including nuPlan, have some unavoidable simulation errors. 
A perfect score of 100 is usually not possible.

We use the human trajectory here to demonstrate that nuPlans provided LQR controller, which is used in most nuPlan papers to convert the trajectory predicted by the planner to the actions of the car, is suboptimal.
\tabref{tab:val14_2} shows that the performance of the human trajectory is 93.5 with the standard LQR controller of nuplan.
We also report the log replay performance with an iterative LQR (iLQR), which is a more accurate controller that uses more computation.
The iLQR improves the LQR controller by almost 3 CLS points, significantly raising the upper bound on nuPlan to 96.24. In particular, it improves collisions (Coll.) and drivable area compliance (Driv.).
Additionally, we report the result of perfect tracking (Perf.), which perfectly tracks the human trajectory by teleporting the bicycle model.
This raises the score by a small 0.15 and is the actual performance of the human driver (although the human does not necessarily follow the constraints of a bicycle model here).

We argue that simulators and benchmarks for autonomous driving methods should not make the control part of the simulation, as controllers are typically specific to a particular representation (here, trajectories) and can be suboptimal, as demonstrated here.
The iLQR controller achieves great performance, yet is expensive, and giving researchers the flexibility to choose their approach for control can lead to innovative solutions that reduce this cost.
In our work, we have merged planning and control into a single model which achieves both high performance and efficient inference.

As a side note, the human perfect tracking result might not be the best possible score one can achieve in nuPlan.
Human drivers often intentionally exceed the speed limit by a bit to progress faster. 
The lowest auxiliary metric of the human driver is the speed limit compliance (Speed.).
Almost all baselines are better at this metric, so it might be possible to achieve scores slightly above 96.4.

\begin{table*}[t!]
\centering
    \resizebox{\textwidth}{!}{
    \begin{tabular}{l|c|c|cccc|cccc}
        \toprule
        \multirow{2}{*}{\textbf{Method}} & \multirow{2}{*}{\textbf{Type}} & \multirow{2}{*}{\textbf{CLS} $\uparrow$} & \multicolumn{4}{c|}{\textbf{Multiplier} $\uparrow$} & \multicolumn{4}{c}{\textbf{Weighted} $\uparrow$} \\
        & & & \textbf{Coll.} & \textbf{Driv.} & \textbf{Dir.} & \textbf{MP} & \textbf{TTC} & \textbf{RC} & \textbf{Speed.} & \textbf{Comf.}\\
        \midrule
        IDM~\cite{Treiber2000PRE} & \multirow{2}{*}{Rule} & 75.60 & 88.55 & 93.56 & 99.28 & 95.44 & 78.89 & 85.05 & 97.43 & 93.47 \\
        PDM-Closed~\cite{Dauner2023CORL} & & 92.84 & \textbf{98.12} & 99.64 & \textbf{99.96} & 99.11 & 93.29 & 92.14 & \textbf{99.83} & 95.08 \\
        \midrule
        PLUTO~\cite{Cheng2024ARXIV} & Hybr. & 92.63 & 97.85 & 99.11 & 99.73 & 99.46 & \textbf{94.36} & 92.98 & 98.22 & 91.86 \\
        \specialrule{1pt}{0.25em}{0.25em} 
        Urban Driver~\cite{Scheel2021CORL} & \multirow{6}{*}{IL} & 50.42 & 69.14 & 79.25 & 95.21 & 97.67 & 61.99 & 91.50 & 81.75 & \textbf{98.75} \\
        GC-PGP~\cite{Hallgarten2023ITSC} & & 59.60 & 84.97 & 88.73 & 98.39 & 89.80 & 79.25 & 60.19 & 99.31 & 89.00 \\
        PlanCNN~\cite{Renz2022CORL} & & 70.81 & 87.66 & 91.95 & 96.74 & 94.54 & 82.38 & 83.81 & 97.86 & 87.39 \\
        PlanTF~\cite{Cheng2024ICRA} & & 84.63 & 94.23 & 95.89 & 99.20 & 98.75 & 89.00 & 90.67 & 97.68 & 93.20 \\
        PLUTO{*}~\cite{Cheng2024ARXIV} & & 88.74 & 95.84 & 98.48 & 99.42 & 98.57 & 92.49 & 89.43 & 98.00 & 96.60 \\
         Diff. Planner{*}~\cite{Zheng2025ICLR} & & 89.62 & 95.89 & 98.30 & 99.60 & \textbf{99.91} & 90.52 & 94.19 & 97.27 & 94.63 \\
        \midrule
        CaRL (Ours) & \multirow{2}{*}{RL} & 91.25 & 97.36 & 99.28 & 99.02 & 99.37 & 92.04 & 94.44 & 98.74 & 89.09 \\
        CaRL (nuPlan tuned) & & \textbf{93.87} & 97.23 & \textbf{100.00} & 99.78 & 99.55 & 93.47 & \textbf{96.14} & 98.41 & 97.14 \\
        \specialrule{1pt}{0.25em}{0.25em} 
        \textit{Log Replay (LQR)}  & \multirow{3}{*}{\textit{Human}} & \textit{93.53} & \textit{98.79} & \textit{97.94} & \textit{99.06} & \textit{100.00} & \textit{94.28} & \textit{99.00} & \textit{96.50} & \textit{99.28} \\
        \textit{Log Replay (iLQR)}  & & \textit{96.24} & \textit{99.55} & \textit{99.55} & \textit{99.24} & \textit{100.00} & \textit{96.51} & \textit{99.32} & \textit{96.46} & \textit{98.93} \\
        \textit{Log Replay (Perf.)}  & & \textit{96.39} & \textit{99.78} & \textit{99.46} & \textit{99.37} & \textit{100.00} & \textit{96.78} & \textit{99.28} & \textit{95.86} & \textit{99.20} \\
        \bottomrule
    \end{tabular}}
    \caption{\textbf{Performance with non-reactive traffic on Val14}  (nuPlan)}
    \label{tab:val14_2}
\end{table*}

\begin{table*}[t!]
\centering
    \resizebox{\textwidth}{!}{
    \begin{tabular}{l|c|c|cccc|cccc}
        \toprule
        \multirow{2}{*}{\textbf{Method}} & \multirow{2}{*}{\textbf{Type}} & \multirow{2}{*}{\textbf{CLS} $\uparrow$} & \multicolumn{4}{c|}{\textbf{Multiplier} $\uparrow$} & \multicolumn{4}{c}{\textbf{Weighted} $\uparrow$} \\
        & & & \textbf{Coll.} & \textbf{Driv.} & \textbf{Dir.} & \textbf{MP} & \textbf{TTC} & \textbf{RC} & \textbf{Speed.} & \textbf{Comf.}\\
        \midrule
        IDM~\cite{Treiber2000PRE} & \multirow{2}{*}{Rule} & 77.33 & 89.22 & 93.02 & 99.24 & 96.33 & 81.57 & 85.20 & 97.20 & 93.11 \\
        PDM-Closed~\cite{Dauner2023CORL} & & 92.13 & \textbf{97.90} & 99.46 & \textbf{99.96} & \textbf{99.11} & 93.83 & 90.26 & \textbf{99.83} & 94.72 \\
        \midrule
        PLUTO~\cite{Cheng2024ARXIV} & Hybr. & 89.66 & 97.09 & 99.28 & 99.87 & 98.57 & \textbf{94.10} & 86.11 & 98.86 & 91.86 \\
        \specialrule{1pt}{0.25em}{0.25em} 
        Urban Driver~\cite{Scheel2021CORL} & \multirow{6}{*}{IL} & 53.05 & 72.45 & 83.36 & 96.87 & 94.28 & 66.37 & 87.17 & 81.66 & \textbf{99.28} \\
        GC-PGP~\cite{Hallgarten2023ITSC} & & 55.28 & 83.94 & 88.37 & 98.35 & 85.51 & 79.43 & 57.67 & 99.26 & 92.04 \\
        PlanCNN~\cite{Renz2022CORL} & & 70.14 & 87.88 & 91.59 & 97.09 & 93.92 & 83.18 & 83.94 & 97.48 & 86.85 \\
        PlanTF~\cite{Cheng2024ICRA} & & 76.14 & 95.21 & 96.33 & 99.33 & 89.53 & 90.61 & 77.21 & 98.50 & 93.56 \\
        PLUTO{*}~\cite{Cheng2024ARXIV} & & 77.97 & 96.69 & 97.59 & 99.69 & 87.66 & 93.92 & 74.56 & 98.70 & 95.62 \\
         Diff. Planner{*}~\cite{Zheng2025ICLR} & & 82.73 & 93.07 & 97.85 & 99.82 & 96.42 & 88.19 & 85.88 & 98.29 & 89.45 \\
        \midrule
        CaRL (Ours) & \multirow{2}{*}{RL} & 90.60 & 97.05 & 99.91 & 99.15 & \textbf{99.11} & 92.31 & 91.30 & 99.36 & 88.55 \\
        CaRL (nuPlan tuned) & & \textbf{93.12} & 97.09 & \textbf{100.00} & 99.91 & \textbf{99.11} & 93.29 & \textbf{93.56} & 99.17 & 97.05 \\
        \specialrule{1pt}{0.25em}{0.25em} 
        \textit{Log Replay (LQR)}  & \multirow{3}{*}{\textit{Human}}  & 80.32 & 85.64 & 97.94 & 99.06 & 100.00 & 79.79 & 99.00 & 96.50 & 99.28 \\
        \textit{Log Replay (iLQR)}  & & 82.05 & 85.69 & 99.55 & 99.28 & 100.00 & 80.77 & 99.32 & 96.46 & 98.93 \\
        \textit{Log Replay (Perf.)}  & & 82.02 & 85.60 & 99.46 & 99.37 & 100.00 & 80.77 & 99.28 & 95.86 & 99.20 \\
        \bottomrule
    \end{tabular}}
    \caption{\textbf{Performance with reactive traffic on Val14}  (nuPlan) $^*$without post-prosessing}
    \label{tab:val14_3}
    \vspace{-0.5cm}
\end{table*}

\section{Reward}
\label{sec:reward}

This section describes the details of our proposed reward.

\begin{equation}
    r_t = RC_t * \left(\prod p_t\right) - T
\label{eqn:infraction_multiplier2}
\end{equation}

$RC_t$ is the percentage of the route completed during the current time step.

\subsection{Terminal Penalties}
\label{sec:reward_hard}

Our reward uses 6 terminal penalties, which lead to termination of the episode if violated.
\begin{itemize}
    \item \textbf{Collision:} The ego vehicle collided with any object.
    \item \textbf{Off road:} The ego vehicle drove off the drivable area.
    \item \textbf{Run red light:} The ego vehicle entered the intersection while its traffic light was red.
    \item \textbf{Run stop sign:} The ego vehicle crossed a stop sign without decelerating to 0 m/s first.
    \item \textbf{Route deviation:} The ego vehicle deviated from the route by more than 30 meters (wrong turn at intersections).
    \item \textbf{Blocked:} The ego vehicle did not move faster than 0.1 m/s for more than 90 seconds
\end{itemize}

The terminal penalty $T$ is 1.0 for violating red lights and collisions and 0.0 for the other infractions.

\subsection{Soft Penalties}
\label{sec:reward_soft}
We use the following soft penalties $p_t$ that multiplicatively reduce the obtained route completion if violated:

\begin{itemize}
    \item \textbf{Outside lanes:} The ego vehicle drives on the sidewalk or on lanes of the opposing traffic direction. $p_t = 0$
    \item \textbf{Distance to nearest lane center:} The ego vehicle is penalized by the lateral distance to the nearest centerline. The penalty factor $p_t$ interpolates linearly from 1.0 while driving on a centerline to 0.0 while driving on a lane marking. Only applied outside intersections.
    \item \textbf{Speeding:} The agent exceeds the current speed limit. The penalty factor $p_t$ is linearly decreased from 1.0 to 0.0 while the agent is between 0-8 km/h too fast.
    \item \textbf{Time to collision:} If the agent violated the time to collision (TTC) metric. $p_t = 0.5$
    \item \textbf{Comfort:} The agent is penalized if it exceeds certain comfort thresholds. There are 6 comfort metrics described below. Each reduce the penalty by $p_t = 1.0 - 0.5 * \#/6$ with $\#$ being the number of violated comfort constraints. 
\end{itemize}

The penalty "Distance to nearest lane center" is there to encode the purpose of lane markings. Without it, the agent still learns to drive but does not know about the concept of lanes, leading to an unnatural-looking driving style that only considers drivable area.
Speed limits are initialized at 30 km/h when vehicles are spawned in CARLA and updated once the vehicles pass the next speed sign.
The comfort bounds in CARLA are shown in \tabref{tab:comfort_values}.

\begin{table*}[t!]
\centering
\begin{tabular}{l|l|l|l}
\toprule
\textbf{Parameter} & \textbf{CARLA} & \textbf{nuPlan} & \textbf{Unit} \\
\midrule
Longitudinal acceleration & $(-20, 10) $ & $(-4.05, 2.40)$ & $\text{m}/\text{s}^{2}$\\
Lateral acceleration & $(-9, 9)$ & $(-4.89, 4.89)$ & $\text{m}/\text{s}^{2}$\\
Absolute Jerk & $(-30, 30)$ & $(-8.37, 8.37)$& $\text{m}/\text{s}^{3}$\\
Longitudinal Jerk & $(-30, 30)$ & $(-4.13, 4.13)$ & $\text{m}/\text{s}^{3}$\\
Yaw rate & $(-1.0, 1.0)$ & $(-0.95, 0.95)$ & $\text{rad}/\text{s}$\\
Yaw acceleration & $(-3.0, 3.0)$ & $(-1.93, 1.93)$ & $\text{rad}/\text{s}^2$\\
\bottomrule
\end{tabular}
\caption{\textbf{Comfort Thresholds.}}
\label{tab:comfort_values}
\end{table*}

Comfort penalties and TTC if violated are applied for the next 500 simulator steps in CARLA and until the end of the episode in nuPlan.
All other soft penalties are applied for one simulator step.
For the CARLA vehicle physics, there are no human reference bounds for comfort, which is a subjective feeling and hence needs to be tuned based on data.
We set quite loose bounds to ensure the agent can adhere to them, but future work may tune the bounds to achieve more comfortable policies.

Time to collision (TTC) is computed by performing a constant speed forecast of all actors in the scene. We forecast for 1 second, discretized to 5 200 ms intervals, with a kinematic bicycle model. TTC is violated if the ego bounding box intersects with another bounding box in the forecast.

\subsection{nuPlan Specific Adaptations}

\boldparagraph{Terminal Penalties} (1) \textbf{Collision}: We terminate the simulation if the bounding box of an ego vehicle intersects with another agent or static object. We ignore collisions if the ego vehicle is stationary (\ie has a speed of $<0.05 \text{ m}/\text{s}$), due to some unavoidable collisions with non-reactive agents. (2) \textbf{Off road:} The drivable area in nuPlan is defined by the polygons of the road and car-park areas. If the ego agent leaves the drivable surface, we terminate the simulation. We do not use the terminal penalties for (3) \textbf{Red light} and (4) \textbf{Stop sign}, as they are not accounted for in the nuPlan CLS and pose several challenges for implementation. For example, the human vehicle operator regularly drives over red lights and stop sign polygons, due to label noise and human behavior, making an optimal score not feasible. Similarly, we ignore the (5) \textbf{Route deviation} and (6) \textbf{Blocked} penalties, as it is hard to select a distance or duration threshold for these penalties in the short durations of the nuPlan simulation (\ie~15 seconds). 

\boldparagraph{Soft Penalties} (1) \textbf{Outside Lanes}: We apply a penalty of $p_t = 0$ if the ego agent leaves the route polygons, which coincides with driving in the opposite traffic or car park areas. For (2) \textbf{Distance to nearest lane center}, we allow a deviation of $0.5 \text{m}$ (\ie $p_t = 1.0$) from the lane center and linearly decrease the penalty from $1.0$ to $0.0$ starting from $0.5 \text{m}$ to the lane marking. Allowing some center deviation was necessary, as driving in the middle of the lane can be suboptimal in nuPlan (\eg if cars are parked on the side of the road). The (3) \textbf{Speeding}, (4) \textbf{TTC}, and (5) \textbf{Comfort} penalties are defined as in \secref{sec:reward_soft}. Note that we use stricter comfort boundaries, as shown in \tabref{tab:comfort_values}, which are the same thresholds used in the nuPlan comfort metric and selected based on human trajectories from the dataset~\cite{Karnchanachari2024ICRA}. 

\boldparagraph{Survival Bonus} Due to the short training scenes in nuPlan, it is then possible to reach 100\% of the route completion before the simulation ends. In this case, the reward from \eqref{eqn:infraction_multiplier2} does not provide feedback to the ego agent and no incentives to avoid terminal penalties. We found that PPO planners would behave aggressively to foster short-term progress while putting less of an emphasis on collision avoidance. We circumvent this problem by combining the regular reward with a constant survival bonus, as follows 
\begin{equation}
    r_t^{*} = (1-s) \times r_t + \frac{s \times 100}{N},
\end{equation}
where $s\in[0,1]$ is the survival ratio and $N$ the total number of simulation iterations (\ie ~$150$ for $15\text{s}$ at $10\text{Hz}$). The weighted combination ensures the sum of rewards remains in the range of 0-100, as previously. We use a survival ratio of $s = 0.6$.

\end{document}